\documentclass[sigconf]{acmart}
\AtBeginDocument{%
  }
\usepackage{amsmath}    % 数学公式支持（如 align, \text 等）
\usepackage{amsthm}
\usepackage{mathrsfs}
\usepackage{mathtools}
\usepackage{booktabs}
\usepackage{url}            % simple URL typesetting
\usepackage{bm}
\usepackage{nicefrac}       % compact symbols for 1/2, etc.
\usepackage{adjustbox}
\usepackage{blindtext}
\usepackage{enumitem}
\usepackage{listings}
\usepackage{pifont}
\usepackage{multirow}
\usepackage{multicol}
\usepackage{makecell}
\usepackage{algorithm}
\usepackage{rotating}
\usepackage{algorithm}
\usepackage{algorithmic}
\usepackage{colortbl}
\usepackage{soul}
\usepackage{tcolorbox}
\usepackage{mathrsfs}

\usepackage{url}            % simple URL typesetting
\usepackage{bm}
\usepackage{booktabs}       % professional-quality tables
\usepackage{nicefrac}       % compact symbols for 1/2, etc.
\usepackage{microtype}      % microtypography
\usepackage{adjustbox}
\usepackage{blindtext}
\usepackage{enumitem}
\usepackage{listings}
\usepackage{pifont}
\usepackage{epsfig} % for postscript graphics files
\usepackage{times}
\usepackage{color}
\usepackage{multirow}
\usepackage{multicol}
\usepackage{makecell}
\usepackage{algorithm}
\usepackage{rotating}
\usepackage{subfigure}
\usepackage[T1]{fontenc}
\usepackage{algorithm}
\usepackage{algorithmic}
\usepackage{caption}
\usepackage[capitalize,noabbrev]{cleveref}
\usepackage{threeparttable}
\usepackage{wrapfig}
\usepackage{colortbl}
\usepackage{color, soul}
\usepackage{arydshln} % For dashline
\usepackage{tcolorbox}
\usepackage[capitalize,noabbrev]{cleveref}
\definecolor{babyblue}{rgb}{0.54, 0.81, 0.94}
\definecolor{bisque}{rgb}{1.0, 0.89, 0.77}
\definecolor{bshade}{rgb}{0.55,0.75,0.95}

\definecolor{mygray}{gray}{.6}
\definecolor{myblue}{RGB}{89,158,254}
\definecolor{mygreen1}{RGB}{81,150,111}
\definecolor{mygreen2}{RGB}{93,174,86}
\definecolor{myred}{RGB}{160,0,0}
\definecolor{myyellow}{RGB}{227,207,87}
\theoremstyle{plain}

\theoremstyle{definition}

\theoremstyle{remark}

\definecolor{babyblue}{rgb}{0.54, 0.81, 0.94}
\definecolor{bisque}{rgb}{1.0, 0.89, 0.77}
\definecolor{bshade}{rgb}{0.55,0.75,0.95}

\definecolor{mygray}{gray}{.6}
\definecolor{myblue}{RGB}{89,158,254}
\definecolor{mygreen1}{RGB}{81,150,111}
\definecolor{mygreen2}{RGB}{93,174,86}
\definecolor{myred}{RGB}{160,0,0}
\definecolor{myyellow}{RGB}{227,207,87}
\usepackage{bbding}
\newcommand{\mA}{\boldsymbol{A}}
\newcommand{\mE}{\boldsymbol{E}}
\newcommand{\mF}{\boldsymbol{F}}
\newcommand{\mH}{\boldsymbol{H}}
\newcommand{\mI}{\boldsymbol{I}}
\newcommand{\mM}{\boldsymbol{M}}
\newcommand{\mO}{\boldsymbol{O}}
\newcommand{\mP}{\boldsymbol{P}}
\newcommand{\mQ}{\boldsymbol{Q}}

\newcommand{\mX}{\boldsymbol{X}}
\usepackage{enumitem}
\let\oldding\ding% Store old \ding in \oldding
\renewcommand{\ding}[2][1]{\scalebox{#1}{\oldding{#2}}}
\usepackage[textsize=tiny]{todonotes}

\newcommand{\cc}{\color[rgb]{0,0.6,0.3}\checkmark}
\newcommand{\xx}{\color[rgb]{0.6,0,0}{\ding{55}}}
\copyrightyear{2025}
\acmYear{2025}
\setcopyright{acmlicensed}\acmConference[MM '25]{Proceedings of the 33rd
ACM International Conference on Multimedia}{October 27--31, 2025}{Dublin, Ireland}
\acmBooktitle{Proceedings of the 33rd ACM International Conference on
Multimedia (MM '25), October 27--31, 2025, Dublin, Ireland}
\acmDOI{10.1145/3746027.3755433}
\acmISBN{979-8-4007-2035-2/2025/10}

\begin{document}

\title{MQuant: Unleashing the Inference Potential of Multimodal Large Language Models via Static Quantization}

\author{JiangYong Yu}
\authornotemark[1]
\affiliation{
  \institution{HOUMO AI}
  \city{Nanjing}
  \country{China}
}
\email{jiangyongyufocus@gmail.com}

\author{Sifan Zhou}
\authornote{Both authors contributed equally to this work.}
\affiliation{
  \institution{Southeast University}
  \city{Nanjing}
  \country{China}
}
\email{sifanjay@gmail.com}

\author{Dawei Yang}
\authornote{Corresponding Author.}
\affiliation{
  \institution{HOUMO AI}
  \city{Nanjing}
  \country{China}
}
\email{dawei.yang@houmo.ai}

\author{Shuoyu Li}
\affiliation{
  \institution{Xi'an Jiaotong University}
  \city{Xi'an}
  \country{China}
}
\email{1027057721@qq.com}

\author{Shuo Wang}
\affiliation{
  \institution{HOUMO AI}
  \city{Nanjing}
  \country{China}
}
\email{wangshuo514@sina.com}

\author{Xing Hu}
\affiliation{
  \institution{HOUMO AI}
  \city{Nanjing}
  \country{China}
}
\email{xing.hu@houmo.ai}

\author{Chen Xu}
\affiliation{
  \institution{HOUMO AI}
  \city{Nanjing}
  \country{China}
}
\email{xuchen19970925@gmail.com}

\author{Zukang Xu}
\affiliation{
  \institution{HOUMO AI}
  \city{Nanjing}
  \country{China}
}
\email{zukang.xu@houmo.ai}

\author{Changyong Shu}
\affiliation{
  \institution{HOUMO AI}
  \city{Nanjing}
  \country{China}
}
\email{changyong.shu89@gmail.com}

\author{Zhihang Yuan}
\affiliation{
  \institution{HOUMO AI}
  \city{Nanjing}
  \country{China}
}
\email{hahnyuan@gmail.com}

% \author{JiangYong Yu, Sifan Zhou, Dawei Yang, Shuo Wang, Shuoyu Li, Xing Hu, \\ Chen Xu, Zukang Xu, Changyong Shu, Zhihang Yuan}
% \email{jiangyongyufocus@gmail.com, sifanjay@gmail.com, dawei.yang@houmo.ai}

% \renewcommand{\shortauthors}{Jiangyong Yu, Sifan Zhou et al.}
\renewcommand{\shortauthors}{JiangYong Yu \& Sifan Zhou et al.}
% \renewcommand\footnotetextcopyrightpermission[1]{}

% \settopmatter{printacmref=false}
% \authornote{Both authors contributed equally to this research.}
% \email{trovato@corporation.com}
% \orcid{1234-5678-9012}
% \author{G.K.M. Tobin}
% \authornotemark[1]
% \email{webmaster@marysville-ohio.com}
% \affiliation{%
%   \institution{Institute for Clarity in Documentation}
%   \city{Dublin}
%   \state{Ohio}
%   \country{USA}
% }

% \author{Lars Th{\o}rv{\"a}ld}
% \affiliation{%
%   \institution{The Th{\o}rv{\"a}ld Group}
%   \city{Hekla}
%   \country{Iceland}}
% \email{larst@affiliation.org}

%%
%% The abstract is a short summary of the work to be presented in the
%% article.
\begin{abstract}
Multimodal large language models (MLLMs) have garnered widespread attention due to their ability to understand multimodal input. However, their large parameter sizes and substantial computational demands severely hinder their practical deployment and application.
While quantization is an effective way to reduce model size
and inference latency, its application to MLLMs remains underexplored. In this paper, we propose MQuant, a post-training quantization (PTQ) framework designed to tackle the unique challenges of multimodal large language models (MLLMs). Conventional quantization often struggles with MLLMs because of (a) high inference latency from large visual token counts, (b) distributional disparities between visual and textual tokens, and
(c) extreme outliers introduced by Hadamard-based transformations. To address these issues, MQuant introduces: • Modality-Specific Static Quantization (MSQ), assigning distinct static scales for visual vs. textual tokens; • Attention-Invariant Flexible Switching (AIFS), reordering tokens to preserve casual attention while eliminating expensive token-wise scale computations; • Rotation Magnitude Suppression (RMS), mitigating weight outliers arising from online Hadamard rotations. On five mainstream MLLMs (including Qwen-VL, MiniCPM-V, CogVLM2), MQuant under W4A8 achieves near-floating-point accuracy (<1\% degradation) while reducing inference latency by up to 30\%, significantly outperforming existing PTQ baselines. Our MQuant effectively bridges the gap for efficient and accurate MLLMs inference in resource-constrained devices. Code is released in \url{https://github.com/StiphyJay/MQuant}.
\end{abstract}

% \begin{CCSXML}
% <ccs2012>
%  <concept>
%   <concept_id>00000000.00000000.00000000</concept_id>
%   <concept_desc>Do Not Use This Code, Generate the Correct Terms for Your Paper</concept_desc>
%   <concept_significance>100</concept_significance>
%  </concept>
% </ccs2012>
% \end{CCSXML}

% \ccsdesc[500]{Do Not Use This Code~Generate the Correct Terms for Your Paper}
% \ccsdesc[300]{Do Not Use This Code~Generate the Correct Terms for Your Paper}
% \ccsdesc{Computing methodologies~\textbf{Computer Vision}}
\begin{CCSXML}
<ccs2012>
   <concept>
       <concept_id>10010147.10010178.10010179.10010182</concept_id>
       <concept_desc>Computing methodologies~Natural language generation</concept_desc>
       <concept_significance>500</concept_significance>
       </concept>
   <concept>
       <concept_id>10010147.10010178.10010224.10010225.10010231</concept_id>
       <concept_desc>Computing methodologies~Visual content-based indexing and retrieval</concept_desc>
       <concept_significance>500</concept_significance>
       </concept>
 </ccs2012>
\end{CCSXML}

\ccsdesc[500]{Computing methodologies~Natural language generation}
\ccsdesc[500]{Computing methodologies~Visual content-based indexing and retrieval}

% \ccsdesc[100]{Do Not Use This Code~Generate the Correct Terms for Your Paper}

\keywords{Multimodal Large Language Model, Model Compression, Model Efficiency, Post-Training Quantization}

%%
%% The code below is generated by the tool at http://dl.acm.org/ccs.cfm.
%% Please copy and paste the code instead of the example below.
%%

%%
%% Keywords. The author(s) should pick words that accurately describe
%% the work being presented. Separate the keywords with commas.

%%
%% This command processes the author and affiliation and title
%% information and builds the first part of the formatted document.
\maketitle
\vspace{-2mm}
\section{Introduction}
\label{intro}
% \vspace{-2mm}

\begin{figure*}[t]
    \centering    
    \includegraphics[width=1.0\textwidth]{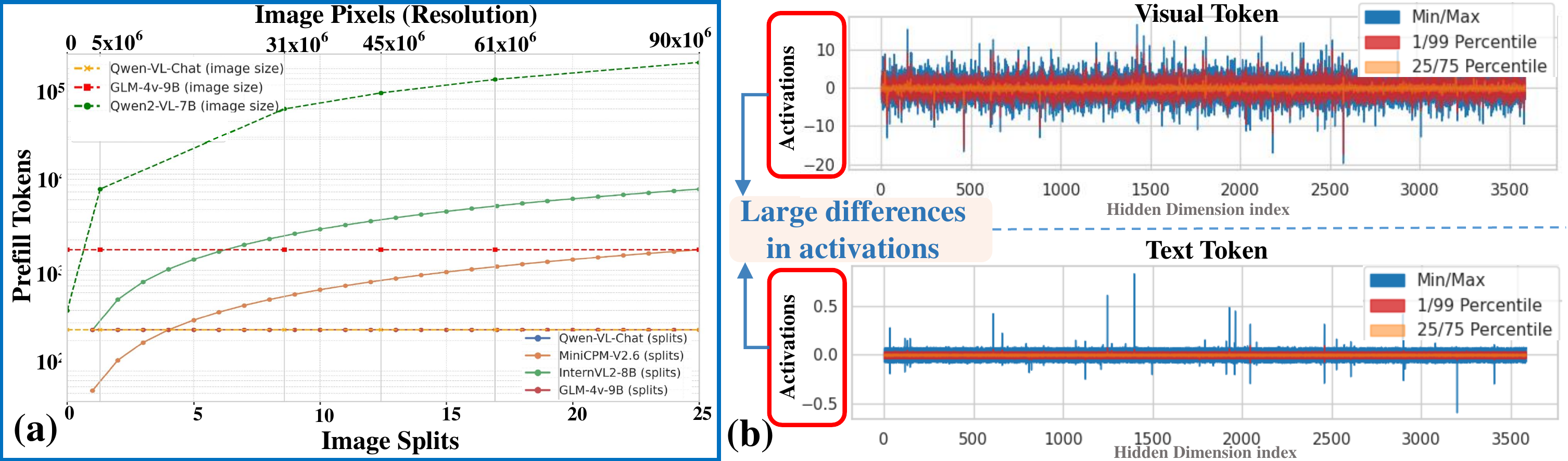}
    \vspace{-2.9mm}
    \caption{(a) Prefill visual tokens counts across different MLLMs as the image splits/resolution increases. (b) The activation values of visual tokens range from $-20$ to $10$, while textual tokens concentrate near $0$ (rarely exceeding $\pm0.5$.} 
    \label{fig:obs12}
\vspace{-3mm}
\end{figure*}
Recent advances in large language models (LLMs)~\citep{brown2020language,touvron2023llama,llama2,llama3} have led
to remarkable performance on a wide range of natural language processing tasks. However, these models often struggle
when dealing with non-textual data such as images or videos. Multimodal large language models (MLLMs)~\citep{reid2024gemini,achiam2023gpt4,wang2023cogvlm}
address this limitation by integrating visual and textual modalities, thereby enabling comprehensive understanding
and reasoning. Despite these benefits, their large parameter sizes, coupled with substantial computational demands, poses
a major challenge for real-world deployment, particularly in resource-constrained or privacy-sensitive environments.

\textbf{Challenges in MLLMs Quantization.} Quantization has proven an effective strategy for reducing memory usage
and inference costs in LLMs~\citep{yuan2023rptq,2023omniquant,gsq}, by converting high-precision parameters (e.g., FP32)
into lower-bit representations (e.g., INT8). Yet, the transition from LLMs to MLLMs brings three unique difficulties: \raisebox{-0.5pt}{\ding[1.1]{182\relax}} \textbf{Time-to-First-Token (TTFT) Explosion.} MLLMs often generate large numbers of visual tokens (e.g., patch embeddings or region proposals) with the resolution and aspect ratio of input images or videos. As shown in Fig \ref{fig:obs12}(a), in models such as Qwen2-VL~\citep{Qwen2VL}, the number of prefill visual tokens grows as image resolution increases (detailed figure in Appendix A.6). As image or video resolution increases, the prefill visual tokens can escalate dramatically---leading to high TTFT and negatively impacting latency-sensitive tasks. Moreover, in higher-demand scenarios such as video-based tasks and multi-image dialogues, the accumulation of visual tokens becomes even more pronounced, further exacerbating TTFT growth. Although the commonly adopted per-token dynamic quantization~\cite{qwenvl,Qwen2VL,liu2023llava,chen2024internvl} offers flexibility to accommodate the larger activation distribution variance across different tokens in vision modal, its requirement for online token-wise scale recalculation during inference significantly amplifies this computational overhead. Moreover, recent studies~\citep{chen2024prefixquant, tan2024mobilequant} have underscored the inefficiencies of per-token dynamic quantization, especially for resource-constrained edge devices, where such methods exacerbate latency and memory pressures. \raisebox{-0.5pt}{\ding[1.1]{183\relax}} \textbf{Disparate Modality Distributions.} As shown in Fig \ref{fig:obs12} (b), the activation distributions between visual and textual tokens reveal substantial numerical discrepancies, where visual token activations can span a significantly broader range (e.g., $-20$ to $10$) than textual tokens, which typically center near $0$. A single global scale factor for both visual and textual tokens leads to either aggressive clipping of visual outliers or increased quantization granularity for text, harming overall accuracy. \raisebox{-0.5pt}{\ding[1.1]{184\relax}} \textbf{Visual Token Outliers.} High-magnitude outliers in visual tokens, often reflecting salient image regions, make traditional clipping-based methods unsuitable. As shown in Table~\ref{fig:abla_visual_txt}, naive clipping outliers significantly degrades accuracy, highlighting the necessity for judicious outlier treatment.

\begin{table}[h]
\renewcommand\arraystretch{1.0} 
\centering
\small
% \vspace{-2mm}
\caption{Ablations on the clip range for different tokens.}
\vspace{-3mm}
\label{fig:abla_visual_txt}
\setlength{\tabcolsep}{0.8mm}
\scalebox{1.0}{
\resizebox{\columnwidth}{!}{
\begin{tabular}{cc|cc}
\toprule
Clip Range & Bits (W/A) & Visual Tokens & Textual Tokens \\ \midrule
-& BF16 / BF16 & \multicolumn{2}{c}{\textbf{61.40}} \\ \midrule
(0-1.0) & BF16 / INT16 & 61.20 (\color{myred}{$\downarrow$0.20}) & 61.25 (\color{myred}{$\downarrow$0.15})\\
\rowcolor{myblue!20}(0-0.99999) & BF16 / INT16 & \textbf{18.92} (\textbf{\color{myred}{$\downarrow$42.48}}) & 60.09 (\color{myred}{$\downarrow$1.31}) \\
\hline
\end{tabular}%
}
}
\vspace{-4mm}
\end{table}

To this end, we propose \emph{MQuant}, an accurate and efficient post-training quantization (PTQ) solution specifically designed for multimodal large language models (MLLMs). First, to reduce the TTFT while maintaining accuracy, we introduce \textbf{Modality-Specific Static Quantization (MSQ)}, which apply distinct static per-tensor scales for visual vs. textual tokens based on the distinct distribution differences. Second, for further acceleration, we design an \textbf{Attention-Invariant Flexible Switching (AIFS)} scheme. AIFS transforms mixed multimodal tokens into unified, modality-decoupled tokens without the need for dynamic position vectors, maintaining computational equivalence and memory efficiency by avoiding the overhead associated with dynamic processing. Based on MSQ and AIFS, MQuant
slashes TTFT and preserves accuracy. Third, we reveal the weight outliers caused by the online Hadamard rotations through theoretical analysis and propose \textbf{Rotation Magnitude Suppression (RMS)} to mitigate them. Our theoretical analysis reveals the emergence of large mean channels in MLLM weights, and RMS effectively reduces these outliers with minimal overhead. 

We evaluate \emph{MQuant} on five mainstream MLLMs, including InternVL~\citep{internvl15}, Qwen-VL~\citep{qwenvl},  MiniCPM-V~\citep{yao2024minicpmv}, CogVLM2~\citep{CogVLM2} and Qwen2-VL~\citep{Qwen2VL}. The extensive results on diverse multimodal reasoning tasks demonstrate that \emph{MQuant} achieves less than 1\% accuracy loss on all MLLMs with 23\% and 100\% speedup in prefill and decode stage under the W4A8 setting, highlighting its superior performance and wide practical value. Our main contributions are summarized as follows:

\vspace{-1mm}
\begin{itemize}
    \item \textbf{Insightful Observation.} To our best knowledge, we present the first comprehensive analysis of quantization issues in MLLMs, revealing the root causes of performance collapse and identifying inference speed bottlenecks as well as the quantization challenges posed by modality differences. 
    \item \textbf{MSQ and AIFS.} We design Modality-specific Static Quantization (MSQ) and Attention-Invariant Flexible Switching (AIFS) to accelerate inference for heterogeneous variable-sequence multimodal inputs while maintaining accuracy.
    \item \textbf{RMS.} We identify weight outliers caused by online rotation, and propose Rotation Magnitude Suppression to enhance quantization performance.
    \item \textbf{MQuant.} We propose \emph{MQuant}, a general PTQ framework designed for MLLMs, demonstrating both near-lossless performance and with significant speedup. 
    \vspace{-2.0mm}
\end{itemize}
\section{Preliminaries and Related Work}
\subsection{Multimodal Large Language Models}
\begin{figure}[h]
    \centering    
    \includegraphics[width=0.6\linewidth]{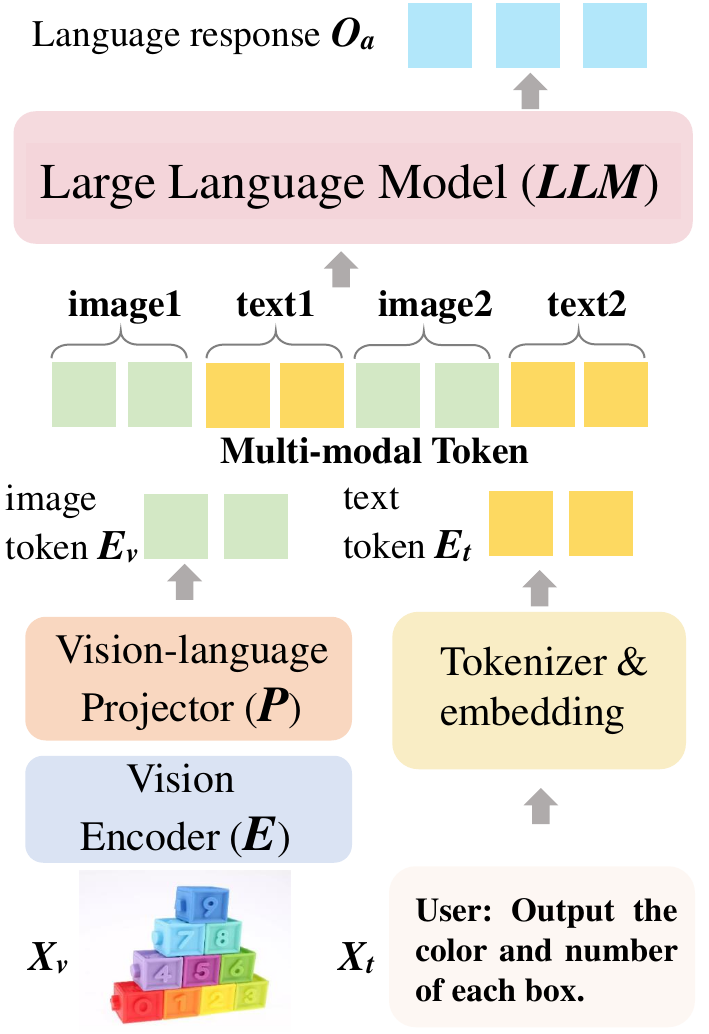}
    \vspace{-2.9mm}
    \caption{MLLM's architecture.} 
    \label{fig:mllms}
\vspace{-3mm}
\end{figure}

As shown in the Fig.~\ref{fig:mllms}, the existing MLLMs framework ~\citep{qwenvl,chen2024internvl} mainly consists of three modules: \textbf{(1)} \textbf{Visual encoder $\mE$} for processing visual inputs, \textbf{(2)} \textbf{Vision-language projector $\mP$} for aligning the text and visual modalities, \textbf{(3)} \textbf{Large language model $LLM$} that handles the multi-modal tokens and performs reasoning. 
\paragraph{Vision Encoder.} Taking the input image or video $\mX_v$ as input and compressing the original vision information into more compact patch features $\mF_v$. This process typically utilizes a Vision Transformer (ViT) ~\citep{vit}, such as CLIP~\citep{radford2021clip} and OpenCLIP~\citep{openclip}. It can be formulated as: $\mF_v = E(\mX_v)$.
\paragraph{Vision-Language Projector.} The task of the vision-language projector $\mP$ is to map the visual patch features $\mF_v$ into the textual feature space: $\mE_v = P(\mF_v)$.
\paragraph{Large Language Model.} The pre-trained large language model is the core component of MLLMs, endowing the framework with exceptional capabilities, such as zero-shot generalization, instruction following, and in-context learning. Typically, a text tokenizer is integrated with the LLM, mapping text prompts $\mX_t$ to the text tokens $\mE_t$. The text tokens $\mE_t$ and the visual tokens $\mE_v$ are then concatenated to form the input for MLLMs, which outputs the final response sequence $\mO_a$ in an autoregressive manner:
\vspace{-3mm}
\begin{equation}
            {LLM}(\mO_a|\mE_v,\mE_t)=\prod^l_{i=1}{LLM}(y_i|\mE_v,\mE_t,y_{<i})
\vspace{-1mm}
\label{eq1}
\end{equation}
where $l$ denotes the length of $\mO_a$. The parameter sizes of large language models (LLMs) range from 3 billion to tens of billions. Commonly used open-source LLMs include the Llama series~\citep{touvron2023llama,llama2}, Qwen~\citep{qwen}, InternLM~\citep{internlm}, MiniCPM~\citep{hu2024minicpm}, ChatGLM~\citep{chatglm}. 

MLLMs’ foundational components have greatly benefited from rapid advancements in LLM technology. Flamingo~\citep{alayrac2022flamingo} pioneered connecting pre-trained visual encoders to LLMs, demonstrating strong generalization across visual-language tasks. Following ChatGPT~\citep{achiam2023gpt}, numerous open-source models based on pre-trained LLMs, such as LLaMA series~\citep{touvron2023llama, llama2, llama3}, have been proposed~\citep{li2023blip2, huang2023kosmos1, zhu2023minigpt, liu2023llava}. Subsequent efforts, including Qwen-VL~\citep{qwenvl}, InternVL~\citep{chen2024internvl}, and CogVLMV2~\citep{CogVLM2}, have enhanced MLLMs by improving high-resolution inputs and scaling training data. Besides, smaller MLLMs like Mini-Gemini~\citep{li2024minigemini}, MobileVLM~\citep{chu2024mobilevlm2}, and MiniCPM-V~\citep{yao2024minicpmv} have emerged. Despite those advancements, the large parameter sizes of MLLMs lead to high computational costs, yet dedicated quantization methods to reduce memory usage and accelerate inference still remain underexplored.

\vspace{-3mm}
\subsection{PTQ for LLMs/MLLMs}

PTQ~\citep{zhou2024lidar, jiang2024ptq4ris, yu2025q} serves as a potential strategy for model compression. By
converting the high-precision variables of pre-trained models into low-bit integers, it achieves memory reduction and inference speed acceleration. For uniform quantization, given a floating-point (FP) tensor $x$ (weights or activations), it can be uniformly quantized to $b$-bits in signed quantization as follows:
\vspace{-1mm}
\begin{align}\label{eq:quant}
    \hat{\mathbf{x}} = \mathbf{Q}_U(\mathbf{x},b) =(clamp(\lfloor \frac{\mathbf{x}}{s} \rceil+z, q_{min}, q_{max}) - z) \cdot s
    % x_{int} = clamp(\lfloor \frac{x}{s} \rceil+z,q_{min},q_{max}),
% \vspace{-2mm}
\end{align} 
where $s=\frac{\max(|\mathbf{x}|)}{2^{b-1}-1}$ is the scale factor, $\lfloor \cdot \rceil$ is the rounding-to-nearest operator, and the function $clamp(\cdot)$ clips values outside the integer range $\left[q_{min}, q_{max} \right]$. $z$ is zero-point. $s$ denotes the quantization scale factor, which reflects the proportional relationship between FP values and integers. $\left[q_{min}, q_{max} \right]$ is the quantization range determined by the bit-width $b$. Generally, when we quantize the network's weight with 4-bit and activations with 8-bit, called it as W4A8. We can calculate $s$ offline using the activations from calibration samples, known as \textbf{static quantization}. We can also use the runtime statistics of activations to get $s$, referred to as \textbf{dynamic quantization}.  More details are in Appendix A. 12. %\ref{quant_granularity}.

Existing post-training quantization (PTQ) methods for LLMs are categorized into weight-only and weight-activation quantization~\citep{yuan2024llm,li2023fptq,li2024norm,yue2024wkvquant,hu2024llm,hu2025ostquant,chenmoequant,xumambaquant,yuerwkvquant}. Weight-only methods like GPTQ~\citep{frantar2022gptq}, QuIP~\citep{chee2024quip}, and AWQ~\citep{lin2023awq} achieve high compression rates but offer limited inference acceleration. In contrast, weight-activation quantization methods~\citep{xiao2022smoothquant, wei2022outlier, yuan2023asvd, zhang2024qqq} quantize both weights and activations, improving memory usage and latency. The main challenge is activation outliers causing quantization errors. Techniques like SmoothQuant~\citep{xiao2022smoothquant} shift quantization difficulty from activations to weights, while OmniQuant~\citep{2023omniquant} optimizes performance by training quantization parameters. SliceGPT~\citep{ashkboos2024slicegpt} reduces memory demands by designing a Pre-LN + Rotate Scheme for LLMs sparsification based on computational invariance. Recently, Quarot~\citep{ashkboos2024quarot} introduced applying \emph{offline} Hadamard transforms and a \emph{partially online} Hadamard transform to eliminate outliers, achieving state-of-the-art quantization results on LLMs. However, this solution is not applicable to MLLMs due to inherent modality differences. Building on the above studies, research on PTQ specifically tailored for MLLMs remains notably scarce. Existing studies, such as Q-VLM~\citep{qvlm}, propose a block-wise quantization framework that optimizes cross-layer dependencies via entropy-guided partitioning. MBQ~\citep{li2024mbq} uses gradients of the supervised fine-tuning loss with respect to vision and language tokens as sensitivity indicators to balance reconstruction loss during calibration. Despite efforts, existing methods still employ per-token dynamic quantization for activation, which inherently introduces a significant computational overhead than per-tensor static quantization discussed earlier. In contrast, our MQuant framework addresses these challenges by extending full static quantization to both weights and activations, specifically designed to unleash the inference potential of MLLMs while overcoming the modality-specific limitations faced by LLM-focused PTQ methods, which are not directly applicable to MLLMs due to significant modality differences.

% \vspace{-2mm}
\section{Method}\label{sec: method}

In this section, we present \emph{MQuant}, a post-training quantization solution specifically designed for MLLMs. In Sec. \ref{UniToekn}, we describe proposed modality-specific static quantization (MSQ) and attention-invariant flexible switching (AIFS) mechanisms. In Sec. \ref{sec:fht-outlier}, we identify the weight outliers caused by the online Hadamard rotations and state Rotation Magnitude Suppression (RMS). We provide the detailed MQuant algorithm for FP MLLMs in Appendix~\ref{Mquant} Algorithm \ref{alg:MQuant}.

% \vspace{-2mm}
\subsection{Modality-Specific Static Quantization and Attention-Invariant Flexible Switching}
\label{UniToekn}

We address this issue with a two-part method:
(1) \textbf{Modality-Specific Static Quantization (MSQ)} and
(2) \textbf{Attention-Invariant Flexible Switching (AIFS)}. MSQ applies different static scaling factors for visual and textual tokens, while AIFS reorders their positions to avoid irregular data slicing. MSQ+AIFS retains accuracy while eliminating the expensive token-wise scale online computation during inference. This approach achieves efficiency gains by leveraging per-tensor static quantization, especially for high-resolution inputs.

\begin{figure}[h]
    \centering
    \includegraphics[width=0.99\linewidth]{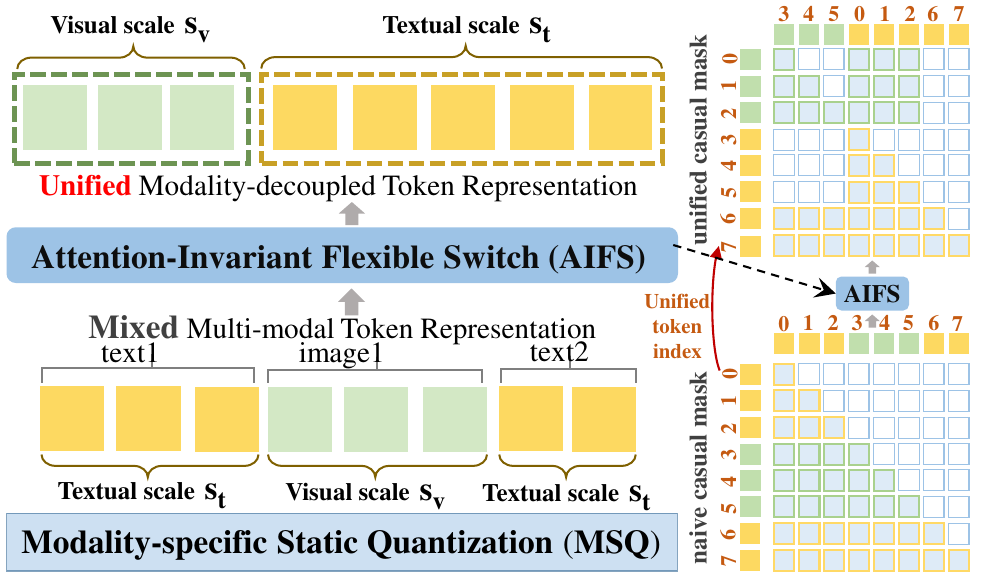}
    \vspace{-1.5mm}
    \caption{Overview of Modality-Specific Static Quantization (MSQ) and Attention-Invariant Flexible Switching (AIFS). AIFS reorders tokens so that all visual tokens appear first, then the textual tokens, while adjusting the causal mask to preserve the original model logic.}
    \label{fig:AIFS}
    \vspace{-4mm}
\end{figure}

%-------------------------------------
\vspace{-2mm}
\label{MSQ}
\noindent\paragraph{\textbf{Modality-Specific Static Quantization (MSQ)}} Let $E$ be an input sequence of length $L$ that intermixes textual and visual tokens. Denote $E = \{\;e^t_1, \dots, e^t_{m-1},\, e^v_m, \dots, e^v_n,\, e^t_{n+1}, \dots, e^t_L\}$, where $m,n$ specify the visual token segment. We observe that visual tokens often have larger activation magnitudes, which can overshadow textual features. To handle these differences, we apply two distinct sets of \emph{static per-tensor} quantization parameters:  
\begin{equation}
\label{eq:ms-ptq}
    E \;=\;
    \underbrace{(e^t_1, \dots, e^t_{m-1})}_{\textstyle \text{text scale } s_{t}}
    \;\;\underbrace{(e^v_m, \dots, e^v_n)}_{\textstyle \text{visual scale } s_{v}}
    \;\;\underbrace{(e^t_{n+1}, \dots, e^t_L)}_{\textstyle \text{text scale } s_{t}}.
\end{equation}
The scaling factors $s_{v}$ and $s_{t}$ correspond to visual and textual tokens, respectively. Our innovation lies in performing \emph{single-shot calibration} of these factors before inference, thereby eliminating per-token computational overhead. This per-tensor static quantization strategy provides three distinct advantages: \textbf{(1)} complete avoidance of dynamic scale updates, \textbf{(2)} hardware-friendly implementation benefiting from simplified quantization kernels, and \textbf{(3)} prevention of distribution saturation where high-magnitude visual activations could overwhelm the narrower dynamic range of textual tokens.

\vspace{-2mm}
\label{AIFS}
\noindent\paragraph{\textbf{Attention-Invariant Flexible Switching (AIFS)}} Despite the benefits of MSQ, it complicates data handling if textual and visual tokens remain interleaved in $E$. Na\"ive slicing and concatenating at runtime can increase memory traffic and reduce efficiency for \texttt{GEMM}-based layers like QK and FC. To overcome this, we propose \emph{Attention-Invariant Flexible Switching (AIFS)}. Figure~\ref{fig:AIFS} illustrates the idea: 
we \emph{reorder} the sequence so that all visual tokens appear first, followed by textual tokens. We then alter the causal mask to preserve the original auto-regressive relationships. Causal attention~\citep{vaswani2017attention} ensures each token can only attend to earlier tokens. Formally, for a naive sequence $E$, the attention matrix is  
\begin{equation}
    \begin{aligned}
    \mathbf{A} = \mathrm{Softmax}\!\bigl(\tfrac{\mathbf{Q}\mathbf{K}^\top}{\sqrt{D}} + M_{i,j}\bigr), 
    \quad
    M_{i,j} =
    \begin{cases}
    0, & \text{if } j \leq i,\\
    -\infty, & \text{otherwise}.
    \end{cases}
    \end{aligned}
\label{eq:casual_mask_mt}
\end{equation}
With AIFS mechanism, we can reorder $E$ and obtain an unified causal mask $M^u_{i,j}$, the new $E^{u} = \{\;e^v_m, \dots, e^v_n,\; e^t_1,\dots,e^t_{m-1},\, \dots, e^t_{n+1},$ $\dots, e^t_L\}$, and $M^u_{i,j}$ could be formulated as:

\begin{equation}
    \begin{aligned}
    M^{u}_{i,j} = 
\begin{cases} 
0 & \text{if one of the following conditions is met:} \\
& ( i \leq (n-m), j \leq i \text{ or } (n-m) < j \leq n) \\
&\text{or } ((n-m)<i \leq n, (n-m) < j \leq i) \\
&\text{or } (i > n, j \leq i) \\
-\infty &\text{otherwise} 
\end{cases}   
\end{aligned}
\label{eq:casual_unified}
\end{equation}

The position embeddings are also shifted accordingly (see Appendix~\ref{rope}), preserving \emph{numerical equivalence of attention scores} with the original sequence. This maintains auto-regressive consistency while enabling streamlined memory access: all visual/textual tokens can now be processed through respective static-scale multiplications. 
\vspace{-2mm}
\paragraph{\textbf{Seamless Integration with FlashAttention}} MSQ and AIFS, which optimizes MLLMs through optimizing on input token efficiency, while FlashAttention~\cite{dao2022flashattention} accelerates attention computation via reduced memory access and tiling. These approaches are conceptually orthogonal, suggesting seamless integration with additive benefits without inherent conflicts. Specifically, at the beginning of inference, MSQ+AIFS reorders the intermixed multimodal tokens and passes the original visual modality start/end indices ($m$ and $n$) to FlashAttention to ensure that irrelevant positions are correctly masked. Detailed results are in Table ~\ref{table:flash}.

\vspace{-2mm}
\noindent \paragraph{\textbf{Efficiency and Practical Benefits}}
Our experiments in Section~\ref{sec:exp} show that MSQ + AIFS delivers three main advantages: \raisebox{-0.5pt}{\ding[1.1]{182\relax}} \emph{High Compatibility and Strong Theoretical Equivalence:} The causal-mask transformation ensures that the model’s output is numerically the same as if tokens were not reordered. This reordering can be integrated into existing MLLM implementations without major code changes. \raisebox{-0.5pt}{\ding[1.1]{183\relax}} \emph{Reduced Inference Latency:} We replace per-token dynamic quantization with a static per-modality approach. This removes the need to recalculate scales per token, cutting runtime overhead and boosting throughput. \raisebox{-0.5pt}{\ding[1.1]{184\relax}} \emph{Enhanced Memory Efficiency:} Sequential processing of visual (via $s_{v}$) and textual ($s_{t}$) tokens with dedicated scale factors minimizes memory fragmentation, yielding up to 24.7\% speedups and 152.9\% memory efficiency gains (Table~\ref{tab:ablation2}) through eliminated padding/slicing operations.

In summary, \emph{MSQ + AIFS} offers a straightforward and effective way to handle the unique challenges of multi-modal token distributions. It is valuable on edge devices or other constrained platforms where dynamic quantization is hard to implement at scale, yet per-tensor static quantization is hardware-friendly and fast~\cite{2024_mobilequant}. More discussions is in Appendix~\ref{aifs_msq_pros}

\vspace{-2mm}

\subsection{FHT-induced Weight Outlier Mitigation: Rotation Magnitude Suppression}
\label{sec:fht-outlier}

% \paragraph{Background and Motivation.}
QuIP ~\citep{chee2024quip} and Quip\#~\citep{tseng2024quip+} formalized \emph{incoherence} to measure the difficulty of quantization. A lower incoherence indicates a simpler quantization scenario. For a weight matrix\(W \in \mathbb{R}^{m\times n}\)  with singular vectors \(e_i\), \(e_j\) and  Frobenius norm \(\|W_{\ell_2}\|_F\). the \emph{incoherence coefficient} \(\mu\) is defined as the minimal constant satisfying:
\begin{equation}
\vspace{-1mm}
\label{eq:incoherence}
\begin{aligned}
|W_{ij}| \;=\; \bigl|\,e_i^\top W\,e_j\bigr|\;\;\le\;\;\mu\,\frac{\|W\|_F}{\sqrt{mn}},
\end{aligned}
% \vspace{-1mm}
\end{equation}
Generally, the smaller \(\mu\) is, the easier it is to quantize \(W\). They also showed that applying a Hadamard transform to both weights and activations can effectively reduce \(\mu\). However, Multi-modal LLMs (MLLMs) combine an LLM component with a \emph{visual encoder}, which often relies on LayerNorm. Quarot’s partial online transform cannot be applied directly to all such norms. Inspired by SliceGPT~\citep{ashkboos2024slicegpt}, we convert visual encoders’ LayerNorms into RMSNorms. More
details can be found in the Appendix \ref{pre-LN2RMSN}. This change makes Quarot-like Hadamard transforms applicable to MLLMs. Yet, as Table~\ref{table:main_results} demonstrates, Quarot still underperforms on many MLLM tasks. Quip\#~\citep{tseng2024quip+} also proved that \emph{random Hadamard transforms} (RHT) can reduce incoherence for Hessians and weights, but they did not analyze \emph{online} (fast) Hadamard transforms (FHT). In Quarot, FHT is crucial for low-bit quantization (e.g., 4-bit) as shown in Appendix ~\ref{sec:online-hadamard}. We investigate why FHT can yield fresh outliers in MLLMs and propose a method to suppress them. Besides, due to the difference of MLLMs architecture, we also discuss the rotation-based implementation details before RMS in Appendix ~\ref{section:background-norm}.

\vspace{-1mm}
\begin{figure}[ht]
\vspace{-2mm}
    \centering    
    \includegraphics[width=0.99\linewidth]{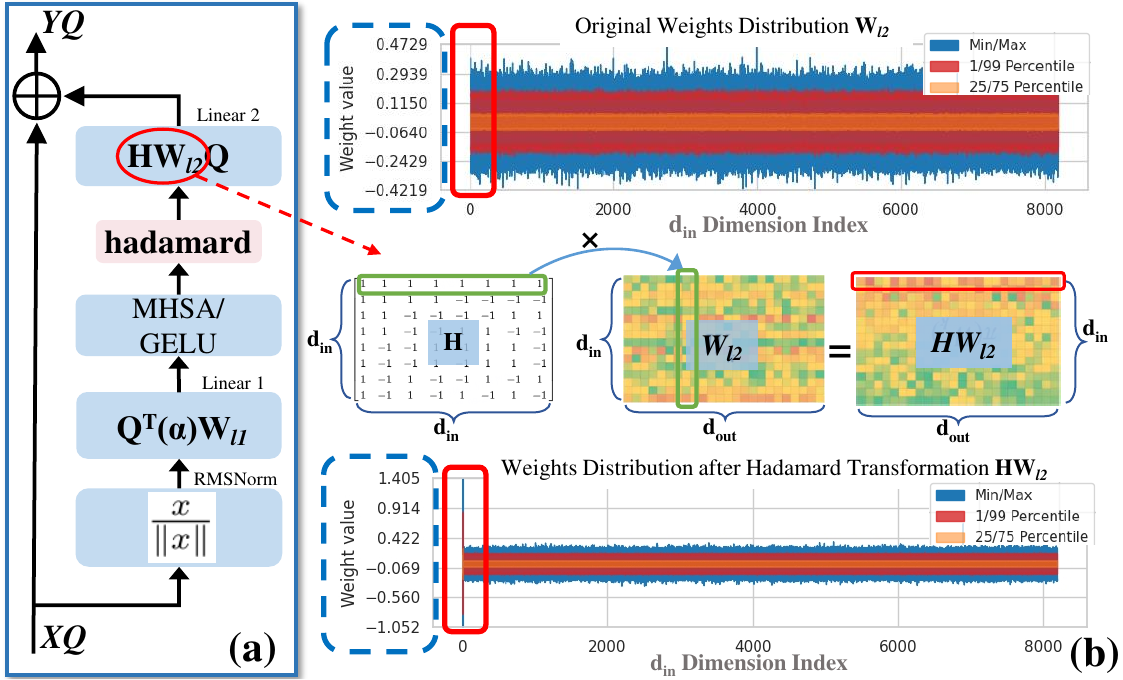}
    \vspace{-2mm}
    \caption{\textbf{(a)} The pipeline of Quarot, showing offline and partially online Fast Hadamard transforms. \textbf{(b)} Weight matrix where applying online FHT produces outliers in the first row.}
    \label{quarot}
    \vspace{-3mm}
\end{figure}

\noindent\paragraph{\textbf{Channel-Mean Outliers in Online FHT}}
Following (Eq.~\ref{eq:incoherence}), let \(W_{\ell_2}\in\mathbb{R}^{n\times m}\) be a weight matrix. A Hadamard matrix \(H \in \{-1,1\}^{n\times n}\) with orthonormal rows preserves this norm, i.e., $\|\,H\,W_{\ell_2}\|_F \\ = \|W_{\ell_2}\|_F$. Suppose \(\mu_{\ell_2}\) is the incoherence coefficient for \(W_{\ell_2}\), satisfying 
\(\max(|W_{\ell_2}|) = \mu_{\ell_2}\,\frac{\|W_{\ell_2}\|_F}{\sqrt{mn}}\).
Let \(HW_{\ell_2}\) be the transformed weight, and let its incoherence coefficient be \(\mu_{H\ell_2}\). Then we have:
\vspace{-2mm}
\begin{align}\label{eq:scale_l2}
\frac{\mu_{H\ell_2}}{\mu_{\ell_2}}
\;=\;
\frac{\max\bigl|\,H\,W_{\ell_2}\bigr|}{\max\bigl|\,W_{\ell_2}\bigr|}
\,.
\end{align}
For many Hadamard transforms, the first row (and first column) contain identical \(\tfrac{1}{\sqrt{n}}\) entries, while other rows sum to zero. Hence, the first channel after transformation is
\begin{align}
(H W_{\ell_2})_{0j}
\;=\;
\sqrt{n}\,\mathrm{mean}\bigl(w_{:,j}\bigr).
\end{align}
If the mean is large, the first element in that row can exceed the original maximum and thus raise \(\mu_{H\ell_2}\). Concretely, when
\begin{align}
\label{eq:scale_l3}
\sqrt{n}\,\mathrm{mean}\bigl(w_{:,j}\bigr) \;>\; \max_{i}\,\bigl(w_{ij}\bigr),
\end{align}
a new \emph{channel-mean outlier} appears in the first row. Figure~\ref{quarot}(b) shows such an occurrence in an MLLM. This issue arises especially in Quarot’s \emph{online} (partial) FHT step, which applies Hadamard rotations per forward pass rather than strictly offline. The detailed approve in Appendix ~\ref{weight_outliers}.

\begin{table}[htbp]
\vspace{-1mm}
\caption{Compliance Ratio with Eq. 9 Across Model Components}
\vspace{-2mm}
\label{tab:ratio_rms}
\centering
\resizebox{1.0\linewidth}{!}{
\begin{tabular}{c|cc|c|c}
\toprule
\multirow{2}{*}{\textbf{Model}} & \multirow{2}{*}{\textbf{Part}} & \textbf{All Blocks} & \textbf{Meet Eq. 9} & \multirow{2}{*}{\textbf{Ratio}}\\
& &\textbf{Number} & \textbf{Number} & \\
\midrule
\multirow{2}{*}{\textbf{Internvl2-8B}}  & Visual & 24 & 24 & 100\% \\
& LLM  & 32 & 4  & 12.5\% \\
\hline
\multirow{2}{*}{\textbf{Qwenvl-9.6B}}     & Visual & 48 & 48 & 100\% \\
& LLM  & 32 & 13 & 40.6\% \\
\hline
\multirow{2}{*}{\textbf{MiniCPM-V-2.6-8B}}      & Visual & 27 & 27 & 100\% \\
& LLM    & 28 & 3  & 10.7\% \\
\hline
\multirow{2}{*}{\textbf{GLM-4V-9B}}        & Visual & 63 & 63 & 100\%\\
 & LLM   & 40 & 6  & 15.0\% \\
\hline
\multirow{2}{*}{\textbf{Qwen2vl-7B}}    & Visual & 32 & 32 & 100\% \\
& LLM            & 28 & 1  & 3.5\% \\
\bottomrule
\end{tabular}
}
\vspace{-4mm}
\end{table}

\noindent\paragraph{\textbf{Prevalence of Eq.~\ref{eq:scale_l3} Compliance}} We conduct systematic block-level verification across five state-of-the-art MLLMs (InternVL2-8B, QwenVL-9.6B, MiniCPM-V-2.6-8B, GLM-4V-9B, and Qwen2VL-7B) to quantify the occurrence of Fast Hadamard Transform (FHT)-induced weight outliers. As evidenced in table ~\ref{tab:ratio_rms}, our analysis reveals two critical findings: \textbf{(1) Universal Presence in Vision Encoders}: All visual $down_{proj}$ layers (100\%) exhibit FHT-induced weight outliers that satisfy Eq.~\ref{eq:scale_l3}, as conceptually illustrated in Fig.~\ref{quarot}(b). \textbf{(2) Variable Manifestation in LLMs}: The phenomenon appears in only 3\%–40\% of LLM $down_{proj}$ layers across different architectures. This empirical validation confirms above theoretical soundness and the critical need for specialized handling of FHT-generated weight outliers, particularly in vision components where their universal presence fundamentally impacts model quantization.

\begin{figure}[ht]
\vspace{-2mm}
    \centering    
    \includegraphics[width=\linewidth]{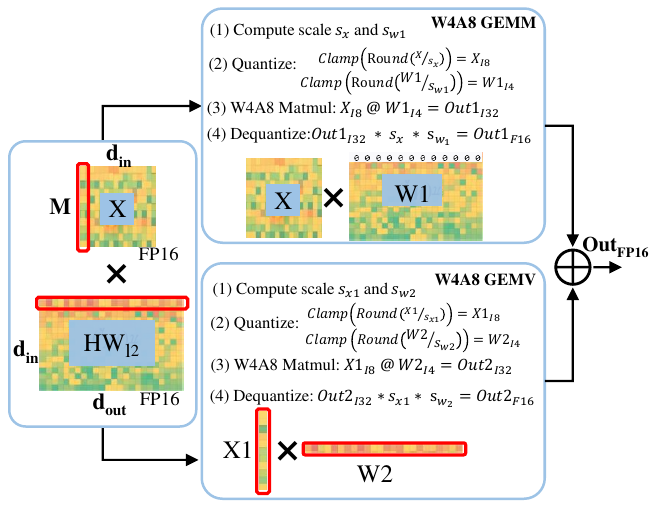}
    \vspace{-2mm}
    \caption{An overview of our proposed \textbf{RMS}. We separate outlier-prone channels into a dedicated \texttt{GEMV} path, zero their row in the main kernel, and then merge the results.}
    \label{oswq}
    \vspace{-4mm}
\end{figure}

\noindent\paragraph{\textbf{Rotation Mitigation Scheme (RMS)}}
To this end, we propose \emph{RMS} to handle these new outliers with minimal overhead. We first identify whether a channel meets (Eq.~\ref{eq:scale_l3}). If it does, we: \raisebox{-0.5pt}{\ding[1.1]{182\relax}} \textbf{Split} the problematic channel from the main \texttt{GEMM} kernel and process it using a separate \texttt{GEMV}. \raisebox{-0.5pt}{\ding[1.1]{183\relax}} \textbf{Zero out} that row in $HW_{\ell_2}$ so that the main kernel does not double-count it. \raisebox{-0.5pt}{\ding[1.1]{184\relax}} \textbf{Add} the partial output (from the split \texttt{GEMV}) back to the main path before the activation.
Figure~\ref{oswq} shows the workflow. This targeted split ensures large-mean channels do not trigger extreme first-row values during the per-forward-pass FHT. The added cost is small, since only channels meeting Eq.~\ref{eq:scale_l3} require this procedure. Detailed RMS algorithm in Appendix~\ref{RMS_algo} outlines how RMS integrates with Quarot’s FHT. Overall, RMS addresses a key shortcoming of online Hadamard transforms by suppressing large-mean channels.

% \vspace{-2mm}
\noindent\paragraph{\textbf{Why Not Subtract Channel Means Directly?}}
A straightforward idea is to separate out each channel's mean before applying $H$ and then re-inject it afterward. However, this leads to two major issues: 1. The separated means are still subject to Hadamard rotation, creating a new linear transformation where the first channel again becomes large. 2. Splitting and then re-injecting the means effectively doubles the linear operations, significantly increasing the computational cost. In contrast, our RMS approach modifies only the row triggering the outlier condition and does not require additional linear layers, thereby offering a more efficient and effective resolution. 

\vspace{-2mm}
\section{Experiments}\label{sec:exp}
\vspace{-1mm}
\label{exp}

\begin{table}[t]
\caption{Comprehensive quantization results of different MLLMs across various multimodal reasoning datasets.}
% \vspace{+2mm}
\label{table:main_results}
\centering
\Huge
\resizebox{\linewidth}{!}{
\begin{tabular}{c|c|cc|cccc}
\toprule
\multirow{2}{*}{\textbf{MLLMs}} &  \multirow{2}{*}{\textbf{Method}}& \multicolumn{2}{|c|}{\textbf{Bits Setting}} & \multirow{2}{*}{\textbf{T.VQA$\uparrow$}} & \multirow{2}{*}{\textbf{D.VQA$\uparrow$}} & \multirow{2}{*}{\textbf{OCRB.$\uparrow$}} & \multirow{2}{*}{\textbf{MME$\uparrow$}} \\
& & Visual & LLM & \\
\toprule
  \rowcolor{gray!12}
\cellcolor{white}& - & BF16 & BF16 & 77.65 & 90.97 & 794 & 2209 \\ 
  & RTN & \multirow{4}{*}{W8A8} &\multirow{4}{*}{W4A8} & 52.02 & 59.04 &542 &1528 \\
  & SQ &  & & 59.88 &59.75 &544 &1540 \\
  & Quarot &  & & 73.34 & 84.07 & 715 & 2067 \\
  \rowcolor{gray!25}
  \cellcolor{white}{InternVL2} & \textbf{MQuant} &  & & \textbf{77.49} & \textbf{90.27} & \textbf{785} & \textbf{2156} \\
  \cline{2-8}
  \cellcolor{white}{-8B} & RTN &\multirow{4}{*}{W4A8} &\multirow{4}{*}{W4A8}& 40.06 &31.58&302 &1482 \\
  & SQ&  & & 46.48 &31.21 &310 &1540 \\
  & Quarot& & &49.10 &33.62 &361 &1941 \\
  \rowcolor{gray!25}
  \cellcolor{white} & \textbf{MQuant} && & \textbf{76.62} & \textbf{88.42} &\textbf{725} &\textbf{2155}\\ 
  \midrule
  \rowcolor{gray!12}
  \cellcolor{white} & - & BF16 & BF16  & 61.40  &60.36 & 493 & 1834 \\
  & RTN & \multirow{4}{*}{W8A8} &\multirow{4}{*}{W4A8} &  0.45 & 0.03 & 189 & 625 \\
  & SQ &  & & 7.45 &7.70 & 160 &797 \\
  & Quarot & &  & 45.32 &42.44 &286 &940 \\
  \rowcolor{gray!25}
  \cellcolor{white}{{{Qwen-VL}}} & \textbf{MQuant} &  & & \textbf{61.16} &\textbf{59.31} &\textbf{483} & \textbf{1691} \\ \cline{2-8}
  \cellcolor{white}{{{-Chat-9.6B}}}& RTN & \multirow{4}{*}{W4A8} &\multirow{4}{*}{W4A8} & 1.02 &0.02 &193 &585 \\
  & SQ &  & &  8.59 &4.28 &188 &921 \\
  & Quarot &  & &  46.77 &37.35 & 289 & 1091 \\
  \rowcolor{gray!25}
  \cellcolor{white} &\textbf{MQuant} &  & & \textbf{60.50} &\textbf{58.72} &\textbf{473} &\textbf{1713} \\
\midrule
\rowcolor{gray!12}
\cellcolor{white}& - & BF16 & BF16 & 79.10 &89.18 &847 &2248 \\
& RTN &\multirow{4}{*}{W8A8} &\multirow{4}{*}{W4A8}& 61.00 & 65.16 & 332 & 1300 \\
  & SQ &  & &  62.40 & 65.76 & 424 & 1510 \\
  & Quarot  &  & & 73.71 &80.04 &736 &1850 \\
  \rowcolor{gray!25}
  \cellcolor{white}{MiniCPM-V} & \textbf{MQuant} &  & & \textbf{80.41} &\textbf{89.15} &\textbf{844} &\textbf{2244} \\ \cline{2-8}
  \cellcolor{white}{2.6-8B}& RTN & \multirow{4}{*}{W4A8} &\multirow{4}{*}{W4A8} &60.70 &62.23 &351 &1404 \\
  & SQ &  & &  65.67 &60.02 &455 &1491\\
  & Quarot &  & & 68.96 &79.63 &685 &1734 \\
  \rowcolor{gray!25}
  \cellcolor{white} & \textbf{MQuant} &  & & \textbf{81.14} & \textbf{89.75} &\textbf{839} &\textbf{2189} \\
\midrule
\rowcolor{gray!12}
\cellcolor{white}& - & BF16 & BF16 & 82.82 &81.16 &782 &2153 \\
& RTN &\multirow{4}{*}{W8A8} &\multirow{4}{*}{W4A8}&  7.05 &3.70 &0.00 &140 \\
  & SQ &  & &  9.05 &4.10 &0.00 &148 \\
  & Quarot  &  & & 82.00 &80.17 &782 &2115 \\
  \rowcolor{gray!25}
  \cellcolor{white}{GLM-4V} & \textbf{MQuant} &  & & \textbf{82.06} &\textbf{80.53} &\textbf{782} &\textbf{2164} \\ \cline{2-8}
  \cellcolor{white}{-9B}& RTN & \multirow{4}{*}{W4A8} &\multirow{4}{*}{W4A8} &7.61 &3.60 &0.00 &163 \\
  & SQ &  & &  9.85 &4.40 &0.00 &188 \\
  & Quarot &  & & 64.16 &45.52 &516 &2048 \\
  \rowcolor{gray!25}
  \cellcolor{white} & \textbf{MQuant} &  & & \textbf{81.58} & \textbf{79.67} &\textbf{754} &\textbf{2120} \\
\midrule
\rowcolor{gray!12}
\cellcolor{white}& - & BF16 & BF16 & 84.43 & 93.87 &842 &2319 \\
& RTN &\multirow{4}{*}{W8A8} &\multirow{4}{*}{W4A8}& 33.92&52.61&442&1298 \\
  & SQ &  & &  49.11 &53.97 &444 &1500 \\
  & Quarot  &  & & 79.36 &89.57 &754 &2045 \\
  \rowcolor{gray!25}
  \cellcolor{white}{Qwen2-VL} & \textbf{MQuant} &  & & \textbf{84.43} &\textbf{93.61} &\textbf{830} &\textbf{2269} \\ \cline{2-8}
  \cellcolor{white}{-7B}& RTN & \multirow{4}{*}{W4A8} &\multirow{4}{*}{W4A8} &40.20 &38.82 &422 &1082 \\
  & SQ &  & &  46.25 &52.36 &411 &1535 \\
  & Quarot &  & & 71.44 & 83.96 & 670 & 1911 \\
  \rowcolor{gray!25}
  \cellcolor{white} & \textbf{MQuant} &  & & \textbf{84.32} & \textbf{93.58} &\textbf{824} &\textbf{2255} \\
\midrule
\rowcolor{gray!12}
\cellcolor{white}& - & BF16 & BF16 & 85.48 &95.95 & 883 &2479\\
& RTN &\multirow{4}{*}{W8A8} &\multirow{4}{*}{W4A8}&37.21 &40.94 &426 &1444 \\
  & SQ &  & &  50.33  & 55.41 & 480 & 1601\\
  & Quarot  &  & & 80.03 & 91.21 & 781 & 2299 \\
  \rowcolor{gray!25}
  \cellcolor{white}{Qwen2-VL} & \textbf{MQuant} &  & & \textbf{85.48} &\textbf{95.90} &\textbf{880} &\textbf{2469} \\ \cline{2-8}
  \cellcolor{white}{-72B}& RTN & \multirow{4}{*}{W4A8} &\multirow{4}{*}{W4A8} &28.21 &25.94 &426 &1137 \\
  & SQ &  & &  47.11 & 54.95 & 413 & 1586 \\
  & Quarot &  & & 71.86 & 86.11 & 701 & 2264 \\
  \rowcolor{gray!25}
  \cellcolor{white} & \textbf{MQuant} &  & & \textbf{85.03} & \textbf{95.49} &\textbf{868} &\textbf{2471} \\
\bottomrule
\end{tabular}
}
\vspace{-7mm}
\end{table}

% \vspace{-2mm}

\vspace{-1mm}
\noindent\paragraph{\textbf{Models and Datasets}} We evaluate our \emph{MQUANT} on five MLLMs: InternVL2-8B~\citep{internvl15}, Qwen-VL-Chat-9.6B~\citep{qwenvl}, MiniCPM-V 2.6-8B~\citep{yao2024minicpmv}, Qwen2-VL-7B~\citep{Qwen2VL}, and GLM-4V-9B~\citep{CogVLM2}. Evaluations are conducted on four benchmarks covering OCR and general question answering: TextVQA~\citep{singh2019textvqa}, DocVQA~\citep{mathew2021docvqa}, OCRBench~\citep{liu2023ocrbench}, and MME~\citep{fu2023mme}, which assesses perception and cognition across 14 tasks. These MLLMs' details are in Appendix~\ref{MLLMs_comparison}.

\vspace{-1mm}
\noindent\paragraph{\textbf{Baselines and Implementation Details}} We test W8A8 and W4A8 quantization settings for both visual encoders and LLMs, comparing RTN, SmoothQuant~\citep{xiao2022smoothquant}, and Quarot~\citep{ashkboos2024quarot}. Notably, we apply static per-tensor activation quantization for both components, unlike the dynamic per-token quantization typically used in existing MLLMs. The calibration dataset consists of 256 randomly selected samples from the corresponding benchmark training sets~\citep{singh2019textvqa, mathew2021docvqa, liu2023ocrbench}. The batch size for latency measurement is 1.

\vspace{-1mm}
\subsection{Overall Results}
% \vspace{-1mm}

As shown in Table~\ref{table:main_results}, \emph{MQuant} can apply to the quantization of various MLLMs, and demonstrates significant improvements over several representative quantization methods. In W8A8 setting, MQuant achieves near-lossless performance to the FP models across diverse datasets and different model scales (7-72B). Notably, even in the more challenging W4A8 setting, MQuant maintains comparable performance with FP models, while other advanced quantization methods exhibit significant performance degradation. These results indicate that our MQuant provide a general and effective PTQ solution with strong compatibility for maintaining high accuracy in MLLMs under various bits settings across diverse multi-modal reasoning tasks. This suggests MQuant maintains robustness (e.g., stability to typical input or models changes) in these frameworks.

\begin{table}[htbp]
  \centering
  \caption{Comparison of existing MLLMs Quantization Methods}
  \label{tab:comp_mllm}
  \resizebox{1.0\linewidth}{!}{
  \begin{tabular}{lcccll}
    \toprule
    $\textbf{Method}$ & $\textbf{W bits}$ & $\textbf{A bits}$ & $\textbf{Scale}$             & $\textbf{GEMM kernel}$ & $\textbf{Extra ops}$         \\
    \midrule
    Q-VLM~\citep{qvlm}           & NF4           & INT4            & per-channel static       & BF16 GEMM          & dequant W\&A               \\
    QSLAW~\citep{xie2024advancing}           & INT4          & BF16            & weight-only              & BF16 GEMM          & dequant W                  \\
    MBQ~\citep{li2024mbq}              & INT4          & INT4            & per-token dynamic        & INT4 GEMM          & per token scale \\
    \rowcolor{myblue!20} MQuant (ours)    & INT4          & INT4            & per-modality static      & INT4 GEMM          & none                       \\
    \bottomrule
  \end{tabular}
  }
  \vspace{-3mm}
\end{table}

\begin{table}[htbp]
  \centering
  \caption{ScienceQA~\cite{ScienceQA} test Results on LLaVA1.5-13B}
  \label{tab:SQA}
  \resizebox{1.0\linewidth}{!}{
  \begin{tabular}{lcccccclc}
    \toprule
    $\textbf{Bits}$ & $\textbf{Method}$     & $\textbf{NAT}$ & $\textbf{SOC}$ & $\textbf{LAN}$ & $\textbf{TXT}$ & $\textbf{IMG}$ & $\textbf{NO}$ & $\textbf{Avg$\uparrow$}$ \\
    \midrule
    BF16            & -                   & 90.19      & 93.14      & 87.09      & 89.39      & 87.06      & 89.83     & 90.00                 \\
    \midrule
    w4a4          & Q-VLM~\citep{qvlm}               & 82.55      & 73.32      & 83.18      & 81.03      & 70.82      & 86.74     & 80.78                 \\
    \rowcolor{myblue!20} w4a4          & MQuant              & \textbf{87.07}      & \textbf{85.24}      & \textbf{87.05}      & \textbf{88.49}      & \textbf{85.50}      & \textbf{89.77}     & \textbf{87.18}                 \\ \midrule
    w4 only       & QSLAW~\citep{xie2024advancing}                & 83.26      & 91.79      & 80.00      & 82.80      & 81.30      & 83.07     & 83.70 \\
    \rowcolor{myblue!20} w4 only       & MQuant (g=128)       & \textbf{90.21}      & \textbf{91.95}      & \textbf{87.05}      & \textbf{89.20}      & \textbf{86.66}      & \textbf{89.94}     & \textbf{89.16} \\
    \bottomrule
  \end{tabular}
  }
  \vspace{-3mm}
\end{table}

\begin{table}[htbp]
  \centering
  \caption{Performance Comparison under Different Precision and Methods (T.VQA, D.VQA, OCRB, MME) on Qwen2VL-72B.}
  \label{tab:MBQ}
  \resizebox{1.0\linewidth}{!}{
  \begin{tabular}{lccccc}
    \toprule
    $\textbf{Bits}$ & $\textbf{Method}$ & $\textbf{T.VQA$\uparrow$}$ & $\textbf{D.VQA$\uparrow$}$ & $\textbf{OCRB.$\uparrow$}$ & $\textbf{MME$\uparrow$}$ \\
    \midrule
    BF16  & - & 85.48 & 95.95 & 883 & 2479 \\
    W4A8  & MBQ~\citep{li2024mbq}      & 83.41 & 92.30 & 854 & 2402 \\
    \rowcolor{myblue!20} W4A8  & MQuant (ours)    & \textbf{85.03} & \textbf{95.49} & \textbf{868} & \textbf{2471} \\
    \bottomrule
  \end{tabular}
  }
  \vspace{-3mm}
\end{table}

\subsection{Comparison with other MLLMs Quantization Methods}

To rigorously evaluate MQuant, we compare it against recent state-of-the-art (SOTA) quantization methods~\citep{qvlm,xie2024advancing, li2024mbq} specifically designed for MLLMs. Table~\ref{tab:comp_mllm} provides a high-level comparison of the quantization schemes. Existing methods often introduce substantial runtime overhead. For example, MBQ~\citep{li2024mbq} employs online dynamic per-token quantization; Q-VLM~\citep{qvlm} and QSLAW~\citep{xie2024advancing} require dequantization and perform GEMM in resource-intensive BF16 precision. In contrast, MQuant is the only method that achieves a fully static, per-modality quantization scheme that is directly compatible with highly efficient \textbf{INT4 GEMM kernels}. This design choice is the foundation of its superior inference speed. Besides, due to their different evaluation benchmark, we compare them on their reported results. The empirical results in Table~\ref{tab:SQA} and Table~\ref{tab:MBQ} confirm MQuant's accuracy advantage. On the ScienceQA benchmark~\cite{ScienceQA} (Table~\ref{tab:SQA}) in LLaVA1.5-13B~\cite{liu2023llava}, our W4A4 MQuant surpasses Q-VLM by a significant margin of \textbf{6.4\%} in average accuracy. In the weight-only setting, MQuant outperforms QSLAW by \textbf{5.46\%}, highlighting the effectiveness of our activation quantization strategy. Similarly, on Qwen2VL-72B\cite{Qwen2VL} (Table~\ref{tab:MBQ}), our W4A8 MQuant achieves near-lossless performance, outperforming MBQ while avoiding its costly per-token dynamic quantization. This superior performance is a direct result of our novel designs. The \textbf{MSQ} effectively handles the distributional disparity between visual and textual tokens, a challenge that other methods do not explicitly address. Furthermore, the \textbf{AIFS} scheme eliminates the need for dynamic scales without compromising model accuracy, enabling the use of fast, hardware-friendly static quantization. Together with \textbf{RMS}, these contributions allow MQuant to achieve both the highest accuracy and the fastest true INT4 inference among existing MLLM quantization methods.

%\paragraph{Multi-round case Experiments.} 

\vspace{-1.5mm}
\subsection{Ablation Study}
% \vspace{-2mm}
In this section, we select Qwen2-VL-7B~\citep{Qwen2VL}, one of the most powerful open-source MLLMs, to validate the effectiveness of our designs.

\begin{table}[h]
\vspace{-2mm}
\caption{Ablation study of proposed designs on Qwen2-VL-7B~\cite{Qwen2VL} with W4A8 setting.}
\vspace{-2mm}
\label{table:ablation}
\centering
\Huge
\resizebox{\linewidth}{!}{
\begin{tabular}{ccc|ccccc}
\toprule
\multicolumn{3}{c|}{\textbf{Methods}}& \multirow{2}{*}{\textbf{T.VQA$\uparrow$}} & \multirow{2}{*}{\textbf{D.VQA$\uparrow$}} & \multirow{2}{*}{\textbf{OCRB.$\uparrow$}} & \multirow{2}{*}{\textbf{MME$\uparrow$}} & \multirow{2}{*}{\textbf{Lat (ms) $\downarrow$}}\\
Static& AIFS + MSQ &RMS & \\
\toprule
\rowcolor{gray!12}
\multicolumn{3}{c|}{\textbf{BF16}} & 84.43 &93.87 &842 &2319 &6523\\ \hline
\cc & \xx &\xx & 71.44 & 83.96 & 670 & 1911 & 5479\\
\cc & \cc& \xx  & 78.95 & 87.55 & 721 & 2095 &5484\\
% &  \cc& \cc&  \cc &\xx& & & 82.48 & 91.91 & 803 & 2174 &\textcolor{myblue}{5452}\\
  \rowcolor{myblue!20}
\cc& \cc& \cc &\textbf{84.32} &\textbf{93.58} &\textbf{824} &\textbf{2255} &5471\\
\bottomrule
\end{tabular}
}
\vspace{-3mm}
\end{table}

\noindent\paragraph{\textbf{Ablation Study of Proposed Quantization Designs}} We perform ablations on Qwen2-VL-7B (Table~\ref{table:ablation}) to isolate the contributions of each quantization component. Beginning with a baseline that applies GPTQ + Hadamard transformations to both the LLM and vision parts with per-tensor static quantization, we progressively introduce \textbf{MSQ+AIFS} for the LLM. This significantly boosts accuracy and latency, highlighting their effectiveness on multimodal input tokens. Next, incorporating \textbf{RMS} to suppress newly arising outliers achieves near-floating-point performance, underscoring the overall robustness and efficiency of our pipeline. We also demonstrate that the design of \textbf{RMS}is also effective for LLMs in Appendix ~\ref{RMS4LLMs}.

% \vspace{-4mm}

\begin{table}[h]
\vspace{-2mm}
\caption{Ablation study about calibration dataset size.}
\vspace{-2mm}
\label{table:calib_size}
\centering
\resizebox{0.8\linewidth}{!}{
\begin{tabular}{c|cccc}
\toprule
\textbf{Calib size}& \textbf{T.VQA$\uparrow$} & \textbf{D.VQA$\uparrow$} & \textbf{OCRB.$\uparrow$} & \textbf{MME$\uparrow$} \\
\midrule
BF16 & 84.43 &93.87 &842 &2319\\
\midrule
128   &84.28 & 93.50 &820 & 2243 \\ 
\rowcolor{myblue!20}
256   &84.32 &93.58 &824  &2255 \\ 
512   &84.32  &93.57 &823 &2254\\ 
\bottomrule
\end{tabular}
}
\label{calib_size}
\vspace{-3mm}
\end{table}

\vspace{-2mm}
\noindent\paragraph{\textbf{Selection of Calibration Dataset Size}} Following ~\citep{xiao2022smoothquant,ashkboos2024quarot}, we selected a sample subset to get the statistics of activations during the calibration process. This ensures the calibration data reflects the distribution in each dataset. To assess sensitivity to calibration dataset size of \emph{MQuant}, we conducted experiments on Qwen2-VL-7B, including [128, 256, 512] samples. The Table~\ref{calib_size} shows the consistency of MQuant in W4A8 with loessless accuracy. These results, revised in supplementary, demonstrate that MQuant is largely insensitive to calibration dataset size. We believe this robustness enhances MQuant’s practical applicability. Therefore, we select 256 randomly selected samples from the corresponding benchmark training set. 

\begin{table}[h]
\vspace{-2mm}
\caption{Latency combining AIFS with FlashAttention from 100 to 8100 tokens ($Q/K/V$ shape is (1, 28, num-tokens, 128))} %(1,28,100,128)
\vspace{-2mm}
\label{table:flash}
\centering
\resizebox{\linewidth}{!}{
\begin{tabular}{c|ccccc}
\toprule
Tokens Number& 100 &900 & 2500 & 4900 &8100\\
\midrule
FlashAttn (ms) &27.3& 33.6 & 35.5 & 165.9 &271.6\\
AIFS+FlashAttn (ms) &27.3 & 34.1& 36.4& 167.5 & 272.1\\
\bottomrule
\end{tabular}
}
\vspace{-3mm}
\end{table}

\begin{table*}[t]
\caption{The accuracy and speedup of our MSQ and AIFS scheme on the linear layer during prefill stage. The input sequence is structured as "text-image-text-image-text-image-text" with an image resolution of $2240 \times 2240$ and 50 textual tokens. Latency were tested on an NVIDIA RTX 6000 Ada Generation.}
% \vspace{+2mm}
\vspace{-2mm}
\label{table:AIFS_laten}
\centering
\resizebox{\textwidth}{!}{
\begin{tabular}{l|l|cccc|c|c}
\toprule
\textbf{Activation}  & \textbf{Weight} &\textbf{T.VQA$\uparrow$} & \textbf{D.VQA$\uparrow$} & \textbf{OCRB.$\uparrow$} & \textbf{MME$\uparrow$}& \textbf{Latency $\downarrow$} (s) & \textbf{Speedup $\uparrow$} \\
\toprule
\rowcolor{gray!10}
BF16 &BF16  &  84.43 &93.87 &842 &2319&1.690 & -\\ \hline
BF16& W4-g128(AWQ)  &83.93 (-0.50) &93.13 (-0.74) &828 (-14) &2252 (-67) &\textbf{2.057} (+0.367) &\color{myred}{-17.8\% }\\ \hline
A8-per-token dyn  & \multirow{4}{*}{W4-per-channel sta}& \cellcolor{yellow!7} 84.32 (-0.11)&\cellcolor{yellow!7} 93.61 (-0.26)	&\cellcolor{yellow!7} 830 (-12)&\cellcolor{yellow!5} 2269 (-50) & \cellcolor{yellow!7} 1.253 (-0.437) &\color{mygreen1}{+34.9\%}   \\ %
A8-per-tensor sta  & &  \cellcolor{yellow!20} 40.20 {(-44.12)} & \cellcolor{yellow!20} 38.82 {(-54.79)}&\cellcolor{yellow!20} 422 {(-408)} &\cellcolor{yellow!20}1082 {(-1187)}&\cellcolor{yellow!20} \textbf{1.016} (\textbf{-0.674}) & \color{mygreen1}{\textbf{+66.3\%}}  \\
% &  & & & AIFS &1.958 (\textbf{\color{mygreen1}{$\uparrow$22.6\%}})\\
% \rowcolor{gray!25}
A8-MSQ & & 84.32 (-0.11)	&93.61 (-0.26) &830 (-12) &2269 (-50)&1.085 (-0.605) & \color{mygreen1}{+55.8\%}\\ \cline{3-8}
\rowcolor{myblue!20}
\textbf{A8-MSQ+AIFS} & & \textbf{84.32} (-0.11)	& \textbf{93.61} (-0.26) &\textbf{830} (-12)&\textbf{2269} (-50)& \underline{1.017} (-0.673) & \color{mygreen1}{\textbf{+66.2\%}}\\
\bottomrule
\end{tabular}
}
% \vspace{-3mm}
\end{table*}

\begin{table*}[h]
    \vspace{-1mm}
    \caption{Comparison of latency and memory saving with Pytorch (BF16), AWQ (W4-only) and MQuant (W4A8) on Qwen2-VL-7B. $\downarrow$ means lower values are better, $\uparrow$ means larger values are better.}
    \vspace{-4.5mm}
    \label{tab:ablation2}
    \centering
    \resizebox{\textwidth}{!}{
        \begin{tabular}{c>{\columncolor{gray!15}}c|>{\columncolor{blue!2}}c>{\columncolor{blue!2}}c|>{\columncolor{myblue!15}}c>{\columncolor{myblue!15}}c|>{\columncolor{gray!15}}c|>{\columncolor{blue!2}}c>{\columncolor{blue!2}}c|>{\columncolor{myblue!15}}c>{\columncolor{myblue!15}}c}\\
        \toprule 
        Image size & \multicolumn{5}{c|}{Latency (s)}& \multicolumn{5}{c}{Memory (G)}\\ \cline{2-11}
        
          H $\times$ W & Pytorch& AWQ$\downarrow$ & Speedup$\uparrow$ & MQuant$\downarrow$& Speedup$\uparrow$ & Pytorch & AWQ$\downarrow$ & Improve$\uparrow$ & MQuant$\downarrow$  & Improve$\uparrow$\\ \cline{2-11}
        840$^2$ & 0.261  &0.304 (+0.043) &\color{myred}{-14.14\%} &0.210 (-0.051) &\textbf{\color{mygreen1}{+24.76\%}}&16.45 &7.45 (-9.00)&\color{mygreen1}{+120.67\%}&6.50 (-9.95) &\textbf{\color{mygreen1}{+152.92\%}}\\
        1960$^2$ & 1.369  &1.598 (+0.229) &\color{myred}{-14.26\%} &1.112 (-0.257) & \textbf{\color{mygreen1}{+16.63\%}} &17.82 &8.82 (-9.00)&\color{mygreen1}{+100.00\%} &7.85 (-9.97)  &\textbf{\color{mygreen1}{+119.93\%}}\\
        3080$^2$ & 5.208  &5.872 (+0.664) &\color{myred}{-11.27\%} &4.488 (-0.720) &\textbf{\color{mygreen1}{+16.02\%}} &20.58 &11.58 (-9.00)&\color{mygreen1}{+77.60\%}&10.61 (-9.97) &\textbf{\color{mygreen1}{+96.45\%}}\\
        5600$^2$ & 8.380  &9.393 (+1.013) &\color{myred}{-10.78\%} &7.469  (-0.911)&\textbf{\color{mygreen1}{+12.19\%}} &22.22 &13.22 (-9.00)&\color{mygreen1}{+57.54\%}&12.25 (-9.97) &\textbf{\color{mygreen1}{+61.65\%}}\\
            \hline
            \end{tabular}  
        }
\label{tab:speedup}
\vspace{-1mm}
\end{table*}

\vspace{-2mm}
\noindent\paragraph{\textbf{Integrate AIFS with Flash Attention}} As evidenced in Table~\ref{table:flash}, we integrated the causal attention mechanism of AIFS into FlashAttention and tested it on an RTX 6000, observing that its speed is nearly identical to that of standard causal attention. The results demonstrates seamless integration MBQ+AIFS with Flash Attention, exhibiting negligible latency overhead (<0.3\% across all tested token lengths). The minimal performance gap (maximum 1.6 ms) stems from two key design choices: (1) AIFS's token reordering operates exclusively in the preprocessing phase without interfering with core attention computation, and (2) our metadata propagation mechanism (visual token indices m/n) preserves Flash Attention's native memory access patterns. Notably, the 0.96\% latency increase at 4,900 tokens validates strong scalability for high-resolution inputs. This highlight the hardware compliance advantages of MBQ and AIFS designs: CUDA-aligned token layouts enable binary compatibility with FlashAttention kernels, which is particularly effective for real-time processing of high-resolution multimodal inputs.

\begin{figure}[h]
    \centering    
    \includegraphics[width=\linewidth]{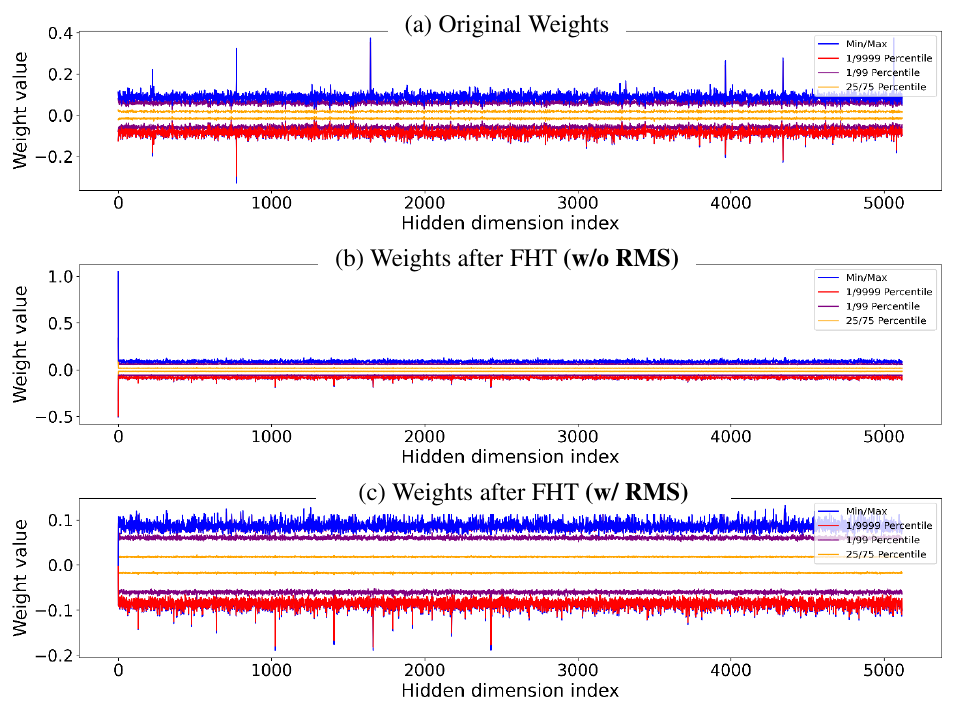}
    \vspace{-5mm}
    \caption{Illustration the weight distributions for the down-proj layer (block 21) in Qwen2VL-7B's visual encoder under three conditions: (a) original weight, (b) weights after FHT, and (c) weights after FHT with our RMS. More resutls refer to Appendix~\ref{weight_distritbuion}} 
    \label{fig:RMS_weight_ablation}
\vspace{-3mm}
\end{figure}

\paragraph{\textbf{Weight Distribution with Rotation Magnitude Suppression (RMS)}} As shown in Fig. ~\ref{fig:RMS_weight_ablation}, we visualizes the magnitude of weights outliers without (w/o) and with (w/) RMS. The weight distribution reveals that FHT amplification increases weight magnitudes, while RMS effectively reduces these magnitudes by a factor of 1.0 to 0.1, significantly improving quantization stability (Other distribution are in Appendix~\ref{weight_distritbuion}). This is also consistent with our ablation results, shown that RMS significantly enhances performance for both MLLMs (Table ~\ref{table:ablation}).  Taking Qwen2-VL as example, the vision encoder's down-projection layer (3584×1280, $C\_in * C\_out$) and LLM's corresponding layer (18944×3584) demonstrate the efficiency of our approach. The RMS implementation converts a W4A8 GEMM operation into a combined W4A8 GEMM + W4A8 GEMV computation, where the GEMV is selectively applied only to affected channels. This introduces negligible overhead, precisely 1/3584 and 1/18944 of the total computation for vision and LLMs parts. 

\begin{figure}[h]
    \centering    
    \includegraphics[width=\linewidth]{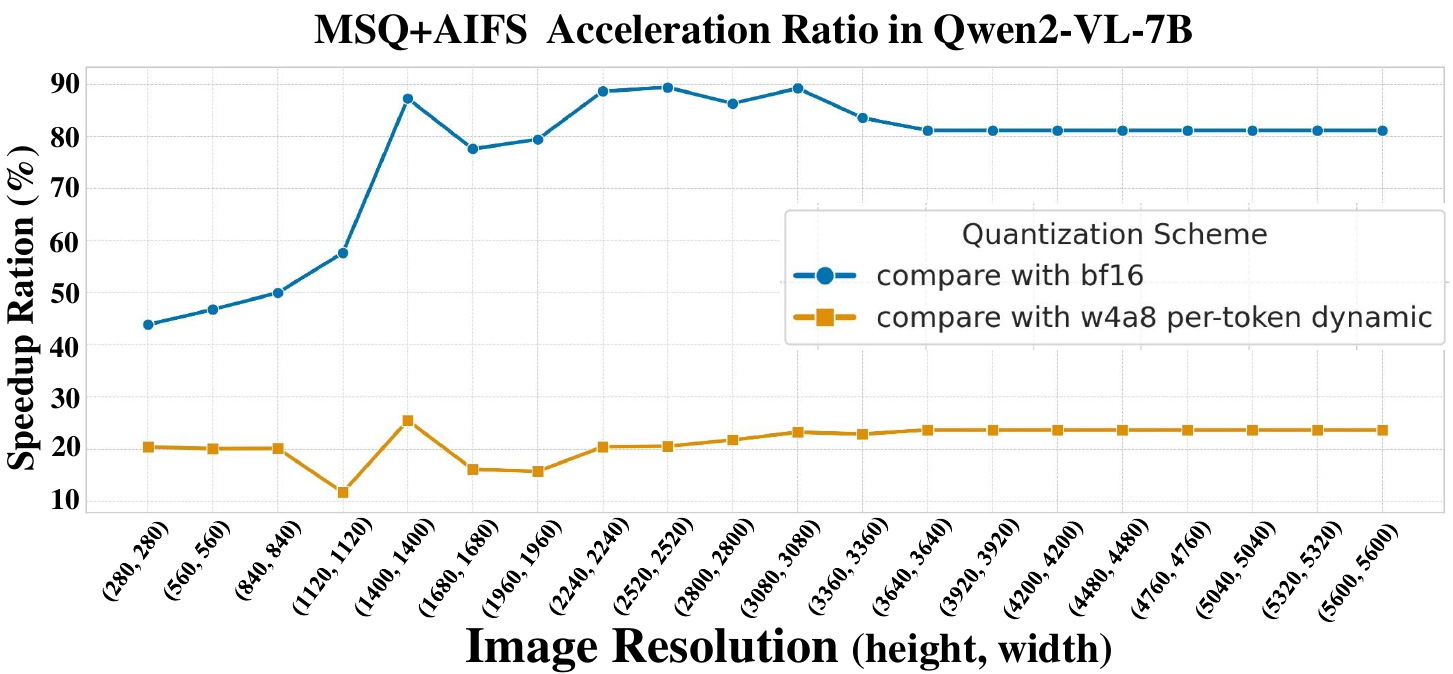}
    \vspace{-2mm}
    \caption{The Speedup of AIFS+MSQ on Qwen2-VL-7B.}
    \label{fig:AIFS_speed}
    \vspace{-2mm}
\end{figure}

\noindent\paragraph{\textbf{Accuracy \& Speedup of AIFS}} While \textbf{MSQ} addresses the modality discrepancy, \textbf{AIFS} further reorders visual and textual tokens into a unified sequence to enable efficient per-tensor static quantization of activations. Table~\ref{table:AIFS_laten} shows that \textbf{MSQ}+\textbf{AIFS} not only achieves the same speedup as naïve per-tensor static quantization but also maintains near-floating-point accuracy across linear layers. Figure~\ref{fig:AIFS_speed} further illustrates that \textbf{AIFS} yields speedups of \textbf{20\%--80\%} as resolution increases, corroborating that rearranging tokens avoids the high overhead of dynamic per-token quantization. 

\vspace{-2mm}
\subsection{Latency and Memory Analysis}
We evaluate \textbf{MQuant}'s performance across three key metrics: (1) inference speedup, (2) memory efficiency (tested over image resolutions from $280^2$-$5600^2$), and (3) decoding acceleration. All experiments employ the established ''\textit{text-image-text}'' input format in~\citep{duan2024vlmevalkit} with a fixed 15 textual tokens unless explicitly varied.

\begin{table}[h]
\vspace{-2mm}
\caption{Speedup of MSQ + AIFS on W4A8 setting.}
\vspace{-2mm}
\label{tab:speed_decode}
\centering
\Huge
\resizebox{\linewidth}{!}{
\begin{tabular}{ccccc|c}
\toprule
\textbf{Stage} & \textbf{BF16} & \textbf{Per-token Dyn.} & \textbf{Ours} & \textbf{Ours + GEMV} & \textbf{Speedup} \\ 
\toprule
\textbf{Prefill} & 1690 & 1253 & 1017 & - & \cellcolor{myblue!15} \color{mygreen1}{\textbf{+23\%}} \\
\textbf{Decode} & 17.5 & 16.4 & 13.06 & 8.2 & \cellcolor{myblue!15} \color{mygreen1}{\textbf{+100\%}} \\
\bottomrule
\end{tabular}
}
\vspace{-4mm}
\end{table}

\vspace{-2mm}
\noindent\paragraph{\textbf{Inference Speedup and Memory Savings}}
\textbf{\raisebox{-0.5pt}{\ding[1.1]{182\relax}} Overall Speedup} As shown in Table~\ref{tab:speedup}, \textbf{MQuant} surpasses BF16 and AWQ (W4-only) across all image resolutions, achieving up to \textbf{24.76\%} speedup over PyTorch at $840\times 840$. Even at higher resolutions (e.g., $5600^2$), \textbf{MQuant} maintains a \textbf{12.19\%} latency improvement, demonstrating scalability. \textbf{\raisebox{-0.5pt}{\ding[1.1]{183\relax}} Memory Efficiency.} Our method provides memory savings over both BF16 and AWQ, with reductions exceeding \textbf{100\%} compared to PyTorch (e.g., \textbf{152.92\%} at $840^2$). These benefits primarily arise from (1)~eliminating the overhead of token-wise scale computation, and (2)~converting mixed tokens into modality-decoupled tokens, avoiding slicing and concatenation when image resolutions become large. More resolution experiments are in Appendix ~\ref{Speed_mem_image}. \textbf{\raisebox{-0.5pt}{\ding[1.1]{184\relax}} Decoding Acceleration.} We also measure decode time for generating 2{,}000 tokens with a custom W4A8 \texttt{GEMV} kernel (Table \ref{tab:speed_decode}). Compared to per-token dynamic quantization, our \textbf{AIFS}+\textbf{MSQ} framework gains \textbf{23\%} speedup in the prefill stage and \textbf{100\%} speedup in decode stage. By shifting online per-toekn dynamic quantization to an offline per-tensor static approach, we greatly reduce inference overhead, especially for long-sequence tasks. Moreover, since visual tokens are generally pricier than textual ones~\citep{duan2024vlmevalkit}, these improvements translate to notable real-world cost reductions (e.g., $\approx30\%$ for OpenAI token pricing).

\begin{table*}[t]
\vspace{-1mm}
\caption{Multi-Batch speedup comparison of MSQ + AIFS on W4A8 setting. Each row shows the cumulative total of text tokens, images, and textual responses for multi-turn inference.}
\vspace{-2mm}
\label{tab:speed_multibatch}
\centering
\Huge
\resizebox{\linewidth}{!}{
\begin{tabular}{c|ccc|cc|c|cc|c|cc|c}
\toprule
\multirow{2}{*}{\textbf{Batch}} 
& \multicolumn{3}{c|}{\textbf{Config (Text+Image+Text)}} 
& \multicolumn{2}{c|}{\textbf{Prefill (s)}} 
& \multirow{2}{*}{\textbf{Improve$\uparrow$}} 
& \multicolumn{2}{c|}{\textbf{Decode (s)}}  
& \multirow{2}{*}{\textbf{Improve$\uparrow$}} 
& \multicolumn{2}{c|}{\textbf{All (s)}}  
& \multirow{2}{*}{\textbf{Improve$\uparrow$}} \\
& \textbf{Text} & \textbf{Image} & \textbf{Text} 
& \textbf{bfp16} & \textbf{MQuant} & 
& \textbf{bfp16} & \textbf{MQuant} & 
& \textbf{bfp16} & \textbf{MQuant} & \\
\midrule
1 & 10 & 2240$\times$2240 & 50 
  & 2.54 & 1.93 & \cellcolor{myblue!15}\color{mygreen1}{\textbf{+31.6\%}} 
  & 18.01 & 12.89 & \cellcolor{myblue!15}\color{mygreen1}{\textbf{+39.7\%}}  
  & 20.55 & 14.82 & \cellcolor{myblue!15}\color{mygreen1}{\textbf{+38.7\%}}  \\
2 & 10/10 & 2240$\times$2240 / 2240$\times$2240 & 50/100 
  & 5.42 & 4.15 & \cellcolor{myblue!15}\color{mygreen1}{\textbf{+30.6\%}} 
  & 37.82 & 31.56 & \cellcolor{myblue!15}\color{mygreen1}{\textbf{+19.8\%}} 
  & 43.24 & 35.71 & \cellcolor{myblue!15}\color{mygreen1}{\textbf{+21.1\%}} \\
3 & 10/10/10 & 2240$\times$2240 / 2240$\times$2240 / 2240$\times$2240 & 50/100/150 
  & 8.24 & 6.42 & \cellcolor{myblue!15}\color{mygreen1}{\textbf{+28.3\%}} 
  & 48.03 & 40.35 & \cellcolor{myblue!15}\color{mygreen1}{\textbf{+19.0\%}} 
  & 56.27 & 46.77 & \cellcolor{myblue!15}\color{mygreen1}{\textbf{+20.3\%}} \\
4 & 10/10/10/10 & 2240$\times$2240 / 2240$\times$2240 / 2240$\times$2240 / 2240$\times$2240 & 50/100/150/200 
  & 11.17 & 8.67 & \cellcolor{myblue!15}\color{mygreen1}{\textbf{+28.9\%}} 
  & 59.09 & 49.92 & \cellcolor{myblue!15}\color{mygreen1}{\textbf{+8.4\%}}
  & 70.26 & 58.59 & \cellcolor{myblue!15}\color{mygreen1}{\textbf{+20.0\%}} \\
\bottomrule
\end{tabular}
}
\vspace{-2mm}
\end{table*}

\begin{table}[h]
\vspace{-2mm}
\caption{Latency comparison under multi-turns setting.}
\vspace{-2mm}
\label{table:multi-turns-comparison}
\centering
\resizebox{\linewidth}{!}{
\begin{tabular}{c|ccc|cc|c}
\toprule
\multirow{2}{*}{\textbf{Turns}}  & \multicolumn{3}{c|}{\textbf{Config in one Turn}} & \multicolumn{2}{c|}{\textbf{All(s)}} & \multirow{2}{*}{\textbf{Improve $\uparrow$}} \\
               & \textbf{Text} & \textbf{Image} & \textbf{Text} & \textbf{bfp16} & \textbf{Ours} & \textbf{} \\ 
\midrule
1              & 10              & 2240x2240               &  50             & 20.55         & 14.82         & \cellcolor{myblue!15}\color{mygreen1}{\textbf{+38.7\%}} \\ 
2              & 10            & 2240x2240      & 50            & 44.06         & 32.61         & \cellcolor{myblue!15} \color{mygreen1}{\textbf{+35.1\%}} \\ 
3              & 10            & 2240x2240      & 50            & 76.67         & 59.48         & \cellcolor{myblue!15} \color{mygreen1}{\textbf{+28.9\%}} \\ 
\bottomrule
\end{tabular}
}
\vspace{-2mm}
\end{table}

\vspace{-2mm}
\noindent\paragraph{\textbf{Acceleration for Multi-batch and Multi-turn Inference}} \textbf{\raisebox{-0.5pt}{\ding[1.1]{182\relax}} For multi-batch scenarios}, we use 2240$\times$2240 resolution images with ”text-image-text” inputs, we evaluate batches containing 50-200 text tokens while generating 512 decoded tokens. Sequence alignment employs left-padding (\texttt{pad\_token\_id}) with proper masking (see Appendix A.4). AIFS maintains full multi-batch compatibility without overhead, enabling consistent 20\% faster inference versus FP baselines across batch sizes 1-4 (Table~\ref{tab:speed_multibatch}). \textbf{\raisebox{-0.5pt}{\ding[1.1]{183\relax}} For multi-turn dialogue}, under identical resolution and input format (50 text tokens per turn + 512 decoded tokens), MQuant achieves 38.7\% faster end-to-end inference than BF16 for 1-3 turns (Table~\ref{table:multi-turns-comparison}) through optimized key-value caches and position IDs across turns. %we preserve context during the dialogue. In Table~\ref{table:multi-turns-comparison}, MQuant reduces end-to-end inference time by up to 38.7\% over the BF16 baseline for 1-3 turns, showing the \emph{MQuant}'s efficiency in multi-turn settings.

\begin{table}[h]
\vspace{-2mm}
\caption{Comparison under weight-only settings on Qwen2-VL-7B~\citep{Qwen2VL}. ${\dagger}$ means re-implementation based on the official weight-only setting with a group size of 128.}
\vspace{-2mm}
\label{table:weight-only}
\centering
\Huge
\resizebox{\linewidth}{!}{
\begin{tabular}{c|cc|cccc}
\toprule
\multirow{2}{*}{\textbf{Method}}& \multicolumn{2}{c|}{\textbf{Bits Setting}} & \multirow{2}{*}{\textbf{T.VQA $\uparrow$}} & \multirow{2}{*}{\textbf{D.VQA $\uparrow$}} & \multirow{2}{*}{\textbf{OCRB.$\uparrow$}} & \multirow{2}{*}{\textbf{MME$\uparrow$}} \\
& Visual & LLM & \\
\toprule
\rowcolor{gray!10}
- & BF16 & BF16 & 84.43 &93.87 &842 &2319\\
GPTQ (g128)$^{\dagger}$  &BF16 &W8& 84.33 &93.97 &842 &2313 \\
GPTQ (g128)$^{\dagger}$ &BF16 &W4& 84.18 &93.25 &831 &2285 \\
 AWQ (g128)$^{\dagger}$ &BF16 &W4 &  83.93 &93.13 &828 &2252 \\ \cline{1-7} 
% \hline
\rowcolor{gray!15}
\textbf{MQuant} (g128)  & BF16 &W4 & 84.55 &93.18 &\textbf{832} &\textbf{2304} \\ \cline{1-7}
% \hline
  \rowcolor{gray!25}
\textbf{MQuant} (g128) & W4 &W4& \textbf{84.70} &\textbf{93.57} &828 &2292 \\ \cline{1-7}
  \rowcolor{myblue!20}
\textbf{MQuant} & W4A8 &W4A8& 84.32 & 93.58 &824 &2255 \\
\bottomrule
\end{tabular}}
\vspace{-2mm}
\end{table}

\vspace{-2mm}
\noindent\paragraph{\textbf{Weight-only Quantization}} We evaluate our weight-only quantization using GPTQ~\citep{frantar2022gptq} and AWQ~\citep{lin2023awq} under consistent settings (group size=128), quantizing only the LLM while keeping the visual encoder in BF16. As shown in Table ~\ref{table:weight-only}, our W4 LLM quantization achieves near-lossless accuracy comparable to existing methods. When extended to W4 quantization, MQuant maintains comparable performance with the BF16 baseline. The W4A8 quantization similarly shows competitive or superior results compared to weight-only approaches. These experiments validate MQuant's effectiveness across: (1) weight-only vs. weight-activation schemes, and (2) partial vs. full-model quantization scenarios.

\vspace{-2mm}
\section{Conclusion} \label{sec:conclusion}
% \vspace{-1mm}
In this paper, we propose \emph{MQuant}, an accurate and efficient PTQ framework specifically designed for MLLMs. Our approach addresses the challenges of MLLMs by applying \textbf{Modality-Specific Static Quantization (MSQ)} to handle distribution mismatches between visual and text tokens, and introducing an \textbf{Attention-Invariant Flexible Switching (AIFS)} mechanism to reduce TTFT with static per-tensor scaling. Furthermore, we reveal that online Hadamard rotations can lead to weight outliers, and propose \textbf{Rotation Magnitude Suppression (RMS)} to mitigate this issue. Extensive experiments on 5 mainstream MLLMs demonstrate that \emph{MQuant} achieves near-floating-point accuracy under the W4A8 setting, highlighting its potential to advance MLLMs quantization and practical deployment in resource-constrained environments. For the quantization of video-based MLLMs, we also discuss in Appendix~\ref{video_MLLMs}.

% \clearpage
% \section*{Acknowledge}
% We thank all anonymous reviewers, ACM MM Program Committee, and Area Committee for their kind help of this work.
\bibliographystyle{ACM-Reference-Format}
\bibliography{samples/sample-base}

%%% -*-BibTeX-*-
%%% Do NOT edit. File created by BibTeX with style
%%% ACM-Reference-Format-Journals [18-Jan-2012].

\begin{thebibliography}{71}

%%% ====================================================================
%%% NOTE TO THE USER: you can override these defaults by providing
%%% customized versions of any of these macros before the \bibliography
%%% command.  Each of them MUST provide its own final punctuation,
%%% except for \shownote{}, \showDOI{}, and \showURL{}.  The latter two
%%% do not use final punctuation, in order to avoid confusing it with
%%% the Web address.
%%%
%%% To suppress output of a particular field, define its macro to expand
%%% to an empty string, or better, \unskip, like this:
%%%
%%% \newcommand{\showDOI}[1]{\unskip}   % LaTeX syntax
%%%
%%% \def \showDOI #1{\unskip}           % plain TeX syntax
%%%
%%% ====================================================================

\ifx \showCODEN    \undefined \def \showCODEN     #1{\unskip}     \fi
\ifx \showDOI      \undefined \def \showDOI       #1{#1}\fi
\ifx \showISBNx    \undefined \def \showISBNx     #1{\unskip}     \fi
\ifx \showISBNxiii \undefined \def \showISBNxiii  #1{\unskip}     \fi
\ifx \showISSN     \undefined \def \showISSN      #1{\unskip}     \fi
\ifx \showLCCN     \undefined \def \showLCCN      #1{\unskip}     \fi
\ifx \shownote     \undefined \def \shownote      #1{#1}          \fi
\ifx \showarticletitle \undefined \def \showarticletitle #1{#1}   \fi
\ifx \showURL      \undefined \def \showURL       {\relax}        \fi
% The following commands are used for tagged output and should be
% invisible to TeX
\providecommand\bibfield[2]{#2}
\providecommand\bibinfo[2]{#2}
\providecommand\natexlab[1]{#1}
\providecommand\showeprint[2][]{arXiv:#2}

\bibitem[Achiam et~al\mbox{.}(2023a)]%
        {achiam2023gpt4}
\bibfield{author}{\bibinfo{person}{Josh Achiam}, \bibinfo{person}{Steven Adler}, \bibinfo{person}{Sandhini Agarwal}, \bibinfo{person}{Lama Ahmad}, \bibinfo{person}{Ilge Akkaya}, \bibinfo{person}{Florencia~Leoni Aleman}, \bibinfo{person}{Diogo Almeida}, \bibinfo{person}{Janko Altenschmidt}, \bibinfo{person}{Sam Altman}, \bibinfo{person}{Shyamal Anadkat}, {et~al\mbox{.}}} \bibinfo{year}{2023}\natexlab{a}.
\newblock \showarticletitle{{GPT-4} technical report}.
\newblock \bibinfo{journal}{\emph{arXiv preprint arXiv:2303.08774}} (\bibinfo{year}{2023}).
\newblock


\bibitem[Achiam et~al\mbox{.}(2023b)]%
        {achiam2023gpt}
\bibfield{author}{\bibinfo{person}{Josh Achiam}, \bibinfo{person}{Steven Adler}, \bibinfo{person}{Sandhini Agarwal}, \bibinfo{person}{Lama Ahmad}, \bibinfo{person}{Ilge Akkaya}, \bibinfo{person}{Florencia~Leoni Aleman}, \bibinfo{person}{Diogo Almeida}, \bibinfo{person}{Janko Altenschmidt}, \bibinfo{person}{Sam Altman}, \bibinfo{person}{Shyamal Anadkat}, {et~al\mbox{.}}} \bibinfo{year}{2023}\natexlab{b}.
\newblock \showarticletitle{Gpt-4 technical report}.
\newblock \bibinfo{journal}{\emph{arXiv preprint arXiv:2303.08774}} (\bibinfo{year}{2023}).
\newblock


\bibitem[Alayrac et~al\mbox{.}(2022)]%
        {alayrac2022flamingo}
\bibfield{author}{\bibinfo{person}{Jean-Baptiste Alayrac}, \bibinfo{person}{Jeff Donahue}, \bibinfo{person}{Pauline Luc}, \bibinfo{person}{Antoine Miech}, \bibinfo{person}{Iain Barr}, \bibinfo{person}{Yana Hasson}, \bibinfo{person}{Karel Lenc}, \bibinfo{person}{Arthur Mensch}, \bibinfo{person}{Katherine Millican}, \bibinfo{person}{Malcolm Reynolds}, {et~al\mbox{.}}} \bibinfo{year}{2022}\natexlab{}.
\newblock \showarticletitle{{Flamingo}: A visual language model for few-shot learning}.
\newblock \bibinfo{journal}{\emph{NeurIPS}}  \bibinfo{volume}{35} (\bibinfo{year}{2022}), \bibinfo{pages}{23716--23736}.
\newblock


\bibitem[Ashkboos et~al\mbox{.}(2024a)]%
        {ashkboos2024slicegpt}
\bibfield{author}{\bibinfo{person}{Saleh Ashkboos}, \bibinfo{person}{Maximilian~L Croci}, \bibinfo{person}{Marcelo Gennari~do Nascimento}, \bibinfo{person}{Torsten Hoefler}, {and} \bibinfo{person}{James Hensman}.} \bibinfo{year}{2024}\natexlab{a}.
\newblock \showarticletitle{Slicegpt: Compress large language models by deleting rows and columns}.
\newblock \bibinfo{journal}{\emph{International Conference on Learning Representations (ICLR)}} (\bibinfo{year}{2024}).
\newblock


\bibitem[Ashkboos et~al\mbox{.}(2024b)]%
        {ashkboos2024quarot}
\bibfield{author}{\bibinfo{person}{Saleh Ashkboos}, \bibinfo{person}{Amirkeivan Mohtashami}, \bibinfo{person}{Maximilian~L Croci}, \bibinfo{person}{Bo Li}, \bibinfo{person}{Martin Jaggi}, \bibinfo{person}{Dan Alistarh}, \bibinfo{person}{Torsten Hoefler}, {and} \bibinfo{person}{James Hensman}.} \bibinfo{year}{2024}\natexlab{b}.
\newblock \showarticletitle{Quarot: Outlier-free 4-bit inference in rotated llms}.
\newblock \bibinfo{journal}{\emph{arXiv preprint arXiv:2404.00456}} (\bibinfo{year}{2024}).
\newblock


\bibitem[Ba(2016)]%
        {layernorm}
\bibfield{author}{\bibinfo{person}{JL Ba}.} \bibinfo{year}{2016}\natexlab{}.
\newblock \showarticletitle{Layer normalization}.
\newblock \bibinfo{journal}{\emph{arXiv preprint arXiv:1607.06450}} (\bibinfo{year}{2016}).
\newblock


\bibitem[Bai et~al\mbox{.}(2023a)]%
        {qwen}
\bibfield{author}{\bibinfo{person}{Jinze Bai}, \bibinfo{person}{Shuai Bai}, \bibinfo{person}{Yunfei Chu}, \bibinfo{person}{Zeyu Cui}, {et~al\mbox{.}}} \bibinfo{year}{2023}\natexlab{a}.
\newblock \showarticletitle{Qwen Technical Report}.
\newblock \bibinfo{journal}{\emph{arXiv preprint arXiv:2309.16609}} (\bibinfo{year}{2023}).
\newblock


\bibitem[Bai et~al\mbox{.}(2023b)]%
        {qwenvl}
\bibfield{author}{\bibinfo{person}{Jinze Bai}, \bibinfo{person}{Shuai Bai}, \bibinfo{person}{Shusheng Yang}, \bibinfo{person}{Shijie Wang}, \bibinfo{person}{Sinan Tan}, \bibinfo{person}{Peng Wang}, \bibinfo{person}{Junyang Lin}, \bibinfo{person}{Chang Zhou}, {and} \bibinfo{person}{Jingren Zhou}.} \bibinfo{year}{2023}\natexlab{b}.
\newblock \showarticletitle{Qwen-vl: A frontier large vision-language model with versatile abilities}.
\newblock \bibinfo{journal}{\emph{arXiv preprint arXiv:2308.12966}} (\bibinfo{year}{2023}).
\newblock


\bibitem[Brown et~al\mbox{.}(2020)]%
        {brown2020language}
\bibfield{author}{\bibinfo{person}{Tom Brown}, \bibinfo{person}{Benjamin Mann}, \bibinfo{person}{Nick Ryder}, \bibinfo{person}{Melanie Subbiah}, \bibinfo{person}{Jared~D Kaplan}, \bibinfo{person}{Prafulla Dhariwal}, \bibinfo{person}{Arvind Neelakantan}, \bibinfo{person}{Pranav Shyam}, \bibinfo{person}{Girish Sastry}, \bibinfo{person}{Amanda Askell}, {et~al\mbox{.}}} \bibinfo{year}{2020}\natexlab{}.
\newblock \showarticletitle{Language models are few-shot learners}.
\newblock \bibinfo{journal}{\emph{Advances in Neural Information Processing Systems}} (\bibinfo{year}{2020}).
\newblock


\bibitem[Cai et~al\mbox{.}(2024)]%
        {internlm}
\bibfield{author}{\bibinfo{person}{Zheng Cai}, \bibinfo{person}{Maosong Cao}, \bibinfo{person}{Haojiong Chen}, \bibinfo{person}{Kai Chen}, {et~al\mbox{.}}} \bibinfo{year}{2024}\natexlab{}.
\newblock \bibinfo{title}{InternLM2 Technical Report}.
\newblock
\newblock
\showeprint[arxiv]{2403.17297}~[cs.CL]


\bibitem[Chee et~al\mbox{.}(2024)]%
        {chee2024quip}
\bibfield{author}{\bibinfo{person}{Jerry Chee}, \bibinfo{person}{Yaohui Cai}, \bibinfo{person}{Volodymyr Kuleshov}, {and} \bibinfo{person}{Christopher~M De~Sa}.} \bibinfo{year}{2024}\natexlab{}.
\newblock \showarticletitle{Quip: 2-bit quantization of large language models with guarantees}.
\newblock \bibinfo{journal}{\emph{Advances in Neural Information Processing Systems}}  \bibinfo{volume}{36} (\bibinfo{year}{2024}).
\newblock


\bibitem[Chen et~al\mbox{.}(2024a)]%
        {chen2024prefixquant}
\bibfield{author}{\bibinfo{person}{Mengzhao Chen}, \bibinfo{person}{Yi Liu}, \bibinfo{person}{Jiahao Wang}, \bibinfo{person}{Yi Bin}, \bibinfo{person}{Wenqi Shao}, {and} \bibinfo{person}{Ping Luo}.} \bibinfo{year}{2024}\natexlab{a}.
\newblock \showarticletitle{Prefixquant: Static quantization beats dynamic through prefixed outliers in llms}.
\newblock \bibinfo{journal}{\emph{arXiv preprint arXiv:2410.05265}} (\bibinfo{year}{2024}).
\newblock


\bibitem[Chen et~al\mbox{.}({[n.\,d.]})]%
        {chenmoequant}
\bibfield{author}{\bibinfo{person}{Zhixuan Chen}, \bibinfo{person}{Xing Hu}, \bibinfo{person}{Dawei Yang}, \bibinfo{person}{Zukang Xu}, \bibinfo{person}{Zhihang Yuan}, \bibinfo{person}{Sifan Zhou}, {et~al\mbox{.}}} \bibinfo{year}{[n.\,d.]}\natexlab{}.
\newblock \showarticletitle{MoEQuant: Enhancing Quantization for Mixture-of-Experts Large Language Models via Expert-Balanced Sampling and Affinity Guidance}. In \bibinfo{booktitle}{\emph{Forty-second International Conference on Machine Learning}}.
\newblock


\bibitem[Chen et~al\mbox{.}(2024b)]%
        {internvl15}
\bibfield{author}{\bibinfo{person}{Zhe Chen}, \bibinfo{person}{Weiyun Wang}, \bibinfo{person}{Hao Tian}, \bibinfo{person}{Shenglong Ye}, \bibinfo{person}{Zhangwei Gao}, \bibinfo{person}{Erfei Cui}, \bibinfo{person}{Wenwen Tong}, \bibinfo{person}{Kongzhi Hu}, \bibinfo{person}{Jiapeng Luo}, \bibinfo{person}{Zheng Ma}, {et~al\mbox{.}}} \bibinfo{year}{2024}\natexlab{b}.
\newblock \showarticletitle{How Far Are We to GPT-4V? Closing the Gap to Commercial Multimodal Models with Open-Source Suites}.
\newblock \bibinfo{journal}{\emph{arXiv preprint arXiv:2404.16821}} (\bibinfo{year}{2024}).
\newblock


\bibitem[Chen et~al\mbox{.}(2024c)]%
        {chen2024internvl}
\bibfield{author}{\bibinfo{person}{Zhe Chen}, \bibinfo{person}{Jiannan Wu}, \bibinfo{person}{Wenhai Wang}, \bibinfo{person}{Weijie Su}, \bibinfo{person}{Guo Chen}, \bibinfo{person}{Sen Xing}, \bibinfo{person}{Muyan Zhong}, \bibinfo{person}{Qinglong Zhang}, \bibinfo{person}{Xizhou Zhu}, \bibinfo{person}{Lewei Lu}, {et~al\mbox{.}}} \bibinfo{year}{2024}\natexlab{c}.
\newblock \showarticletitle{Internvl: Scaling up vision foundation models and aligning for generic visual-linguistic tasks}. In \bibinfo{booktitle}{\emph{Proceedings of the IEEE/CVF Conference on Computer Vision and Pattern Recognition}}. \bibinfo{pages}{24185--24198}.
\newblock


\bibitem[Chu et~al\mbox{.}(2024)]%
        {chu2024mobilevlm2}
\bibfield{author}{\bibinfo{person}{Xiangxiang Chu}, \bibinfo{person}{Limeng Qiao}, \bibinfo{person}{Xinyu Zhang}, \bibinfo{person}{Shuang Xu}, \bibinfo{person}{Fei Wei}, \bibinfo{person}{Yang Yang}, \bibinfo{person}{Xiaofei Sun}, \bibinfo{person}{Yiming Hu}, \bibinfo{person}{Xinyang Lin}, \bibinfo{person}{Bo Zhang}, {et~al\mbox{.}}} \bibinfo{year}{2024}\natexlab{}.
\newblock \showarticletitle{MobileVLM V2: Faster and Stronger Baseline for Vision Language Model}.
\newblock \bibinfo{journal}{\emph{arXiv preprint arXiv:2402.03766}} (\bibinfo{year}{2024}).
\newblock


\bibitem[Dao et~al\mbox{.}(2022)]%
        {dao2022flashattention}
\bibfield{author}{\bibinfo{person}{Tri Dao}, \bibinfo{person}{Dan Fu}, \bibinfo{person}{Stefano Ermon}, \bibinfo{person}{Atri Rudra}, {and} \bibinfo{person}{Christopher R{\'e}}.} \bibinfo{year}{2022}\natexlab{}.
\newblock \showarticletitle{Flashattention: Fast and memory-efficient exact attention with io-awareness}.
\newblock \bibinfo{journal}{\emph{Advances in Neural Information Processing Systems}}  \bibinfo{volume}{35} (\bibinfo{year}{2022}), \bibinfo{pages}{16344--16359}.
\newblock


\bibitem[Dosovitskiy(2021)]%
        {vit}
\bibfield{author}{\bibinfo{person}{Alexey Dosovitskiy}.} \bibinfo{year}{2021}\natexlab{}.
\newblock \showarticletitle{An image is worth 16x16 words: Transformers for image recognition at scale}.
\newblock \bibinfo{journal}{\emph{ICLR}} (\bibinfo{year}{2021}).
\newblock


\bibitem[Duan et~al\mbox{.}(2024)]%
        {duan2024vlmevalkit}
\bibfield{author}{\bibinfo{person}{Haodong Duan}, \bibinfo{person}{Junming Yang}, \bibinfo{person}{Yuxuan Qiao}, \bibinfo{person}{Xinyu Fang}, \bibinfo{person}{Lin Chen}, \bibinfo{person}{Yuan Liu}, \bibinfo{person}{Xiaoyi Dong}, \bibinfo{person}{Yuhang Zang}, \bibinfo{person}{Pan Zhang}, \bibinfo{person}{Jiaqi Wang}, {et~al\mbox{.}}} \bibinfo{year}{2024}\natexlab{}.
\newblock \showarticletitle{Vlmevalkit: An open-source toolkit for evaluating large multi-modality models}. In \bibinfo{booktitle}{\emph{Proceedings of the 32nd ACM International Conference on Multimedia}}. \bibinfo{pages}{11198--11201}.
\newblock


\bibitem[Dubey et~al\mbox{.}(2024)]%
        {llama3}
\bibfield{author}{\bibinfo{person}{Abhimanyu Dubey}, \bibinfo{person}{Abhinav Jauhri}, \bibinfo{person}{Abhinav Pandey}, \bibinfo{person}{Abhishek Kadian}, \bibinfo{person}{Ahmad Al-Dahle}, \bibinfo{person}{Aiesha Letman}, \bibinfo{person}{Akhil Mathur}, \bibinfo{person}{Alan Schelten}, \bibinfo{person}{Amy Yang}, \bibinfo{person}{Angela Fan}, {et~al\mbox{.}}} \bibinfo{year}{2024}\natexlab{}.
\newblock \showarticletitle{The llama 3 herd of models}.
\newblock \bibinfo{journal}{\emph{arXiv preprint arXiv:2407.21783}} (\bibinfo{year}{2024}).
\newblock


\bibitem[Frantar et~al\mbox{.}(2022)]%
        {frantar2022gptq}
\bibfield{author}{\bibinfo{person}{Elias Frantar}, \bibinfo{person}{Saleh Ashkboos}, \bibinfo{person}{Torsten Hoefler}, {and} \bibinfo{person}{Dan Alistarh}.} \bibinfo{year}{2022}\natexlab{}.
\newblock \showarticletitle{GPTQ: Accurate Post-Training Quantization for Generative Pre-trained Transformers}.
\newblock \bibinfo{journal}{\emph{arXiv preprint arXiv:2210.17323}} (\bibinfo{year}{2022}).
\newblock


\bibitem[Fu et~al\mbox{.}(2023)]%
        {fu2023mme}
\bibfield{author}{\bibinfo{person}{Chaoyou Fu}, \bibinfo{person}{Peixian Chen}, \bibinfo{person}{Yunhang Shen}, \bibinfo{person}{Yulei Qin}, \bibinfo{person}{Mengdan Zhang}, \bibinfo{person}{Xu Lin}, \bibinfo{person}{Jinrui Yang}, \bibinfo{person}{Xiawu Zheng}, \bibinfo{person}{Ke Li}, \bibinfo{person}{Xing Sun}, \bibinfo{person}{Yunsheng Wu}, {and} \bibinfo{person}{Rongrong Ji}.} \bibinfo{year}{2023}\natexlab{}.
\newblock \showarticletitle{{MME}: A Comprehensive Evaluation Benchmark for Multimodal Large Language Models}.
\newblock \bibinfo{journal}{\emph{arXiv preprint arXiv:2306.13394}} (\bibinfo{year}{2023}).
\newblock


\bibitem[GLM et~al\mbox{.}(2024)]%
        {chatglm}
\bibfield{author}{\bibinfo{person}{Team GLM}, \bibinfo{person}{Aohan Zeng}, \bibinfo{person}{Bin Xu}, \bibinfo{person}{Bowen Wang}, \bibinfo{person}{Chenhui Zhang}, \bibinfo{person}{Da Yin}, \bibinfo{person}{Diego Rojas}, \bibinfo{person}{Guanyu Feng}, \bibinfo{person}{Hanlin Zhao}, \bibinfo{person}{Hanyu Lai}, {et~al\mbox{.}}} \bibinfo{year}{2024}\natexlab{}.
\newblock \showarticletitle{ChatGLM: A Family of Large Language Models from GLM-130B to GLM-4 All Tools}.
\newblock \bibinfo{journal}{\emph{arXiv preprint arXiv:2406.12793}} (\bibinfo{year}{2024}).
\newblock


\bibitem[Hong et~al\mbox{.}(2024)]%
        {CogVLM2}
\bibfield{author}{\bibinfo{person}{Wenyi Hong}, \bibinfo{person}{Weihan Wang}, \bibinfo{person}{Ming Ding}, \bibinfo{person}{Wenmeng Yu}, \bibinfo{person}{Qingsong Lv}, \bibinfo{person}{Yan Wang}, \bibinfo{person}{Yean Cheng}, \bibinfo{person}{Shiyu Huang}, \bibinfo{person}{Junhui Ji}, \bibinfo{person}{Zhao Xue}, {et~al\mbox{.}}} \bibinfo{year}{2024}\natexlab{}.
\newblock \showarticletitle{CogVLM2: Visual Language Models for Image and Video Understanding}.
\newblock \bibinfo{journal}{\emph{arXiv preprint arXiv:2408.16500}} (\bibinfo{year}{2024}).
\newblock


\bibitem[Hu et~al\mbox{.}(2024b)]%
        {hu2024minicpm}
\bibfield{author}{\bibinfo{person}{Shengding Hu}, \bibinfo{person}{Yuge Tu}, \bibinfo{person}{Xu Han}, \bibinfo{person}{Chaoqun He}, \bibinfo{person}{Ganqu Cui}, \bibinfo{person}{Xiang Long}, \bibinfo{person}{Zhi Zheng}, \bibinfo{person}{Yewei Fang}, \bibinfo{person}{Yuxiang Huang}, \bibinfo{person}{Weilin Zhao}, {et~al\mbox{.}}} \bibinfo{year}{2024}\natexlab{b}.
\newblock \showarticletitle{{MiniCPM}: Unveiling the Potential of Small Language Models with Scalable Training Strategies}.
\newblock \bibinfo{journal}{\emph{arXiv preprint arXiv:2404.06395}} (\bibinfo{year}{2024}).
\newblock


\bibitem[Hu et~al\mbox{.}(2024a)]%
        {hu2024llm}
\bibfield{author}{\bibinfo{person}{Xing Hu}, \bibinfo{person}{Yuan Chen}, \bibinfo{person}{Dawei Yang}, \bibinfo{person}{Sifan Zhou}, \bibinfo{person}{Zhihang Yuan}, \bibinfo{person}{Jiangyong Yu}, {and} \bibinfo{person}{Chen Xu}.} \bibinfo{year}{2024}\natexlab{a}.
\newblock \showarticletitle{I-LLM: Efficient Integer-Only Inference for Fully-Quantized Low-Bit Large Language Models}.
\newblock \bibinfo{journal}{\emph{arXiv preprint arXiv:2405.17849}} (\bibinfo{year}{2024}).
\newblock


\bibitem[Hu et~al\mbox{.}(2025)]%
        {hu2025ostquant}
\bibfield{author}{\bibinfo{person}{Xing Hu}, \bibinfo{person}{Yuan Cheng}, \bibinfo{person}{Dawei Yang}, \bibinfo{person}{Zukang Xu}, \bibinfo{person}{Zhihang Yuan}, \bibinfo{person}{Jiangyong Yu}, \bibinfo{person}{Chen Xu}, \bibinfo{person}{Zhe Jiang}, {and} \bibinfo{person}{Sifan Zhou}.} \bibinfo{year}{2025}\natexlab{}.
\newblock \showarticletitle{OstQuant: Refining Large Language Model Quantization with Orthogonal and Scaling Transformations for Better Distribution Fitting}.
\newblock \bibinfo{journal}{\emph{arXiv preprint arXiv:2501.13987}} (\bibinfo{year}{2025}).
\newblock


\bibitem[Huang et~al\mbox{.}(2024)]%
        {huang2023kosmos1}
\bibfield{author}{\bibinfo{person}{Shaohan Huang}, \bibinfo{person}{Li Dong}, \bibinfo{person}{Wenhui Wang}, \bibinfo{person}{Yaru Hao}, \bibinfo{person}{Saksham Singhal}, \bibinfo{person}{Shuming Ma}, \bibinfo{person}{Tengchao Lv}, \bibinfo{person}{Lei Cui}, \bibinfo{person}{Owais~Khan Mohammed}, \bibinfo{person}{Barun Patra}, {et~al\mbox{.}}} \bibinfo{year}{2024}\natexlab{}.
\newblock \showarticletitle{Language is not all you need: Aligning perception with language models}.
\newblock \bibinfo{journal}{\emph{NeurIPS}}  \bibinfo{volume}{36} (\bibinfo{year}{2024}).
\newblock


\bibitem[Ilharco et~al\mbox{.}(2021)]%
        {openclip}
\bibfield{author}{\bibinfo{person}{Gabriel Ilharco}, \bibinfo{person}{Mitchell Wortsman}, \bibinfo{person}{Ross Wightman}, \bibinfo{person}{Cade Gordon}, \bibinfo{person}{Nicholas Carlini}, \bibinfo{person}{Rohan Taori}, \bibinfo{person}{Achal Dave}, \bibinfo{person}{Vaishaal Shankar}, \bibinfo{person}{Hongseok Namkoong}, \bibinfo{person}{John Miller}, \bibinfo{person}{Hannaneh Hajishirzi}, \bibinfo{person}{Ali Farhadi}, {and} \bibinfo{person}{Ludwig Schmidt}.} \bibinfo{year}{2021}\natexlab{}.
\newblock \bibinfo{booktitle}{\emph{OpenCLIP}}.
\newblock
\urldef\tempurl%
\url{https://doi.org/10.5281/zenodo.5143773}
\showDOI{\tempurl}
\newblock
\shownote{If you use this software, please cite it as below.}.


\bibitem[Jiang et~al\mbox{.}(2024)]%
        {jiang2024ptq4ris}
\bibfield{author}{\bibinfo{person}{Xiaoyan Jiang}, \bibinfo{person}{Hang Yang}, \bibinfo{person}{Kaiying Zhu}, \bibinfo{person}{Xihe Qiu}, \bibinfo{person}{Shibo Zhao}, {and} \bibinfo{person}{Sifan Zhou}.} \bibinfo{year}{2024}\natexlab{}.
\newblock \showarticletitle{Ptq4ris: Post-training quantization for referring image segmentation}.
\newblock \bibinfo{journal}{\emph{arXiv preprint arXiv:2409.17020}} (\bibinfo{year}{2024}).
\newblock


\bibitem[Li et~al\mbox{.}(2023a)]%
        {li2023blip2}
\bibfield{author}{\bibinfo{person}{Junnan Li}, \bibinfo{person}{Dongxu Li}, \bibinfo{person}{Silvio Savarese}, {and} \bibinfo{person}{Steven Hoi}.} \bibinfo{year}{2023}\natexlab{a}.
\newblock \showarticletitle{{BLIP-2}: Bootstrapping language-image pre-training with frozen image encoders and large language models}.
\newblock \bibinfo{journal}{\emph{ICML}} (\bibinfo{year}{2023}), \bibinfo{pages}{19730--19742}.
\newblock


\bibitem[Li et~al\mbox{.}(2024b)]%
        {li2024norm}
\bibfield{author}{\bibinfo{person}{Liang Li}, \bibinfo{person}{Qingyuan Li}, \bibinfo{person}{Bo Zhang}, {and} \bibinfo{person}{Xiangxiang Chu}.} \bibinfo{year}{2024}\natexlab{b}.
\newblock \showarticletitle{Norm tweaking: High-performance low-bit quantization of large language models}. In \bibinfo{booktitle}{\emph{Proceedings of the AAAI Conference on Artificial Intelligence}}, Vol.~\bibinfo{volume}{38}. \bibinfo{pages}{18536--18544}.
\newblock


\bibitem[Li et~al\mbox{.}(2023b)]%
        {li2023fptq}
\bibfield{author}{\bibinfo{person}{Qingyuan Li}, \bibinfo{person}{Yifan Zhang}, \bibinfo{person}{Liang Li}, \bibinfo{person}{Peng Yao}, \bibinfo{person}{Bo Zhang}, \bibinfo{person}{Xiangxiang Chu}, \bibinfo{person}{Yerui Sun}, \bibinfo{person}{Li Du}, {and} \bibinfo{person}{Yuchen Xie}.} \bibinfo{year}{2023}\natexlab{b}.
\newblock \showarticletitle{Fptq: Fine-grained post-training quantization for large language models}.
\newblock \bibinfo{journal}{\emph{arXiv preprint arXiv:2308.15987}} (\bibinfo{year}{2023}).
\newblock


\bibitem[Li et~al\mbox{.}(2024a)]%
        {li2024mbq}
\bibfield{author}{\bibinfo{person}{Shiyao Li}, \bibinfo{person}{Yingchun Hu}, \bibinfo{person}{Xuefei Ning}, \bibinfo{person}{Xihui Liu}, \bibinfo{person}{Ke Hong}, \bibinfo{person}{Xiaotao Jia}, \bibinfo{person}{Xiuhong Li}, \bibinfo{person}{Yaqi Yan}, \bibinfo{person}{Pei Ran}, \bibinfo{person}{Guohao Dai}, \bibinfo{person}{Shengen Yan}, \bibinfo{person}{Huazhong Yang}, {and} \bibinfo{person}{Yu Wang}.} \bibinfo{year}{2024}\natexlab{a}.
\newblock \bibinfo{title}{MBQ: Modality-Balanced Quantization for Large Vision-Language Models}.
\newblock
\newblock
\showeprint[arxiv]{2412.19509}~[cs.CV]
\urldef\tempurl%
\url{https://arxiv.org/abs/2412.19509}
\showURL{%
\tempurl}


\bibitem[Li et~al\mbox{.}(2024c)]%
        {li2024minigemini}
\bibfield{author}{\bibinfo{person}{Yanwei Li}, \bibinfo{person}{Yuechen Zhang}, \bibinfo{person}{Chengyao Wang}, \bibinfo{person}{Zhisheng Zhong}, \bibinfo{person}{Yixin Chen}, \bibinfo{person}{Ruihang Chu}, \bibinfo{person}{Shaoteng Liu}, {and} \bibinfo{person}{Jiaya Jia}.} \bibinfo{year}{2024}\natexlab{c}.
\newblock \showarticletitle{{Mini-Gemini}: Mining the Potential of Multi-modality Vision Language Models}.
\newblock \bibinfo{journal}{\emph{arXiv preprint arXiv:2403.18814}} (\bibinfo{year}{2024}).
\newblock


\bibitem[Lin et~al\mbox{.}(2023)]%
        {lin2023awq}
\bibfield{author}{\bibinfo{person}{Ji Lin}, \bibinfo{person}{Jiaming Tang}, \bibinfo{person}{Haotian Tang}, \bibinfo{person}{Shang Yang}, \bibinfo{person}{Xingyu Dang}, {and} \bibinfo{person}{Song Han}.} \bibinfo{year}{2023}\natexlab{}.
\newblock \showarticletitle{AWQ: Activation-aware Weight Quantization for LLM Compression and Acceleration}.
\newblock \bibinfo{journal}{\emph{arXiv preprint arXiv:2306.00978}} (\bibinfo{year}{2023}).
\newblock


\bibitem[Liu et~al\mbox{.}(2024)]%
        {liu2023llava}
\bibfield{author}{\bibinfo{person}{Haotian Liu}, \bibinfo{person}{Chunyuan Li}, \bibinfo{person}{Qingyang Wu}, {and} \bibinfo{person}{Yong~Jae Lee}.} \bibinfo{year}{2024}\natexlab{}.
\newblock \showarticletitle{Visual Instruction Tuning}.
\newblock \bibinfo{journal}{\emph{NeurIPS}}  \bibinfo{volume}{36} (\bibinfo{year}{2024}).
\newblock


\bibitem[Liu et~al\mbox{.}(2023)]%
        {liu2023ocrbench}
\bibfield{author}{\bibinfo{person}{Yuliang Liu}, \bibinfo{person}{Zhang Li}, \bibinfo{person}{Hongliang Li}, \bibinfo{person}{Wenwen Yu}, \bibinfo{person}{Mingxin Huang}, \bibinfo{person}{Dezhi Peng}, \bibinfo{person}{Mingyu Liu}, \bibinfo{person}{Mingrui Chen}, \bibinfo{person}{Chunyuan Li}, \bibinfo{person}{Lianwen Jin}, {et~al\mbox{.}}} \bibinfo{year}{2023}\natexlab{}.
\newblock \showarticletitle{On the hidden mystery of {OCR} in large multimodal models}.
\newblock \bibinfo{journal}{\emph{arXiv preprint arXiv:2305.07895}} (\bibinfo{year}{2023}).
\newblock


\bibitem[Lu et~al\mbox{.}(2022)]%
        {ScienceQA}
\bibfield{author}{\bibinfo{person}{Pan Lu}, \bibinfo{person}{Swaroop Mishra}, \bibinfo{person}{Tanglin Xia}, \bibinfo{person}{Liang Qiu}, \bibinfo{person}{Kai-Wei Chang}, \bibinfo{person}{Song-Chun Zhu}, \bibinfo{person}{Oyvind Tafjord}, \bibinfo{person}{Peter Clark}, {and} \bibinfo{person}{Ashwin Kalyan}.} \bibinfo{year}{2022}\natexlab{}.
\newblock \showarticletitle{Learn to explain: Multimodal reasoning via thought chains for science question answering}.
\newblock \bibinfo{journal}{\emph{Advances in Neural Information Processing Systems}}  \bibinfo{volume}{35} (\bibinfo{year}{2022}), \bibinfo{pages}{2507--2521}.
\newblock


\bibitem[Mathew et~al\mbox{.}(2021)]%
        {mathew2021docvqa}
\bibfield{author}{\bibinfo{person}{Minesh Mathew}, \bibinfo{person}{Dimosthenis Karatzas}, {and} \bibinfo{person}{CV Jawahar}.} \bibinfo{year}{2021}\natexlab{}.
\newblock \showarticletitle{{DocVQA}: A dataset for {VQA} on document images}. In \bibinfo{booktitle}{\emph{WACV}}. \bibinfo{pages}{2200--2209}.
\newblock


\bibitem[Radford et~al\mbox{.}(2021)]%
        {radford2021clip}
\bibfield{author}{\bibinfo{person}{Alec Radford}, \bibinfo{person}{Jong~Wook Kim}, \bibinfo{person}{Chris Hallacy}, \bibinfo{person}{Aditya Ramesh}, \bibinfo{person}{Gabriel Goh}, \bibinfo{person}{Sandhini Agarwal}, \bibinfo{person}{Girish Sastry}, \bibinfo{person}{Amanda Askell}, \bibinfo{person}{Pamela Mishkin}, \bibinfo{person}{Jack Clark}, {et~al\mbox{.}}} \bibinfo{year}{2021}\natexlab{}.
\newblock \showarticletitle{Learning transferable visual models from natural language supervision}. In \bibinfo{booktitle}{\emph{ICML}}. PMLR, \bibinfo{pages}{8748--8763}.
\newblock


\bibitem[Reid et~al\mbox{.}(2024)]%
        {reid2024gemini}
\bibfield{author}{\bibinfo{person}{Machel Reid}, \bibinfo{person}{Nikolay Savinov}, \bibinfo{person}{Denis Teplyashin}, \bibinfo{person}{Dmitry Lepikhin}, \bibinfo{person}{Timothy Lillicrap}, \bibinfo{person}{Jean-baptiste Alayrac}, \bibinfo{person}{Radu Soricut}, \bibinfo{person}{Angeliki Lazaridou}, \bibinfo{person}{Orhan Firat}, \bibinfo{person}{Julian Schrittwieser}, {et~al\mbox{.}}} \bibinfo{year}{2024}\natexlab{}.
\newblock \showarticletitle{{Gemini 1.5}: Unlocking multimodal understanding across millions of tokens of context}.
\newblock \bibinfo{journal}{\emph{arXiv preprint arXiv:2403.05530}} (\bibinfo{year}{2024}).
\newblock


\bibitem[Shao et~al\mbox{.}(2023)]%
        {2023omniquant}
\bibfield{author}{\bibinfo{person}{Wenqi Shao}, \bibinfo{person}{Mengzhao Chen}, \bibinfo{person}{Zhaoyang Zhang}, \bibinfo{person}{Peng Xu}, \bibinfo{person}{Lirui Zhao}, \bibinfo{person}{Zhiqian Li}, \bibinfo{person}{Kaipeng Zhang}, \bibinfo{person}{Peng Gao}, \bibinfo{person}{Yu Qiao}, {and} \bibinfo{person}{Ping Luo}.} \bibinfo{year}{2023}\natexlab{}.
\newblock \showarticletitle{OmniQuant: Omnidirectionally Calibrated Quantization for Large Language Models}.
\newblock \bibinfo{journal}{\emph{CoRR}}  \bibinfo{volume}{abs/2308.13137} (\bibinfo{year}{2023}).
\newblock


\bibitem[Singh et~al\mbox{.}(2019)]%
        {singh2019textvqa}
\bibfield{author}{\bibinfo{person}{Amanpreet Singh}, \bibinfo{person}{Vivek Natarajan}, \bibinfo{person}{Meet Shah}, \bibinfo{person}{Yu Jiang}, \bibinfo{person}{Xinlei Chen}, \bibinfo{person}{Dhruv Batra}, \bibinfo{person}{Devi Parikh}, {and} \bibinfo{person}{Marcus Rohrbach}.} \bibinfo{year}{2019}\natexlab{}.
\newblock \showarticletitle{Towards {VQA} models that can read}. In \bibinfo{booktitle}{\emph{CVPR}}. \bibinfo{pages}{8317--8326}.
\newblock


\bibitem[Su et~al\mbox{.}(2021)]%
        {su2021roformer}
\bibfield{author}{\bibinfo{person}{Jianlin Su}, \bibinfo{person}{Yu Lu}, \bibinfo{person}{Shengfeng Pan}, \bibinfo{person}{Bo Wen}, {and} \bibinfo{person}{Yunfeng Liu}.} \bibinfo{year}{2021}\natexlab{}.
\newblock \bibinfo{title}{RoFormer: Enhanced Transformer with Rotary Position Embedding}.
\newblock
\newblock
\showeprint[arxiv]{2104.09864}~[cs.CL]


\bibitem[Tan et~al\mbox{.}(2024a)]%
        {tan2024mobilequant}
\bibfield{author}{\bibinfo{person}{Fuwen Tan}, \bibinfo{person}{Royson Lee}, \bibinfo{person}{{\L}ukasz Dudziak}, \bibinfo{person}{Shell~Xu Hu}, \bibinfo{person}{Sourav Bhattacharya}, \bibinfo{person}{Timothy Hospedales}, \bibinfo{person}{Georgios Tzimiropoulos}, {and} \bibinfo{person}{Brais Martinez}.} \bibinfo{year}{2024}\natexlab{a}.
\newblock \showarticletitle{Mobilequant: Mobile-friendly quantization for on-device language models}.
\newblock \bibinfo{journal}{\emph{arXiv preprint arXiv:2408.13933}} (\bibinfo{year}{2024}).
\newblock


\bibitem[Tan et~al\mbox{.}(2024b)]%
        {2024_mobilequant}
\bibfield{author}{\bibinfo{person}{Fuwen Tan}, \bibinfo{person}{Royson Lee}, \bibinfo{person}{Łukasz Dudziak}, \bibinfo{person}{Shell~Xu Hu}, \bibinfo{person}{Sourav Bhattacharya}, \bibinfo{person}{Timothy Hospedales}, \bibinfo{person}{Georgios Tzimiropoulos}, {and} \bibinfo{person}{Brais Martinezs}.} \bibinfo{year}{2024}\natexlab{b}.
\newblock \showarticletitle{MobileQuant: Mobile-friendly Quantization for On-device Language Models}. In \bibinfo{booktitle}{\emph{The 2024 Conference on Empirical Methods in Natural Language Processing}}.
\newblock
\urldef\tempurl%
\url{https://openreview.net/forum?id=48ptWWA54E}
\showURL{%
\tempurl}


\bibitem[Touvron et~al\mbox{.}(2023a)]%
        {touvron2023llama}
\bibfield{author}{\bibinfo{person}{Hugo Touvron}, \bibinfo{person}{Thibaut Lavril}, \bibinfo{person}{Gautier Izacard}, \bibinfo{person}{Xavier Martinet}, \bibinfo{person}{Marie-Anne Lachaux}, \bibinfo{person}{Timoth{\'e}e Lacroix}, \bibinfo{person}{Baptiste Rozi{\`e}re}, \bibinfo{person}{Naman Goyal}, \bibinfo{person}{Eric Hambro}, \bibinfo{person}{Faisal Azhar}, {et~al\mbox{.}}} \bibinfo{year}{2023}\natexlab{a}.
\newblock \showarticletitle{Llama: Open and efficient foundation language models}.
\newblock \bibinfo{journal}{\emph{arXiv preprint arXiv:2302.13971}} (\bibinfo{year}{2023}).
\newblock


\bibitem[Touvron et~al\mbox{.}(2023b)]%
        {llama2}
\bibfield{author}{\bibinfo{person}{Hugo Touvron}, \bibinfo{person}{Louis Martin}, \bibinfo{person}{Kevin Stone}, \bibinfo{person}{Peter Albert}, \bibinfo{person}{Amjad Almahairi}, \bibinfo{person}{Yasmine Babaei}, \bibinfo{person}{Nikolay Bashlykov}, \bibinfo{person}{Soumya Batra}, \bibinfo{person}{Prajjwal Bhargava}, \bibinfo{person}{Shruti Bhosale}, {et~al\mbox{.}}} \bibinfo{year}{2023}\natexlab{b}.
\newblock \showarticletitle{{Llama 2}: Open foundation and fine-tuned chat models}.
\newblock \bibinfo{journal}{\emph{arXiv preprint arXiv:2307.09288}} (\bibinfo{year}{2023}).
\newblock


\bibitem[Tseng et~al\mbox{.}(2024)]%
        {tseng2024quip+}
\bibfield{author}{\bibinfo{person}{Albert Tseng}, \bibinfo{person}{Jerry Chee}, \bibinfo{person}{Qingyao Sun}, \bibinfo{person}{Volodymyr Kuleshov}, {and} \bibinfo{person}{Christopher De~Sa}.} \bibinfo{year}{2024}\natexlab{}.
\newblock \showarticletitle{Quip\#: Even better LLM quantization with hadamard incoherence and lattice codebooks}.
\newblock \bibinfo{journal}{\emph{Forty-first International Conference on Machine Learning}} (\bibinfo{year}{2024}).
\newblock


\bibitem[Vaswani et~al\mbox{.}(2017)]%
        {vaswani2017attention}
\bibfield{author}{\bibinfo{person}{Ashish Vaswani}, \bibinfo{person}{Noam Shazeer}, \bibinfo{person}{Niki Parmar}, \bibinfo{person}{Jakob Uszkoreit}, \bibinfo{person}{Llion Jones}, \bibinfo{person}{Aidan~N Gomez}, \bibinfo{person}{{\L}ukasz Kaiser}, {and} \bibinfo{person}{Illia Polosukhin}.} \bibinfo{year}{2017}\natexlab{}.
\newblock \showarticletitle{Attention is all you need}.
\newblock \bibinfo{journal}{\emph{Advances in neural information processing systems}} (\bibinfo{year}{2017}).
\newblock


\bibitem[Wang et~al\mbox{.}(2024b)]%
        {qvlm}
\bibfield{author}{\bibinfo{person}{Changyuan Wang}, \bibinfo{person}{Ziwei Wang}, \bibinfo{person}{Xiuwei Xu}, \bibinfo{person}{Yansong Tang}, \bibinfo{person}{Jie Zhou}, {and} \bibinfo{person}{Jiwen Lu}.} \bibinfo{year}{2024}\natexlab{b}.
\newblock \showarticletitle{Q-{VLM}: Post-training Quantization for Large Vision-Language Models}. In \bibinfo{booktitle}{\emph{The Thirty-eighth Annual Conference on Neural Information Processing Systems}}.
\newblock


\bibitem[Wang et~al\mbox{.}(2024a)]%
        {Qwen2VL}
\bibfield{author}{\bibinfo{person}{Peng Wang}, \bibinfo{person}{Shuai Bai}, \bibinfo{person}{Sinan Tan}, \bibinfo{person}{Shijie Wang}, \bibinfo{person}{Zhihao Fan}, \bibinfo{person}{Jinze Bai}, \bibinfo{person}{Keqin Chen}, \bibinfo{person}{Xuejing Liu}, \bibinfo{person}{Jialin Wang}, \bibinfo{person}{Wenbin Ge}, \bibinfo{person}{Yang Fan}, \bibinfo{person}{Kai Dang}, \bibinfo{person}{Mengfei Du}, \bibinfo{person}{Xuancheng Ren}, \bibinfo{person}{Rui Men}, \bibinfo{person}{Dayiheng Liu}, \bibinfo{person}{Chang Zhou}, \bibinfo{person}{Jingren Zhou}, {and} \bibinfo{person}{Junyang Lin}.} \bibinfo{year}{2024}\natexlab{a}.
\newblock \showarticletitle{Qwen2-VL: Enhancing Vision-Language Model's Perception of the World at Any Resolution}.
\newblock \bibinfo{journal}{\emph{arXiv preprint arXiv:2409.12191}} (\bibinfo{year}{2024}).
\newblock


\bibitem[Wang et~al\mbox{.}(2023)]%
        {wang2023cogvlm}
\bibfield{author}{\bibinfo{person}{Weihan Wang}, \bibinfo{person}{Qingsong Lv}, \bibinfo{person}{Wenmeng Yu}, \bibinfo{person}{Wenyi Hong}, \bibinfo{person}{Ji Qi}, \bibinfo{person}{Yan Wang}, \bibinfo{person}{Junhui Ji}, \bibinfo{person}{Zhuoyi Yang}, \bibinfo{person}{Lei Zhao}, \bibinfo{person}{Xixuan Song}, {et~al\mbox{.}}} \bibinfo{year}{2023}\natexlab{}.
\newblock \showarticletitle{{CogVLM}: Visual expert for pretrained language models}.
\newblock \bibinfo{journal}{\emph{arXiv preprint arXiv:2311.03079}} (\bibinfo{year}{2023}).
\newblock


\bibitem[Wei et~al\mbox{.}(2022)]%
        {wei2022outlier}
\bibfield{author}{\bibinfo{person}{Xiuying Wei}, \bibinfo{person}{Yunchen Zhang}, \bibinfo{person}{Xiangguo Zhang}, \bibinfo{person}{Ruihao Gong}, \bibinfo{person}{Shanghang Zhang}, \bibinfo{person}{Qi Zhang}, \bibinfo{person}{Fengwei Yu}, {and} \bibinfo{person}{Xianglong Liu}.} \bibinfo{year}{2022}\natexlab{}.
\newblock \showarticletitle{Outlier suppression: Pushing the limit of low-bit transformer language models}.
\newblock \bibinfo{journal}{\emph{Advances in Neural Information Processing Systems}} (\bibinfo{year}{2022}).
\newblock


\bibitem[Xiao et~al\mbox{.}(2022)]%
        {xiao2022smoothquant}
\bibfield{author}{\bibinfo{person}{Guangxuan Xiao}, \bibinfo{person}{Ji Lin}, \bibinfo{person}{Mickael Seznec}, \bibinfo{person}{Julien Demouth}, {and} \bibinfo{person}{Song Han}.} \bibinfo{year}{2022}\natexlab{}.
\newblock \showarticletitle{Smoothquant: Accurate and efficient post-training quantization for large language models}.
\newblock \bibinfo{journal}{\emph{arXiv preprint arXiv:2211.10438}} (\bibinfo{year}{2022}).
\newblock


\bibitem[Xie et~al\mbox{.}(2024)]%
        {xie2024advancing}
\bibfield{author}{\bibinfo{person}{Jingjing Xie}, \bibinfo{person}{Yuxin Zhang}, \bibinfo{person}{Mingbao Lin}, \bibinfo{person}{Liujuan Cao}, {and} \bibinfo{person}{Rongrong Ji}.} \bibinfo{year}{2024}\natexlab{}.
\newblock \showarticletitle{Advancing multimodal large language models with quantization-aware scale learning for efficient adaptation}. In \bibinfo{booktitle}{\emph{Proceedings of the 32nd ACM International Conference on Multimedia}}. \bibinfo{pages}{10582--10591}.
\newblock


\bibitem[Xu et~al\mbox{.}({[n.\,d.]})]%
        {yuerwkvquant}
\bibfield{author}{\bibinfo{person}{Chen Xu}, \bibinfo{person}{Yuxuan Yue}, \bibinfo{person}{Zukang Xu}, \bibinfo{person}{Xing Hu}, \bibinfo{person}{Zhixuan Chen}, \bibinfo{person}{Sifan Zhou}, \bibinfo{person}{Zhihang Yuan}, \bibinfo{person}{Dawei Yang}, {et~al\mbox{.}}} \bibinfo{year}{[n.\,d.]}\natexlab{}.
\newblock \showarticletitle{RWKVQuant: Quantizing the RWKV Family with Proxy Guided Hybrid of Scalar and Vector Quantization}. In \bibinfo{booktitle}{\emph{Forty-second International Conference on Machine Learning}}.
\newblock


\bibitem[Xu et~al\mbox{.}(2025)]%
        {xumambaquant}
\bibfield{author}{\bibinfo{person}{Zukang Xu}, \bibinfo{person}{Yuxuan Yue}, \bibinfo{person}{Xing Hu}, \bibinfo{person}{Dawei Yang}, \bibinfo{person}{Zhihang Yuan}, \bibinfo{person}{Zixu Jiang}, \bibinfo{person}{Zhixuan Chen}, \bibinfo{person}{Sifan Zhou}, {et~al\mbox{.}}} \bibinfo{year}{2025}\natexlab{}.
\newblock \showarticletitle{MambaQuant: Quantizing the Mamba Family with Variance Aligned Rotation Methods}.
\newblock \bibinfo{journal}{\emph{The Thirteenth International Conference on Learning Representations}} (\bibinfo{year}{2025}).
\newblock


\bibitem[Yao et~al\mbox{.}(2024)]%
        {yao2024minicpmv}
\bibfield{author}{\bibinfo{person}{Yuan Yao}, \bibinfo{person}{Tianyu Yu}, \bibinfo{person}{Ao Zhang}, \bibinfo{person}{Chongyi Wang}, \bibinfo{person}{Junbo Cui}, \bibinfo{person}{Hongji Zhu}, \bibinfo{person}{Tianchi Cai}, \bibinfo{person}{Haoyu Li}, \bibinfo{person}{Weilin Zhao}, \bibinfo{person}{Zhihui He}, {et~al\mbox{.}}} \bibinfo{year}{2024}\natexlab{}.
\newblock \showarticletitle{Minicpm-v: A gpt-4v level mllm on your phone}.
\newblock \bibinfo{journal}{\emph{arXiv preprint arXiv:2408.01800}} (\bibinfo{year}{2024}).
\newblock


\bibitem[Yu et~al\mbox{.}(2025)]%
        {yu2025q}
\bibfield{author}{\bibinfo{person}{Jiangyong Yu}, \bibinfo{person}{Changyong Shu}, \bibinfo{person}{Dawei Yang}, \bibinfo{person}{Sifan Zhou}, \bibinfo{person}{Zichen Yu}, \bibinfo{person}{Xing Hu}, {and} \bibinfo{person}{Yan Chen}.} \bibinfo{year}{2025}\natexlab{}.
\newblock \showarticletitle{Q-PETR: Quant-aware Position Embedding Transformation for Multi-View 3D Object Detection}.
\newblock \bibinfo{journal}{\emph{arXiv preprint arXiv:2502.15488}} (\bibinfo{year}{2025}).
\newblock


\bibitem[Yuan et~al\mbox{.}(2023a)]%
        {yuan2023rptq}
\bibfield{author}{\bibinfo{person}{Zhihang Yuan}, \bibinfo{person}{Lin Niu}, \bibinfo{person}{Jiawei Liu}, \bibinfo{person}{Wenyu Liu}, \bibinfo{person}{Xinggang Wang}, \bibinfo{person}{Yuzhang Shang}, \bibinfo{person}{Guangyu Sun}, \bibinfo{person}{Qiang Wu}, \bibinfo{person}{Jiaxiang Wu}, {and} \bibinfo{person}{Bingzhe Wu}.} \bibinfo{year}{2023}\natexlab{a}.
\newblock \showarticletitle{RPTQ: Reorder-based Post-training Quantization for Large Language Models}.
\newblock \bibinfo{journal}{\emph{arXiv preprint arXiv:2304.01089}} (\bibinfo{year}{2023}).
\newblock


\bibitem[Yuan et~al\mbox{.}(2023b)]%
        {yuan2023asvd}
\bibfield{author}{\bibinfo{person}{Zhihang Yuan}, \bibinfo{person}{Yuzhang Shang}, \bibinfo{person}{Yue Song}, \bibinfo{person}{Qiang Wu}, \bibinfo{person}{Yan Yan}, {and} \bibinfo{person}{Guangyu Sun}.} \bibinfo{year}{2023}\natexlab{b}.
\newblock \showarticletitle{ASVD: Activation-aware Singular Value Decomposition for Compressing Large Language Models}.
\newblock \bibinfo{journal}{\emph{arXiv preprint arXiv:2312.05821}} (\bibinfo{year}{2023}).
\newblock


\bibitem[Yuan et~al\mbox{.}(2024)]%
        {yuan2024llm}
\bibfield{author}{\bibinfo{person}{Zhihang Yuan}, \bibinfo{person}{Yuzhang Shang}, \bibinfo{person}{Yang Zhou}, \bibinfo{person}{Zhen Dong}, \bibinfo{person}{Chenhao Xue}, \bibinfo{person}{Bingzhe Wu}, \bibinfo{person}{Zhikai Li}, \bibinfo{person}{Qingyi Gu}, \bibinfo{person}{Yong~Jae Lee}, \bibinfo{person}{Yan Yan}, {et~al\mbox{.}}} \bibinfo{year}{2024}\natexlab{}.
\newblock \showarticletitle{Llm inference unveiled: Survey and roofline model insights}.
\newblock \bibinfo{journal}{\emph{arXiv preprint arXiv:2402.16363}} (\bibinfo{year}{2024}).
\newblock


\bibitem[Yue et~al\mbox{.}(2024)]%
        {yue2024wkvquant}
\bibfield{author}{\bibinfo{person}{Yuxuan Yue}, \bibinfo{person}{Zhihang Yuan}, \bibinfo{person}{Haojie Duanmu}, \bibinfo{person}{Sifan Zhou}, \bibinfo{person}{Jianlong Wu}, {and} \bibinfo{person}{Liqiang Nie}.} \bibinfo{year}{2024}\natexlab{}.
\newblock \showarticletitle{Wkvquant: Quantizing weight and key/value cache for large language models gains more}.
\newblock \bibinfo{journal}{\emph{arXiv preprint arXiv:2402.12065}}.
\newblock


\bibitem[Zhang et~al\mbox{.}(2019)]%
        {RMSNorm}
\bibfield{author}{\bibinfo{person}{Biao Zhang} {et~al\mbox{.}}} \bibinfo{year}{2019}\natexlab{}.
\newblock \showarticletitle{Root mean square layer normalization}.
\newblock \bibinfo{journal}{\emph{Advances in Neural Information Processing Systems}}  \bibinfo{volume}{32} (\bibinfo{year}{2019}).
\newblock


\bibitem[Zhang et~al\mbox{.}(2023)]%
        {zhang2023video}
\bibfield{author}{\bibinfo{person}{Hang Zhang}, \bibinfo{person}{Xin Li}, {and} \bibinfo{person}{Lidong Bing}.} \bibinfo{year}{2023}\natexlab{}.
\newblock \showarticletitle{Video-LLaMA: An Instruction-tuned Audio-Visual Language Model for Video Understanding}. In \bibinfo{booktitle}{\emph{EMNLP (Demos)}}.
\newblock


\bibitem[Zhang et~al\mbox{.}(2024)]%
        {zhang2024qqq}
\bibfield{author}{\bibinfo{person}{Ying Zhang}, \bibinfo{person}{Peng Zhang}, \bibinfo{person}{Mincong Huang}, \bibinfo{person}{Jingyang Xiang}, \bibinfo{person}{Yujie Wang}, \bibinfo{person}{Chao Wang}, \bibinfo{person}{Yineng Zhang}, \bibinfo{person}{Lei Yu}, \bibinfo{person}{Chuan Liu}, {and} \bibinfo{person}{Wei Lin}.} \bibinfo{year}{2024}\natexlab{}.
\newblock \showarticletitle{QQQ: Quality Quattuor-Bit Quantization for Large Language Models}.
\newblock \bibinfo{journal}{\emph{arXiv preprint arXiv:2406.09904}} (\bibinfo{year}{2024}).
\newblock


\bibitem[Zhou et~al\mbox{.}(2024)]%
        {zhou2024lidar}
\bibfield{author}{\bibinfo{person}{Sifan Zhou}, \bibinfo{person}{Liang Li}, \bibinfo{person}{Xinyu Zhang}, \bibinfo{person}{Bo Zhang}, \bibinfo{person}{Shipeng Bai}, \bibinfo{person}{Miao Sun}, \bibinfo{person}{Ziyu Zhao}, \bibinfo{person}{Xiaobo Lu}, {and} \bibinfo{person}{Xiangxiang Chu}.} \bibinfo{year}{2024}\natexlab{}.
\newblock \showarticletitle{LiDAR-PTQ: Post-Training Quantization for Point Cloud 3D Object Detection}.
\newblock \bibinfo{journal}{\emph{International Conference on Learning Representations}} (\bibinfo{year}{2024}).
\newblock


\bibitem[Zhou et~al\mbox{.}(2025)]%
        {gsq}
\bibfield{author}{\bibinfo{person}{Sifan Zhou}, \bibinfo{person}{Shuo Wang}, \bibinfo{person}{Zhihang Yuan}, \bibinfo{person}{Mingjia Shi}, \bibinfo{person}{Yuzhang Shang}, {and} \bibinfo{person}{Dawei Yang}.} \bibinfo{year}{2025}\natexlab{}.
\newblock \showarticletitle{{GSQ}-Tuning: Group-Shared Exponents Integer in Fully Quantized Training for {LLM}s On-Device Fine-tuning}. In \bibinfo{booktitle}{\emph{Findings of the Association for Computational Linguistics: ACL 2025}}. \bibinfo{publisher}{Association for Computational Linguistics}, \bibinfo{address}{Vienna, Austria}, \bibinfo{pages}{22971--22988}.
\newblock
\showISBNx{979-8-89176-256-5}
\urldef\tempurl%
\url{https://aclanthology.org/2025.findings-acl.1178/}
\showURL{%
\tempurl}


\bibitem[Zhu et~al\mbox{.}(2023)]%
        {zhu2023minigpt}
\bibfield{author}{\bibinfo{person}{Deyao Zhu}, \bibinfo{person}{Jun Chen}, \bibinfo{person}{Xiaoqian Shen}, \bibinfo{person}{Xiang Li}, {and} \bibinfo{person}{Mohamed Elhoseiny}.} \bibinfo{year}{2023}\natexlab{}.
\newblock \showarticletitle{{MiniGPT-4}: Enhancing Vision-Language Understanding with Advanced Large Language Models}.
\newblock \bibinfo{journal}{\emph{arXiv preprint arXiv:2304.10592}} (\bibinfo{year}{2023}).
\newblock


\end{thebibliography}
\appendix
\section{Appendix}

\subsection{MQuant Algorithm}
\label{Mquant}
Here, we present overall \emph{MQuant} algorithm for MLLMs in Algorithm~\ref{alg:MQuant}.
\begin{algorithm}[!htbp]
\caption{\emph{MQuant} Quantization Algorithm}
\label{alg:MQuant}
\textbf{Input}: Full-precision (FP) MLLM model with a vision encoder $\mE$, visual-language projector $\mP$, and a large language model $\textbf{LLM}$; Calibration dataset $D^c$.\\
\textbf{Output}:
\begin{itemize}
\item For $\mE$ and $\mP$: per-channel weight scale $s^E_{w}$, per-channel weight zero-point $z^E_{w}$, per-tensor activation scales $s^E_{a}$, per-tensor activation zero-point $z^E_{a}$.
\item For $\textbf{LLM}$: per-channel weight scale $s^{llm}_{w}$, per-channel weight zero-point $z^{llm}_{w}$, per-tensor activation scales $s^{^{llm}}_{a^v}$ for visual tokens and $s^{llm}_{a^t}$ for textual tokens, per-tensor activation zero-points $z^{llm}_{a^v}$ for visual tokens and $z^{llm}_{a^s}$ for textual tokens.
\end{itemize}
\begin{algorithmic}[1]
\STATE Apply Hadamard Rotation to the LLM Part as described:
\STATE Apply the offline and online Hadamard rotations to all the weights and activations in $\textbf{LLM}$.
\STATE Quantize Weights of the LLM:
\STATE Input the calibration dataset $D^c$ to the FP MLLM.
\STATE Use GPTQ to quantize the weights for $\textbf{LLM}$, obtaining per-channel weight quantization parameters $s^{llm}_{w}$ and $z^{llm}_{w}$.
\STATE For the LLM Part:
\begin{enumerate}[label=(\alph*)]
\item Input $D^c$ to the FP MLLM.
\item Compute per-tensor activation quantization parameters $s^{^{llm}}_{a^v}$ and $s^{llm}_{a^t}$, $z^{llm}_{a^v}$ and $z^{llm}_{a^t}$ for visual and textual tokens respectively, based on the proposed Modality-Specific Static Quantization (MSQ) in Sec 3.1.
\item Reorder the input mixed token sequence from $E = {e^t_1, \dots, e^v_m, \dots, e^v_n, \dots, e^t_L}$ to a unified modality-decoupled sequence $E^u = {e^v_m, \dots, e^v_n, e^t_1, \dots, e^t_{m-1}, \dots, e^t_{n+1}, \dots, e^t_L}$ using the proposed Attention-Invariant Flexible Switching (AIFS) scheme in Sec 3.1.
\end{enumerate}
\STATE Transform all the LayerNorm to RMSNorm in MLLM vision encoder $\mE$ and Visual-Language Projector $\mP$ using the proposed Post LayerNorm-to-RMSNorm transformation.
\STATE Apply the offline and online Hadamard rotations to all the weights and activations in $\mE$ and $\mP$.
\STATE Quantize Weights of $\mE$ and $\mP$:
\STATE Input $D^c$ to the transformed FP vision encoder $\mE$ and Visual-Language Projector $\mP$.
\STATE Use GPTQ to quantize $\mE$ and $\mP$, obtaining per-channel weight quantization parameters $s^E_{w}$ and $z^E_{w}$.
\STATE Address the weight outliers using caused by online Hadamard based on the proposed Rotation Magnitude Suppression (RMS) in Sec 3.2.
\end{algorithmic}
\end{algorithm}

\subsection{Rotary Position Embedding for Attention-Invariant Flexible Switching}
\label{rope}
Many modern LLMs~\citep{touvron2023llama,llama2,llama3} use rotary position embedding (RoPE)~\citep{su2021roformer} to encode information about the order of tokens in the input sequence. Rotary position embeddings are linear transformations applied to keys and queries defined as: 
\vspace{-2mm}
\begin{equation}
R_{\Theta,m}^{d_h} =
	\begin{pmatrix}
		\cos{i\theta_1}& -\sin{i\theta_1} &\cdots & 0 &0\\
		\sin{i\theta_1}&\cos{i\theta_1}   &\cdots & 0 &0 \\
  %           0 &0 &\cos{i\theta_2}& -\sin{i\theta_2} &\cdots & 0 &0\\
		% 0 & 0 &\sin{i\theta_2}&\cos{i\theta_2}   &\cdots & 0 &0 \\
		\vdots &\vdots &\ddots &\vdots &\vdots &\vdots &\vdots\\
    		0 & 0  &\cdots&\cos{i\theta_{d_h/2}}& -\sin{i\theta_{d_h/2}}\\
		 0 &0 &\cdots&\sin{i\theta_{d_h/2}}&\cos{i\theta_{d_h/2}}
	\end{pmatrix}
    \label{algo:rope}
\end{equation}
where $i \in [1, L]$ is the token index, $\Theta = \{\theta_i=10000^{-2(i-1)/D}, i \in [1, 2, ..., D/2]\}$, and $\theta_i,\, i \in 1..D/2$ are predefined constants.

In the proposed Attention-Invariant Flexible Switching (AIFS) mechanism, we also apply the rearrangement for position embedding to maintain the computation equivalence. For a mixed input token $E = \left\{e^t_1,...,e^v_m,...,e^v_n,...,e^t_L) \in (\mE_v,\mE_t)\right\}$ (in Eq. 3), where $m$ and $n$ denote the start and end indices of the visual tokens. Specifically, after AIFS, the unified token can formulated as: $E = \left\{e^v_m,...,e^v_n,e^t_1,...,e^t_{m-1},...,e^t_L) \in (\mE_v,\mE_t)\right\}$. Therefore, the unified token indices after AIFS can be represented as:
\begin{equation}
(m,...,n,1,...,{m-1},{n+1},...,L) = AIFS{(1,...,m,...,n,...,L)}
\end{equation}
Due to we are aware of the final indices for input token after AIFS, than we can utilize the reorder token indices for Eq~\ref{algo:rope} to get the corresponding position embedding.

\subsection{Weights Outliers after Online Hadamard Transformation} 
\label{weight_outliers}
Following~\citep{tseng2024quip+,ashkboos2024quarot} we make use of fast Hadamard transformation where convenient. Hadamard matrix is an orthogonal matrix with entries proportional to $\{+1, -1\}$. A Walsh-Hadamard matrix is a square matrix of size $2^n$ with For a Hadamard matrix:
\begin{align}
    \mH_2 = \tfrac{1}{\sqrt{2}}\left[\begin{array}{cc}
    1&1\\1&-1\end{array}\right]
    \qquad\textrm{and} \qquad \mH_{2^n} = \mH_2 \otimes \mH_{2^{n-1}}\,. \\
    (AB)_{ij} = \sum_{r=1}^{n} a_{ir} b_{rj} = a_{i1} b_{1j} + a_{i2} b_{2j} + \cdots + a_{in} b_{nj}
\vspace{-4mm}
\end{align}
Thereby the $\mH W_{\ell_2}$ can be formulated as:
\vspace{-2mm}
\begin{align}
    (\mH W_{\ell_2})_{ij} = \sum_{r=1}^{n} h_{ir} w_{rj} = h_{i1} w_{1j} + h_{i2} w_{2j} + \cdots + h_{in} w_{nj}
\vspace{-4mm}
\end{align}
where $\mH\in\mathbb{R}^{d_{in} \times d_{in}}$ and $W_{\ell_2} \in\mathbb{R}^{d_{in} \times d_{out}}$, $d_{in}$ and $d_{out}$ is the dimension of input and output of weight $W_{\ell_2}$. Due to the properties of the Hadamard matrix $\mH$, whose first row consists entirely of $1$, for the first row in $(\mH W_{\ell_2})$, $(\mH W_{\ell_2})_{0j} = \sum_{r=1}^{n} w_{rj}$, due to the property of Hadamard matrix $\mH$. So, the values in $W_{\ell_2}$ are subject to continuous accumulation and summation, resulting in the exists of outliers in the first row of the output matrix $\mH W_{\ell_2}$. Notably, the matrix $\mH W_{\ell_2}\mQ$ still has the same problem, for simplicity, we omit the matrix $\mQ$ in the main paper. 

\subsection{Attention Mask in AIFS when Multi-batch Inference.}
\label{multi-batch mask}

\begin{figure}[!htbp]
    \centering    
    \includegraphics[width=1.0\linewidth]{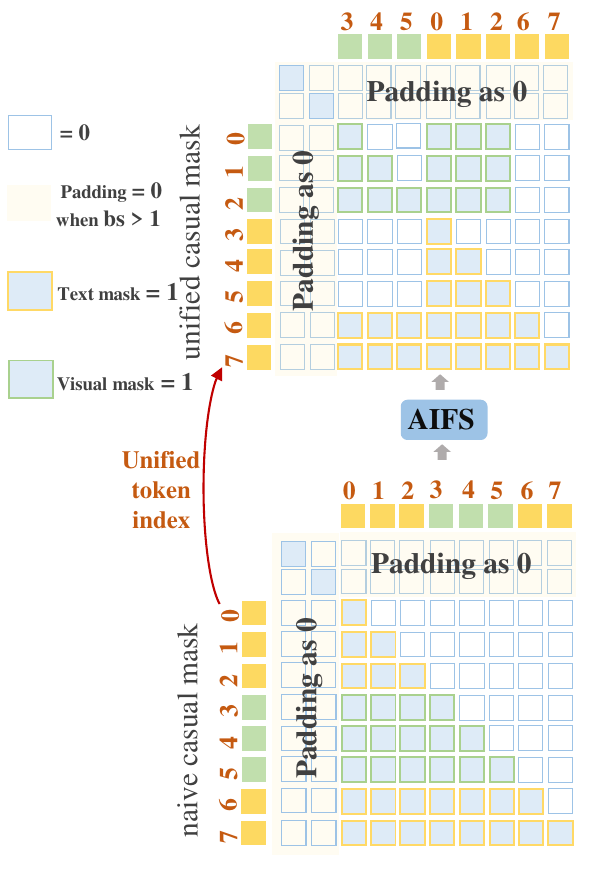}
    \vspace{-2mm}
    \caption{The illustration of causal mask when batch size > 1.} 
    \label{mask_mutli-batch}
\vspace{-3mm}
\end{figure}

% \begin{wrapfigure}{r}{5.5cm}
%   % \vspace{-13mm}
%   % \begin{adjustbox}{max width=\linewidth}
%     \includegraphics[width=1.0\linewidth]{figures/AIFSV3_pad.pdf}
%     \caption{The illustration of causal mask when batch size > 1.}
%   \label{mask_mutli-batch}
%   \vspace{-8mm}
% \end{wrapfigure}
During multi-batch inference, we first identify the longest token length within the batch. Subsequently, we left-pad the shorter sequences with $pad_token_id$ to align all batches to this maximum length. By applying left padding, the padding tokens are associated with the image modality. Additionally, the padded regions are assigned a mask value of 0, ensuring that they do not interfere with attention computations and thereby do not affect the final results. For clarity, we also plot an illustration of causal mask when batch size >1.

\subsection{Effectiveness of Rotational Magnitude Suppression for Weight Outliers in LLMs.}
\label{RMS4LLMs}

\begin{table}[h]
\caption{WikiText-2 perplexity (PPL) on 4-bit quantized LLaMA2 models with an input sequence length of 2048. Lower PPL is better.}
\label{table:quarot+rms}
\centering
\resizebox{0.8\linewidth}{!}{
\begin{tabular}{l|c|ccc}
\toprule
\multirow{2}{*}{\textbf{Method}} & \multirow{2}{*}{\textbf{Weight Quantization}} & \multicolumn{3}{c}{\textbf{PPL}} \\
\cmidrule(lr){3-5}
 & & \textbf{LLaMA2 7B} & \textbf{LLaMA2 13B} & \textbf{LLaMA2 70B} \\
\midrule
Baseline & --    & 5.47 & 4.88 & 3.32 \\
QuaRot   & GPTQ  & 6.10 & 5.40 & 3.79 \\
\rowcolor{blue!10}
QuaRot + RMS & GPTQ & 6.04 & 5.32 & 3.67 \\
\bottomrule
\end{tabular}
}
\end{table}

For LLMs, compared to the original Quarot, integrating RMS with Quarot leads to performance improvements across LLaMA2 models with 7B, 13B, and 70B parameters, as detailed in Table. 4. For MLLMs, ablation studies presented in Table~\ref{table:ablation} demonstrate that the RMS method significantly enhances quantization performance.

\begin{figure*}[h]
    \centering    
    \includegraphics[width=1.0\textwidth]{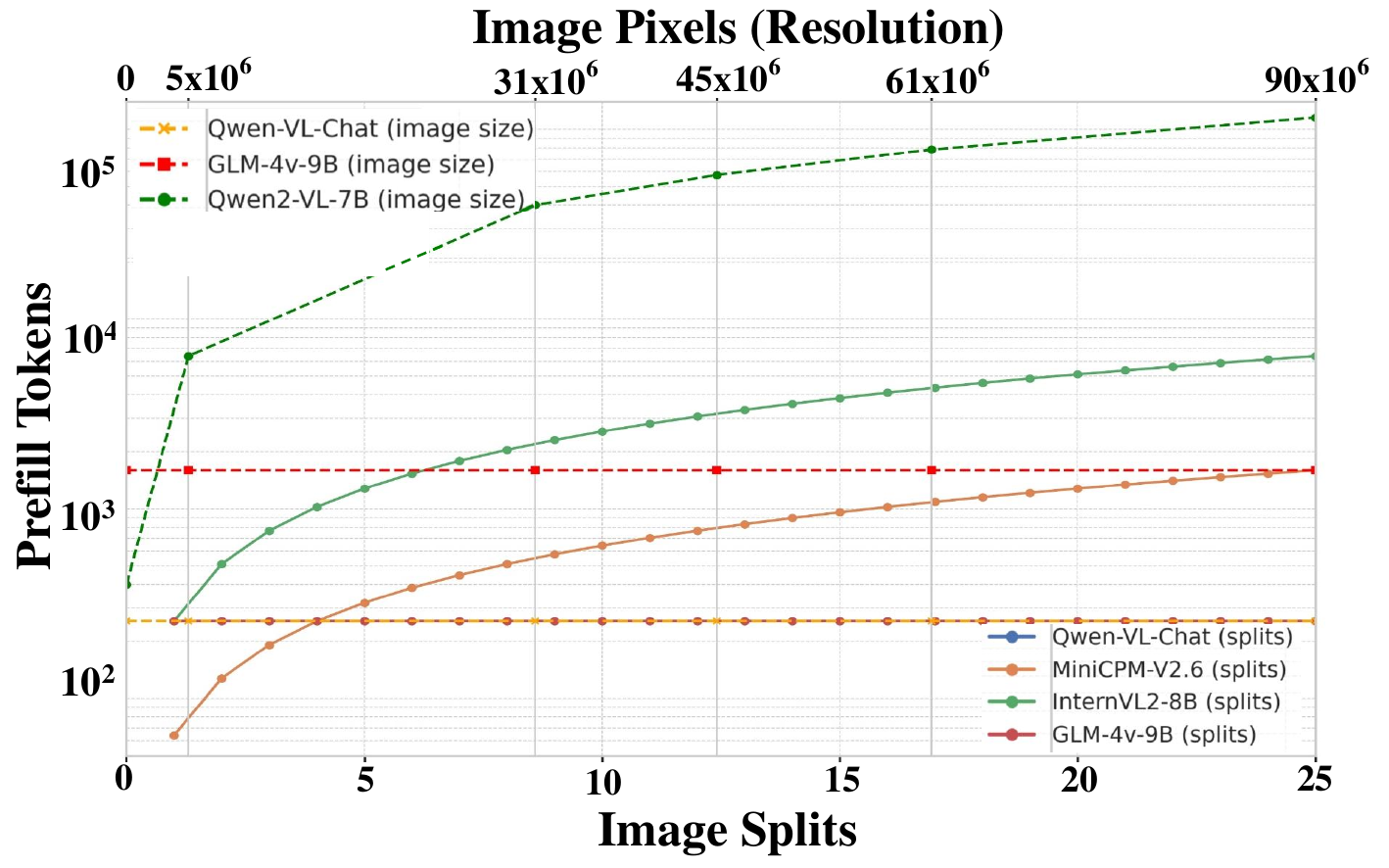}
    \vspace{-3mm}
    \caption{The number of prefill visual tokens across different MLLMs as the image splits or resolution increases.} 
    \label{fig:obs_reso}
\vspace{-3mm}
\end{figure*}

\subsection{Image Tokens in Various MLLMs} 
As shown in Fig \ref{fig:obs_reso}, for different MLLMs~\citep{qwenvl,Qwen2VL,yao2024minicpmv,chen2024internvl}, the number of prefill visual tokens grows as image resolution increases. This rapid expansion in token count exacerbates the inference latency, particularly in per-token dynamic quantization, which requires separate processing for each token, such as memory access and scale computation. As a result, the TTFT in MLLMs increases drastically, severely impacting overall inference latency. Moreover, in higher-demand scenarios such as video-based tasks and multi-image dialogues, the accumulation of visual tokens becomes even more pronounced, further exacerbating the increase in TTFT.

\subsection{RMS Algorithm}
\label{RMS_algo}
Here, we present the algorithm of our \emph{RMS} design with Quarot in Algorithm~\ref{alg:rms+quarot}.

\begin{algorithm}
\small
\caption{RMS Integration with Quarot}
\label{alg:rms+quarot}
\begin{algorithmic}[1]
    \REQUIRE An LLM or MLLM model
    \ENSURE Quantized model with RMS
    
    \STATE Initialize an empty list $marks$
    \FOR{each linear layer in the $model$}
        \IF{Layer satisfies Eq. 9}
            \STATE Mark the layer and append its ID to $marks$
        \ENDIF
    \ENDFOR
    \STATE Apply Quarot or other Hadamard-based transformations to the $model$
    \FOR{each layer ID in $marks$}
        \STATE Modify the layer's implementation using the RMS method
    \ENDFOR
    \STATE Quantize the $model$
    \STATE \textbf{return} $model$
\end{algorithmic}
\end{algorithm}

\begin{table*}[h]
    \caption{Comparison of latency and memory saving with Pytorch and AWQ on Qwen2-VL-7B. $^{\ddag}$ means the Qwen2-VL-7B official implementation.}
    \vspace{-4.5mm}
    \label{tab:ablation2}
    \begin{adjustbox}{max width=0.8\linewidth}
        \begin{tabular}{ccc|cc|cc}\\
        \toprule 
        Image size & \multicolumn{2}{c|}{Pytorch$^{\ddag}$ (BF16)}& \multicolumn{2}{c|}{AWQ$^{\ddag}$ (W4-only)}& \multicolumn{2}{c}{MQuant (W4A8)} \\ \cline{2-7}
         H $\times$ W & Latency(s)& Memory(G) & Latency(s)& Memory(G)&Latency(s)& Memory(G) \\
        \midrule        
        280$^2$ & 0.257 &16.45 &0.286 (\color{myred}{-10.14\%}) &7.45 (\color{mygreen1}{+120.67\%}) &0.220 (\textbf{\color{mygreen1}{+16.82\%}}) &6.50 (\textbf{\color{mygreen1}{+152.92\%}})\\
        560$^2$ &0.252 &16.45 &0.292 (\color{myred}{-13.70\%}) &7.45 (\color{mygreen1}{+120.67\%})&0.211 (\textbf{\color{mygreen1}{+19.48\%}}) &6.50 (\textbf{\color{mygreen1}{+152.92\%}})\\
        840$^2$ & 0.261 &16.45 &0.304 (\color{myred}{-14.14\%}) &7.45 (\color{mygreen1}{+120.67\%})&0.210 (\textbf{\color{mygreen1}{+24.76\%}})&6.50 (\textbf{\color{mygreen1}{+152.92\%}})\\
        1120$^2$ & 0.326 &16.58 &0.384 (\color{myred}{-15.10\%}) &7.56 (\color{mygreen1}{+119.51\%})&0.262 (\textbf{\color{mygreen1}{+24.48\%}})&6.61 (\textbf{\color{mygreen1}{+151.59\%}})\\
        1400$^2$ & 0.559 &16.90 &0.652 (\color{myred}{-14.29\%}) &7.90 (\color{mygreen1}{+113.92\%})&0.432 (\textbf{\color{mygreen1}{+20.24\%}})&6.97 (\textbf{\color{mygreen1}{+142.71\%}})\\
        1680$^2$ & 0.881 &17.33 &1.066 (\color{myred}{-17.39\%}) &8.33 (\color{mygreen1}{+108.57\%})&0.705 (\textbf{\color{mygreen1}{+18.23\%}})&7.40 (\textbf{\color{mygreen1}{+130.27\%}})\\
        1960$^2$ & 1.369 &17.82 &1.598 (\color{myred}{-14.26\%}) &8.82 (\color{mygreen1}{+100.00\%})&1.112 (\textbf{\color{mygreen1}{+16.63\%}})&7.85 (\textbf{\color{mygreen1}{+119.93\%}})\\
        2240$^2$ & 2.013 &18.40 &2.294 (\color{myred}{-12.24\%}) &9.40 (\color{mygreen1}{+95.74\%})&1.653 (\textbf{\color{mygreen1}{+17.83\%}})&8.44 (\textbf{\color{mygreen1}{+117.84\%}})\\
        2520$^2$ & 2.820 &19.04 &3.175 (\color{myred}{-11.14\%}) &10.05 (\color{mygreen1}{+89.59\%})&2.357 (\textbf{\color{mygreen1}{+19.63\%}})&9.10 (\textbf{\color{mygreen1}{+109.92\%}})\\
        2880$^2$ & 3.883 &19.77 &4.345 (\color{myred}{-10.64\%}) &10.77 (\color{mygreen1}{+83.81\%})&3.297 (\textbf{\color{mygreen1}{+17.69\%}})&9.82 (\textbf{\color{mygreen1}{+101.07\%}})\\
        3080$^2$ & 5.208 &20.58 &5.872 (\color{myred}{-11.27\%}) &11.58 (\color{mygreen1}{+77.60\%})&4.488 (\textbf{\color{mygreen1}{+16.02\%}})&10.61 (\textbf{\color{mygreen1}{+96.45\%}})\\
        3360$^2$ & 6.814 &21.46 &7.548 (\color{myred}{-9.73\%}) &12.46 (\color{mygreen1}{+70.65\%})&6.004 (\textbf{\color{mygreen1}{+13.41\%}})&11.50 (\textbf{\color{mygreen1}{+83.01\%}})\\
        3640$^2$ & 8.360 &22.22 &9.377 (\color{myred}{-10.84\%}) &13.22 (\color{mygreen1}{+57.54\%})&7.469 (\textbf{\color{mygreen1}{+11.91\%}})&12.25 (\textbf{\color{mygreen1}{+61.65\%}})\\
        4480$^2$ & 8.349 &22.22 &9.379 (\color{myred}{-10.97\%}) &13.22 (\color{mygreen1}{+57.54\%})&7.469 (\textbf{\color{mygreen1}{+11.71\%}})&12.25 (\textbf{\color{mygreen1}{+61.65\%}})\\
        5600$^2$ & 8.380 &22.22 &9.393 (\color{myred}{-10.78\%}) &13.22 (\color{mygreen1}{+57.54\%})&7.469 (\textbf{\color{mygreen1}{+12.19\%}})&12.25 (\textbf{\color{mygreen1}{+61.65\%}})\\
            \hline
            \end{tabular}  
    \end{adjustbox}
\label{tab:speedup_append}
\vspace{-3mm}
% \end{wraptable}
\end{table*}

\subsection{Speedup and Memory Savings with Scaling Image Resolution}
\label{Speed_mem_image}
We fixed the input sequence as "text-image-text" with 15 textual tokens and presented the detailed changes of speed and memory, varying the image resolution from $280\times280$ to $5600\times5600$. Notably, the "text-image-text" sequence setting is not arbitrarily chosen; instead, it is a common setting in existing evaluation datasets~\citep{duan2024vlmevalkit}. We evaluate speedup and memory savings by comparing PyTorch's BF16, AWQ (W4-only), and our MQuant (W4A8). \textbf{Speedup:} As shown in Table \ref{tab:speedup_append}, MQuant consistently achieves speedups over both PyTorch and AWQ across all resolutions, with a maximum of 24.76\% over PyTorch at $840\times 840$. Notably, MQuant outperforms AWQ, which is slower than PyTorch at most resolutions due to negative speedups. This significant speedup highlights the advantage of our per-tensor static quantization, eliminating the overhead of token-wise scale computation. Even at higher resolutions (e.g., $5600^2$), MQuant maintains a 12.19\% latency improvement, demonstrating scalability across various image sizes. \textbf{Memory Savings:} MQuant offers substantial memory reductions compared to both PyTorch and AWQ. It consistently reduces memory usage by over 100\% compared to PyTorch (e.g., 152.92\% at $840^2$) and significantly outperforms AWQ's memory efficiency, achieving up to 101.07\% savings at higher resolutions. These experiments demonstrate MQuant's strengths in both latency and memory savings, achieving up to 24.76\% faster inference and reducing memory consumption by over 100\% compared to baseline methods. This makes MQuant a more efficient solution for deploying MLLMs in resource-constrained environments.

\subsection{Comparison of Online vs.\ Without Online Hadamard Transform in Quarot}
\label{sec:online-hadamard}

In this section, we evaluate how online Hadamard transformations affect perplexity (PPL) in Quarot for different LLaMA2 models (7B and 13B) under two bit settings, \texttt{W4A8KV4} and \texttt{W4A4KV4}. As shown in Table~\ref{table:online-hadda}, enabling online Hadamard transforms yields consistent improvements, especially under more aggressive quantization (e.g., \texttt{W4A4KV4}).

\begin{table}[h]
\caption{Comparison of perplexity (PPL) with or without online Hadamard transforms in Quarot, evaluated on LLaMA2 models of sizes 7B and 13B. Lower PPL is better.}
\label{table:online-hadda}
\centering
\resizebox{0.8\linewidth}{!}{
\begin{tabular}{l|c|cc}
\toprule
\multirow{2}{*}{\textbf{Model}} & \multirow{2}{*}{\textbf{Bit Setting}} & \multicolumn{2}{c}{\textbf{PPL}} \\
\cmidrule(lr){3-4}
 & & With Online & Without Online \\
\midrule
LLaMA2 7B  & W4A8KV4 & 5.736 & 5.830 \\
LLaMA2 7B  & W4A4KV4 & 6.358 & 14.375 \\
LLaMA2 13B & W4A8KV4 & 5.123 & 5.146 \\
LLaMA2 13B & W4A4KV4 & 5.585 & 24.401 \\
\bottomrule
\end{tabular}
}
\vspace{-4mm}
\end{table}

We observe that online Hadamard transforms provide substantial gains when using \texttt{W4A4KV4} for both 7B and 13B models, reducing PPL from over 14.3 to 6.36 (7B) and from 24.4 to 5.58 (13B), respectively. These results demonstrate the effectiveness of online Hadamard transformations in maintaining quantization accuracy at lower bit precision.

\subsection{Advantage of Proposed MSQ and AIFS}
\label{aifs_msq_pros}
In per-tensor static quantization, the quantization parameters (i.e., scale and zero-point) are precomputed for an entire tensor (e.g., weights or activations) and remain fixed throughout inference. While efficient, this approach often leads to large and unacceptable accuracy loss in MLLMs due to their diverse activation distributions across varying inputs.

In contrast, per-token dynamic quantization computes quantization parameters on-the-fly for each input token during inference. This approach incurs significantly higher computational overhead, as the quantization parameters must be recalculated for every input token, along with multiple additional memory traversals. Such requirements make per-token dynamic quantization unfriendly or impractical for edge devices and some AI accelerators, which struggle with fine-grained dynamic operations~\cite{2024_mobilequant}. This issue is especially severe in MLLMs, where the token count increases significantly with higher image resolution or more video frames. The Modality-Specific Static Quantization (MSQ) in MQuant is a novel per-modality quantization approach specifically designed to address the unique challenges of MLLMs quantization.

Furthermore, MSQ can be naturally applied to the unified modality-decoupled tokens generated by AIFS. By integrating MSQ and AIFS, our designs yields three key advantages: \textbf{(1) Computational Equivalence and Strong Compatibility}: The unified causal mask and token index introduced by AIFS preserves the inherent causal relationships among tokens, ensuring numerical equivalence during attention computations. Moreover, since AIFS requires only a one-time rearrangement of the input data (adjust causal mask and token index in offline), it does not alter the overall computation graph. This characteristic allows for seamless integration with other LLM inference acceleration methods, such as FlashAttention~\citep{dao2022flashattention}, ensuring both computational equivalence and strong compatibility. As shown in Table~\ref{table:main_results}, MQuant achieves SOTA quantization performance across 5 mainstream MLLMs. \textbf{(2) Reduced Inference Latency}: MSQ not only addresses the substantial distributional differences between modalities but also mitigates the significant computational overhead and increased inference latency caused by the surge in token counts from higher image and video resolutions. As shown in Table~\ref{table:AIFS_laten}, MSQ+AIFS significantly reduces latency from 2.057s to 1.1017s, closely matching the speed of the per-tensor static setting while maintaining near-lossless accuracy comparable to the original Float model. \textbf{(3) Enhanced Memory and Computational Efficiency}: By combining MSQ and AIFS, we convert mixed input tokens into unified, modality-decoupled tokens, eliminating the irregular memory operations (e.g., slice, concat, pad) introduced by directly applying MSQ. This transformation reduces memory consumption and improves efficiency of \texttt{GEMM} kernel, which would otherwise be compromised by the interleaved and non-fixed positions of visual and textual tokens. As shown in Table~\ref{tab:ablation2}, MQuant can achieve up to 24.7\% speedup and 152.9\% memory savings.

\subsection{Comparison of Different MLLMs: Input Pre-Process, LayerNorm Architecture in Vision Encoder, Model Parameters and Flops.}
\label{MLLMs_comparison}

\begin{table*}[h]
\caption{Comparison of TTFT sensitivity to image resolution, model Parameters and flops in mainstream MLLMs. $^\dagger$ means the Flops values are measured with the number of visual tokens is 256.}
% \vspace{+2mm}
\label{table:ln-compare}
\centering
\resizebox{\textwidth}{!}{
\begin{tabular}{l|l|c|cc|cc}
\toprule
\multirow{2}{*}{\textbf{MLLMs}} &  \multirow{2}{*}{\textbf{TTFT's sensitivity to input image resolution}} &  \multirow{2}{*}{\textbf{LayerNorm}}& \multicolumn{2}{c|}{\textbf{Params (B)}} & \multicolumn{2}{c}{\textbf{FLOPs (T)$^\dagger$}} \\
& & &Visual & LLM &Visual & LLM \\
\toprule
\rowcolor{gray!25}
InternVL2-8B & TTFT increases with input image aspect ratio & Pre-LN & 0.34 & 7.74 &1.28 &7.54 \\
\rowcolor{gray!10}
Qwen-VL-Chat-9.6B& Fixed input resolution (448×448)  &Pre-LN &1.94 &7.73 &4.23 &3.70 \\
\rowcolor{gray!25}
MiniCPM-V 2.6-8B& TTFT increases with input image aspect ratio &Pre-LN &0.49 &7.61 &4.16 &3.64 \\
\rowcolor{yellow!20}
Qwen2-VL-7B  & TTFT increases \textbf{quadratically} with input image resolution &Pre-LN &0.68 &7.61 &1.31 &3.61 \\
\rowcolor{gray!10}
GLM-4V-9B  & Fixed input resolution (1120×1120) &Post-LN &4.51 & 8.78 &12.10 &4.70 \\  
\bottomrule
\end{tabular}
}
% \vspace{-4mm}
\end{table*}

In this section, we compare the visual input pre-process methods, LayerNorm structures in MLLM vision encoder, model parameters, and Flops across five mainstream MLLMs: InternVL2-8B~\citep{internvl15}, Qwen-VL-Chat-9.6B~\citep{qwenvl}, MiniCPM-V 2.6-8B~\citep{yao2024minicpmv}, Qwen2-VL-7B~\citep{Qwen2VL}, and GLM-4V-9B~\citep{CogVLM2}. As shown in Table~\ref{table:ln-compare}, this comparison highlights the architectural differences between these models, particularly the TTFT sensitivity to input image resolution, illustrating the unique challenges these variations present for quantization and efficient inference.

\subsection{Quantization Granularity} 
\label{quant_granularity}
Furthermore, as mentioned in SoomthQuant ~\citep{xiao2022smoothquant}, there are different different granularity levels. The \textbf{per-tensor static} quantization uses a single step size for the entire matrix. \textbf{Per-token dynamic} quantization employs different $s$ for the activations associated with each token, being a common granularity for activations quantization of existing LLMs. For weights, per-channel quantization applies distinct $s$ for each output channel of the weights, while group-wise quantization utilizes a coarse-grained $s$ for different channel groups. Notably, group-wise quantization is a prevalent granularity for weight quantization in LLMs ~\citep{frantar2022gptq}.

\subsection{LayerNorm, RMSNorm and Computational Invariance}
\label{section:background-norm}
We introduce LayerNorm, RMSNorm, computational Invariance, and their usage in Transformers.

\textbf{Layer Normalization} (LayerNorm, LN) \citep{layernorm} is a technique to normalize the activations of intermediate layers of neural networks.
Given a vector $\mathbf{x} \in \mathbb{R}^d$, LayerNorm normalizes it to obtain a zero-mean unit-variance vector,
\begin{equation}
    \text{LayerNorm}(\mathbf{x}) = \frac{\mathbf{x} - \mu(\mathbf{x})\mathbf{1}}{\sqrt{\lVert{\mathbf{x}}\lVert_2^2/ d - \mu^2(\mathbf{x}) + \epsilon}}, \text{where } \mu(\mathbf{x}) = \frac{\mathbf{1}^T \mathbf{x}}{d}, \epsilon > 0.
\end{equation}
LayerNorm recenters and rescales the activations and gradients in the forward and backward computations, which enables fast and robust training of neural networks.

\textbf{Root Mean Square Normalization} (RMSNorm) \citep{RMSNorm} is another technique used for normalizing the activations.
It is similar to LayerNorm in that it aims to accelerate and stabilize the training but uses a different normalization approach.
Instead of normalizing the inputs based on their mean and variance, RMSNorm normalizes them based on their root mean square (RMS) value.
It is defined in the following equation,
\begin{equation}
    \text{RMSNorm}(\mathbf{x}) = \frac{\mathbf{x}}{\sqrt{\lVert{\mathbf{x}}\lVert_2^2 / d + \epsilon}}, \text{where } \epsilon > 0.
\end{equation}
RMSNorm only rescales the input vector and the corresponding gradients, discarding the recentering process.
As shown in their definitions, RMSNorm is computationally simpler and more efficient than LayerNorm.
It is reported that replacing LayerNorm with RMSNorm can achieve comparable performance and save training and inference time by $7\%-64\%$ \citep{RMSNorm}.

Given a zero-mean vector $\mathbf{x}$, these two kinds of normalization are equivalent.
Formally, if $\mu(\mathbf{x}) = 0$, then $\text{LayerNorm}(\mathbf{x}) = \text{RMSNorm}(\mathbf{x})$.
We may optionally introduce learnable parameters and apply an element-wise affine transformation on the output of LayerNorm and RMSNorm.

\paragraph{LayerNorm (LN) and RMSNorm} Given the input concated token $\mX$ after embeddings with the shape $L\times D$, the $\mX$ is passed through a LayerNorm~\citep{layernorm} operation, which subtracts the mean from each row of the matrix, divides the row by its standard deviation, rescales (columnwise), and adds an offset. Follow ~\citep{ashkboos2024slicegpt}, we write the LayerNorm block as
\begin{equation}
\textrm{LayerNorm}(\mX) = \textrm{RMSNorm}(\mX \mM)\textrm{diag}(\boldsymbol\alpha)\sqrt{D} + \mathbf{1}_N\boldsymbol\beta^\top
\label{eq:LN}
\end{equation}
where $\textrm{RMSNorm}(\mX)$ applies $\mathbf{x}\leftarrow\mathbf{x}/\Vert\mathbf{x}\Vert$ to each row of $\mX$, and $\mX = concat(\mE_v,\mE_t))$ is the concatenation between text tokens $\mE_t$ and the visual tokens $\mE_v$.
The vector parameter $\boldsymbol\alpha$ and offset (vector) parameter $\boldsymbol\beta$ are learned independently at each LayerNorm instance. The constant matrix $\mM=\mI - \frac{1}{D}\mathbf{1}\mathbf{1}^\top$ is a $D\times D$ matrix which subtracts the mean from each row of $\mX$, called recentering operation. Formally, if $\mM=\mI$, the input $\mX$ has a zero-mean, the Eq \ref{eq:LN} is equivalent to RMSNorm. Specifically, LayerNorm is widely employed in visual encoders $E$, whereas RMSNorm ~\citep{RMSNorm} is commonly used in LLMs~\citep{touvron2023llama, llama3} and has been shown to accelerate training and inference time with similar performance \citep{RMSNorm}.

\paragraph{Computational Invariance in RMSNorm.} 
Based on the \emph{computational invariance}, recent studies~\citep{ashkboos2024slicegpt, ashkboos2024quarot} have shown that orthogonal transformations can effectively smooth outliers and improve the quantize-ability of both weights and activations. In particular, for transformers, inserting linear layers with an orthogonal matrices $\mQ$ before and after the RMSNorm~\citep{RMSNorm} layer in a transformer, the network remains unchanged. In detail, given the input $\mX$ and orthogonal matrix $\mQ$ for RMSNorm layer, the \emph{computational invariance} means: $\textrm{RMSNorm}(\mX \mQ)\mQ^\top = \textrm{RMSNorm}$. Here, $\mQ^\top\mQ = \mQ\mQ^\top = \mI$ and a rotation matrix is an orthogonal matrix with $|\mQ| = 1$. Note that multiplying a vector $\mathbf{x}$ by $\mQ$ does not change the norm of the vector, since $\Vert\mQ\mathbf{x}\Vert = \sqrt{\mathbf{x}^\top\mQ^\top \mQ\mathbf{x}} = \sqrt{\mathbf{x}^\top\mathbf{x}} = \Vert\mathbf{x}\Vert$.

\subsection{LayerNorm to RMSNorm Transformation.}
\label{pre-LN2RMSN}

\begin{figure*}[h]
    \centering    
    \includegraphics[width=0.9\textwidth]{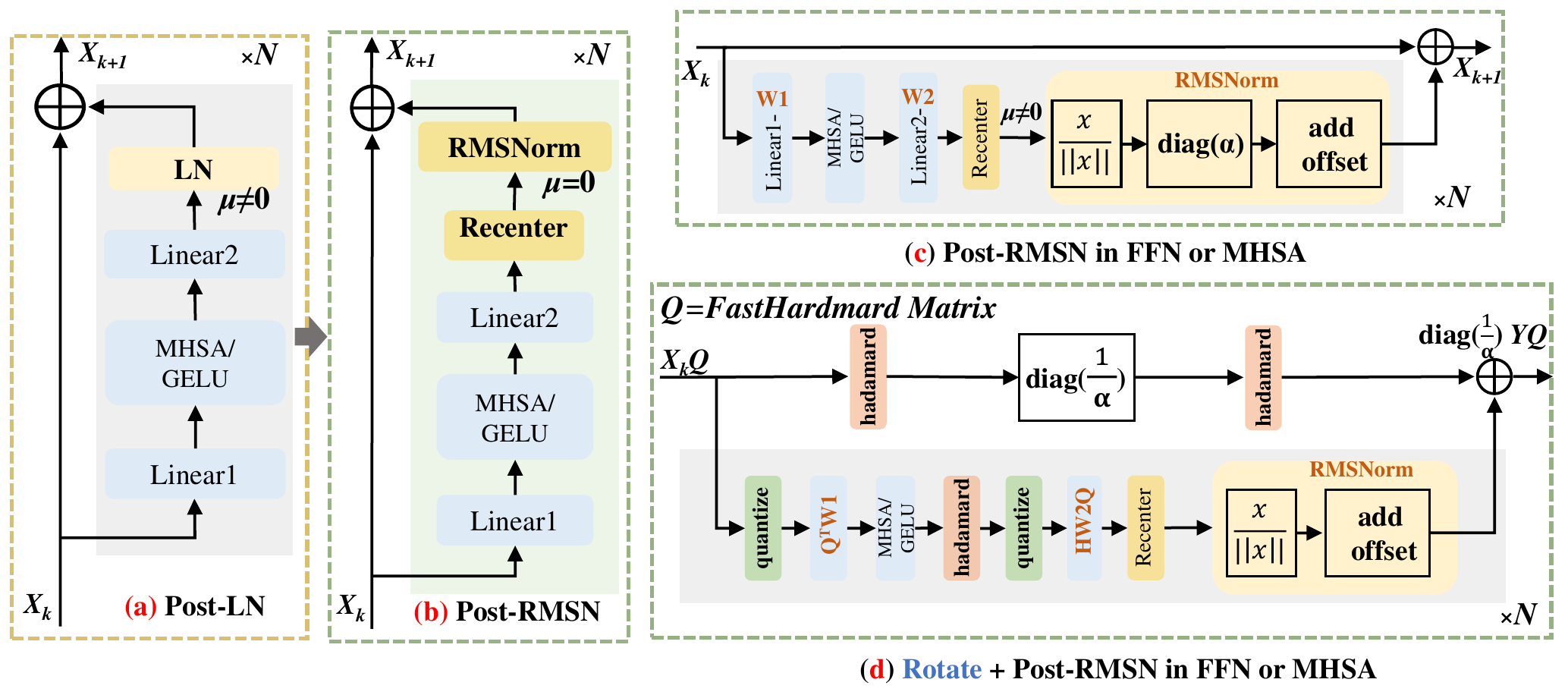}
    \vspace{-4mm}
    \caption{The proposed Post-LN + Rotate Scheme.} 
    \label{fig:post-LN+RMSNorm}
\vspace{-2mm}
\end{figure*}
\paragraph{Post-LayerNorm to RMSNorm Transformation.} As shown in Figure~\ref{fig:post-LN+RMSNorm}, we present the detailed Post-LN + Rotate design. Unlike the Pre-LN + Rotate in SliceGPT~\cite{ashkboos2024slicegpt}, our Post-LN + Rotate scheme extends the applicability for different MLLMs' vision encoder.  

\begin{wrapfigure}{r}{6.5cm}
  % \vspace{-3mm}
  % \begin{adjustbox}{max width=\linewidth}
    \includegraphics[width=1.0\linewidth]{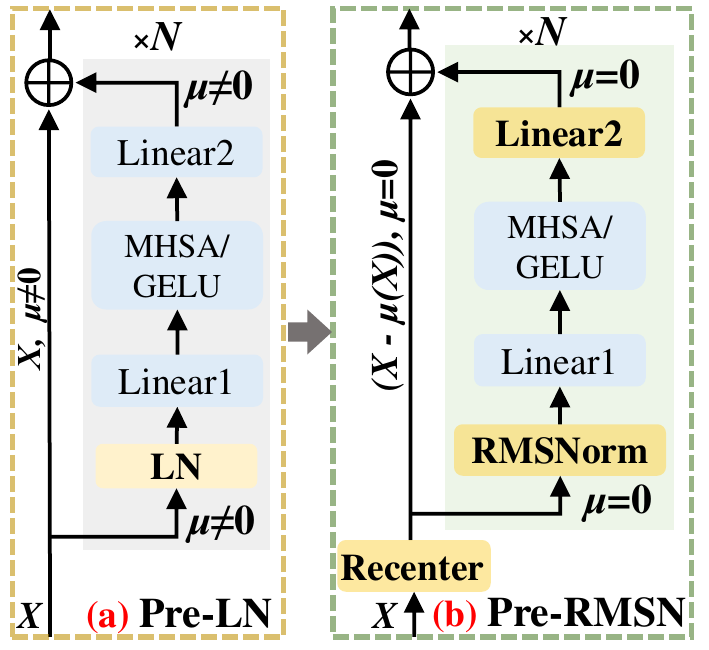}
    \caption{The illustration of transformation from Pre-LN to RMSNorm.}
  % \end{adjustbox}
  % \vspace{-4mm}
  \label{preLN2RMSNorm}
  \vspace{-4mm}
\end{wrapfigure}

\paragraph{Pre-LayerNorm to RMSNorm Transformation.} Here, we propose unified LayerNorm-to-RMSNorm transformation, aiming to synthesize the transformer architecture of MLLMs' vision encoders and LLMs, endowing them with rotation-friendly characteristics that facilitate the effective removal of outliers. We take Pre-LN transformer as an example to show that how transform Pre-LN into RMSNorm layer while guaranting arithmetic equivalence. As shown in Figure \ref{preLN2RMSNorm} (a), for the input $\mX_k$ of the $k$-th block, the main block in pre-LN transformer is $\mX_{k+1} = \mX_k + \ell_2(g(\ell_1(LN(\mX_k))))$, where $k \in [1, N]$, and $N$ is the block number. If $g$ is an activation function, such as GELU, this block is a multi-layer perceptron (MLP) module. If $g$ is a multi-head attention, then this block is the casual attention module \citep{vaswani2017attention}. Due to the recentering operation, LN exhibits invariance to shifts, such that $LN(\mX_k - a\mathbf{1}) = LN(\mX_k), \forall a \in \mathbb{R}$. 
 Therefore, as shown in Figure \ref{preLN2RMSNorm} (b), we can replace LN as RMSNorm layer through two modifications: \raisebox{-0.5pt}{\ding[1.1]{182\relax}} recenter the input $\mX_k$ to $\mX_k - \mu(\mX_k)\mathbf{1}$, ensuring that the input to norm layer maintain a zero mean. \raisebox{-0.5pt}{\ding[1.1]{183\relax}} adjust the weights $\mA_2$ and bias $\mathbf{b}_2$ of the the linear $\ell_2$ to $\hat{\mA_2} = \mA_2 - \frac{1}{D}\mathbf{1} \mathbf{1}^T \mA_2, \hat{\mathbf{b}_2} = \mathbf{b}_2 - \mu(\mathbf{b}_2)\mathbf{1}$. Consequently, the LN can be replaced with an RMSNorm layer with the same arithmetic functionality. The first operation is to recenter $\mX_{k}$, while the second operation is to recenter the output of main branches. Notably, since $\mX_{k+1} = \mX_k + \ell_2(g(\ell_1(LN(\mX_k))))$, after applying \raisebox{-0.5pt}{\ding[1.1]{182\relax}} and \raisebox{-0.5pt}{\ding[1.1]{183\relax}}, the input and the output of main branch are re-centered with zero-mean, while the input of residual branches also maintain a zero mean. Therefore, the output after current blocks, $\mX_{k+1}$ (which serves as the input for next block), still maintain zero-mean. A detailed proof is provided in the Appendix. Ultimately, we establish the equivalence of Pre-LN and Pre-RMSNorm Transformers. Now that every LayerNorm in the transformer has been converted to RMSNorm in MLLMs, we can use any orthogonal matrices $\mQ$ to the model. Therefore, the visual encoder and LLMs are in a rotation-friendly RMSNorm-only transformer architecture.
% \section{Research Methods}
% \subsection{Part One}

\begin{figure*}[h]
    \centering    
    \vspace{-3mm}
    \includegraphics[width=0.9\linewidth]{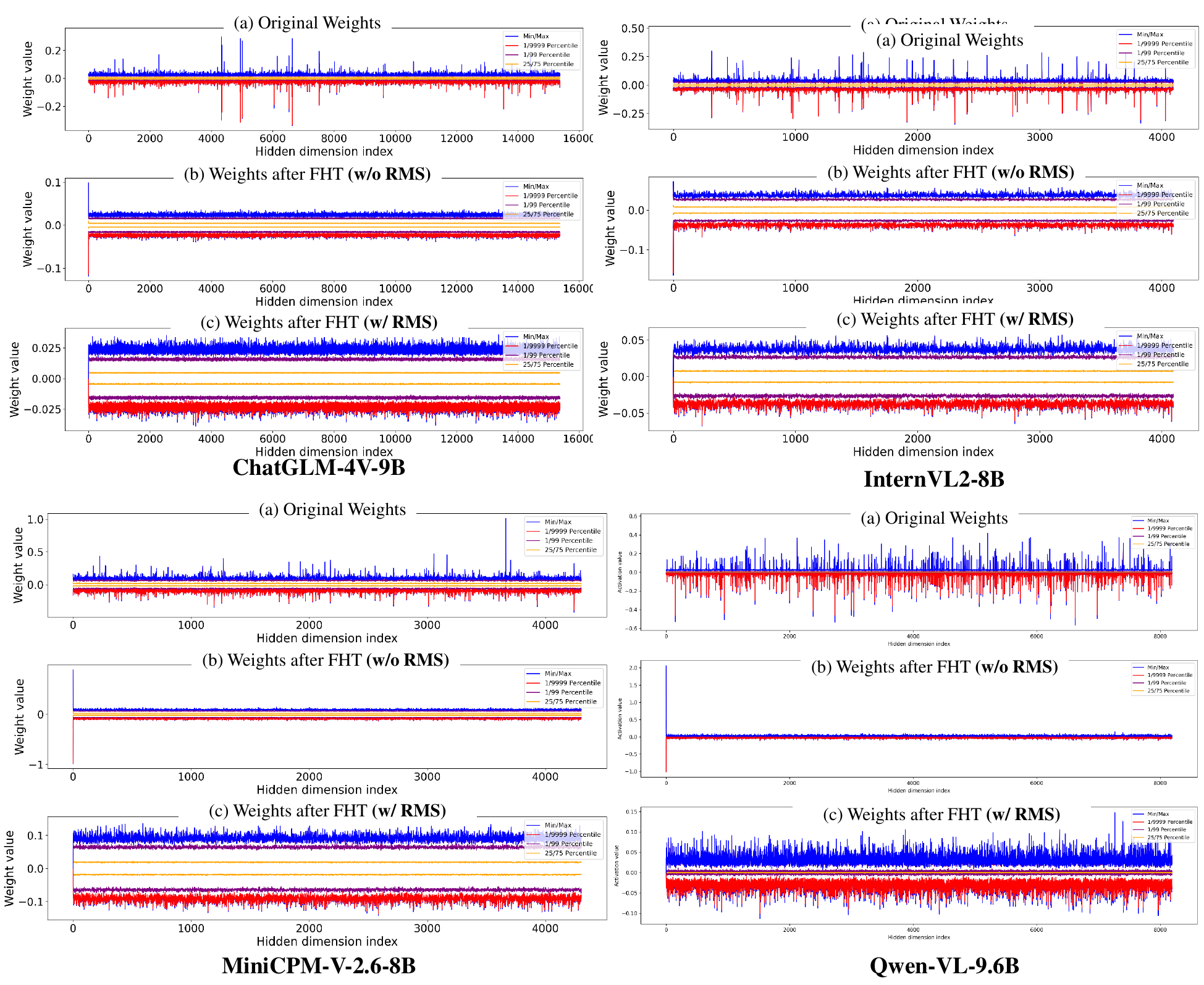}
    \vspace{-5mm}
    \caption{Illustration the weight distributions for the down-proj layer in different MLLMs's visual encoder under three conditions: (a) original weight, (b) weights after FHT, and (c) weight after FHT with our RMS.} 
    \label{fig:RMS_weight_ablation}
\vspace{-3mm}
\end{figure*}

\subsection{Weight Distribution Visualization with RMS.} 
\label{weight_distritbuion}
To facilitate a better analysis of RMS, we also visualize weight distribution across other mainstream MLLMs (InternVL, Qwen-VL, MiniCPM-V and ChatVLM2). As shown in Figure ~\ref{fig:RMS_weight_ablation}, we can observe that the FHT-induced weight outliers presents at all MLLMs. The weight distribution reveals that FHT amplification increases weight magnitudes, while RMS effectively reduces these magnitudes, significantly improving quantization stability. 

\subsection{Discussion About Video-based MLLMs.}
\label{video_MLLMs}
In video-based MLLMs like Video-LLaMA~\citep{zhang2023video} and Qwen2-VL~\citep{Qwen2VL}, the pipeline involves sampling video frames at fixed intervals, extracting visual features per frame using a vision encoder, and modeling temporal relationships with downstream components (e.g., via temporal positional encoding in Video-LLaMA or 3D convolution in Qwen2-VL). Two key features emerge: (1) videos are processed as multi-frame image sequences, and (2) Our Modality-specific static quantization is not affected by the temporal-aware module. Therefore, for an input video, MQuant’s per-modality static quantization applies naturally: we compute quantization parameters offline across these visual frames. Specifically, the dynamic nature of visual tokens is handled as an image sequence, not altering our static quantization’s efficiency or accuracy in MQuant (Section 3).
We recognize this as a promising direction for future work.

\end{document}